\documentclass[acmjacm,acmnow,nohyperref]{acmtrans2m}

\usepackage{amsmath}
\usepackage{amsfonts}
\usepackage{amssymb}
\allowdisplaybreaks[2]

\ifx\pdfoutput\undefined
  \usepackage[dvips]{color}
  \usepackage{graphicx}
  \DeclareGraphicsExtensions{.eps}
\else
 \usepackage[pdftex]{color}
 \usepackage[pdftex]{graphicx}
 \DeclareGraphicsExtensions{.pdf}
\fi

\newcommand{\thmcolon}{}
\newtheorem{THEOREM}{Theorem}[section]
\newenvironment{theorem}{\begin{THEOREM} \thmcolon  }%
                        {\end{THEOREM}}
\newtheorem{LEMMA}[THEOREM]{Lemma}
\newenvironment{lemma}{\begin{LEMMA} \thmcolon  }%
                      {\end{LEMMA}}
\newtheorem{COROLLARY}[THEOREM]{Corollary}
\newenvironment{corollary}{\begin{COROLLARY} \thmcolon  }%
                          {\end{COROLLARY}}
\newtheorem{PROPOSITION}[THEOREM]{Proposition}
\newenvironment{proposition}{\begin{PROPOSITION} \thmcolon  }%
                            {\end{PROPOSITION}}
\newtheorem{DEFINITION}[THEOREM]{Definition}
\newenvironment{definition}{\begin{DEFINITION} \thmcolon  \rm}%
                            {\end{DEFINITION}}
\newtheorem{CLAIM}[THEOREM]{Claim}
                            {\end{CLAIM}}
\newtheorem{EXAMPLE}[THEOREM]{Example}
\newenvironment{example}{\begin{EXAMPLE} \thmcolon  \rm}%
                            {\end{EXAMPLE}}
\newtheorem{REMARK}[THEOREM]{Remark}
\newenvironment{remark}{\begin{REMARK} \thmcolon  \rm}%
                            {\end{REMARK}}

\newcommand{\thm}{\begin{theorem}}
\newcommand{\lem}{\begin{lemma}}
\newcommand{\pro}{\begin{proposition}}
\newcommand{\dfn}{\begin{definition}}
\newcommand{\rem}{\begin{remark}}
\newcommand{\xam}{\begin{example}}
\newcommand{\cor}{\begin{corollary}}
\newcommand{\prf}{\begin{proof}}
\newcommand{\ethm}{\end{theorem}}
\newcommand{\elem}{\end{lemma}}
\newcommand{\epro}{\end{proposition}}
\newcommand{\edfn}{\bbox\end{definition}}
\newcommand{\erem}{\bbox\end{remark}}
\newcommand{\exam}{\bbox\end{example}}
\newcommand{\ecor}{\end{corollary}}
\newcommand{\eprf}{\end{proof}}
\newcommand{\beqn}{\begin{equation}}
\newcommand{\eeqn}{\end{equation}}
\newcommand{\bbox}{\vrule height7pt width4pt depth1pt}
\newenvironment{oldtheorem}[1]
  {\begin{renewcommand}{\theTHEOREM}{#1}}
  {\end{renewcommand}\addtocounter{THEOREM}{-1}}
\newenvironment{oldthm}[1]{\begin{oldtheorem}{#1}\thm}{\ethm\end{oldtheorem}}
\newenvironment{oldlem}[1]{\begin{oldtheorem}{#1}\lem}{\elem\end{oldtheorem}}
\newenvironment{oldcor}[1]{\begin{oldtheorem}{#1}\cor}{\ecor\end{oldtheorem}}
\newenvironment{oldpro}[1]{\begin{oldtheorem}{#1}\pro}{\epro\end{oldtheorem}}
\newcommand{\othm}[1]{\begin{oldthm}{\ref{#1}}}
\newcommand{\eothm}{\end{oldthm}}
\newcommand{\olem}[1]{\begin{oldlem}{\ref{#1}}}
\newcommand{\eolem}{\end{oldlem}}
\newcommand{\ocor}[1]{\begin{oldcor}{\ref{#1}}}
\newcommand{\eocor}{\end{oldcor}}
\newcommand{\opro}[1]{\begin{oldpro}{\ref{#1}}}
\newcommand{\eopro}{\end{oldpro}}
\newcommand{\nthm}[1]{\begin{oldthm}{#1}}
\newcommand{\enthm}{\end{oldthm} \medskip}

\newcommand{\cL}{\mathcal{L}}
\newcommand{\cM}{\mathcal{M}}
\newcommand{\ax}{{\rm\bf AX}}
\newcommand{\MProb}{\cM^{\mathit{prob}}}
\newcommand{\MBel}{\cM^{\mathit{bel}}}
\newcommand{\MPoss}{\cM^{\mathit{poss}}}
\newcommand{\MLP}{\cM^{\mathit{lp}}}
\newcommand{\AXProb}{\ax^{\mathit{prob}}}
\newcommand{\AXBel}{\ax^{\mathit{bel}}}
\newcommand{\AXPoss}{\ax^{\mathit{poss}}}
\newcommand{\AXLP}{\ax^{\mathit{lp}}}

\newcommand{\Lg}{\cL^g}
\newcommand{\Mg}{\cM^g}
\newcommand{\AXg}{\ax^{\mathit{g}}}
\newcommand{\AXf}{\ax^{\mathit{f}}}

\newcommand{\sat}{\models}

\newcommand{\rimp}{\Rightarrow}
\newcommand{\dimp}{\Leftrightarrow}
\newcommand{\bor}{\bigvee}
\newcommand{\union}{\cup}
\newcommand{\inter}{\cap}
\renewcommand{\phi}{\varphi}

\newcommand{\gt}{>}

\newcommand{\F}{{\cal F}}

\newcommand{\M}{{\cal M}}

\renewcommand{\P}{{\cal P}}

\newcommand{\R}{{\cal R}}

\newcommand{\V}{{\cal V}}

\newcommand{\<}{\langle}
\renewcommand{\>}{\rangle}

\newcommand{\respc}{resp.,\ }

\newcommand{\ol}{\setlength{\itemsep}{0pt}\begin{enumerate}}
\newcommand{\eol}{\end{enumerate}\setlength{\itemsep}{-\parsep}}
\newcommand{\ul}{\setlength{\itemsep}{0pt}\begin{itemize}}
\newcommand{\dl}{\setlength{\itemsep}{0pt}\begin{description}}
\newcommand{\edl}{\end{description}\setlength{\itemsep}{-\parsep}}
\newcommand{\eul}{\end{itemize}\setlength{\itemsep}{-\parsep}}

\newcommand{\commentout}[1]{}

\newcommand{\bi}{\begin{itemize}}
\newcommand{\ei}{\end{itemize}}
\newcommand{\be}{\begin{enumerate}}
\newcommand{\ee}{\end{enumerate}}

\newcommand{\LPP}{{\cal L}^{QU}}
\renewcommand{\L}{{\cal L}}

\newcommand{\Poss}{{\rm Poss}}

\newcommand{\intension}[1]{[\![ #1 ]\!]}

\newcommand{\Worlds}{W}

\newcommand{\mi}{\ell}

\newcommand{\Bel}{{\rm Bel}}
\newcommand{\Plaus}{{\rm Plaus}}

\newcommand{\lE}{\underline{E}}
\newcommand{\uE}{\overline{E}}

\newcommand{\COMMENTOUT}[1]{}

\newcommand{\cP}{\ensuremath{\mathcal{P}}}

\newcommand{\axiom}[1]{\textbf{#1}}
\newcommand{\riff}{\ensuremath{\Leftrightarrow}}
\newcommand{\sep}{\ensuremath{~:~}}
\newcommand{\truep}{\mbox{\textit{true}}}
\newcommand{\falsep}{\mbox{\textit{false}}}
\newcommand{\true}{\mbox{\textbf{true}}}
\newcommand{\false}{\mbox{\textbf{false}}}
\newcommand{\event}[2]{\ensuremath{\intension{#1}_{#2}}}
\newcommand{\eventM}[1]{\ensuremath{\intension{#1}_M}}
\renewcommand{\R}{\mathbb{R}}
\newcommand{\LE}{{\cal L}^E}
\newcommand{\LEp}{{{\cal L}^E}'}

\newcommand{\gamb}[2]{\{\!\hspace{-.7pt}|#1|\!\hspace{-.7pt}\}_{#2}}
\newcommand{\gambM}[1]{\gamb{#1}{M}}

\newcommand{\unPs}{\underline{P}^*}
\newcommand{\unEs}{\underline{E}^*}
\newcommand{\unP}{\underline{P}}
\newcommand{\unE}{\underline{E}}

\renewcommand{\emptyset}{\varnothing}

\newenvironment{wideitemize}[1]
   {\begin{list}{$\bullet$}
                     {\setlength{\labelwidth}{#1}
                      \setlength{\leftmargin}{#1}}}
   {\end{list}}
\newenvironment{axiomlist}
   {\begin{wideitemize}{8ex}}
   {\end{wideitemize}}

\title{Characterizing and Reasoning about Probabilistic and
Non-Probabilistic Expectation} 
            
\author{JOSEPH Y. HALPERN\\
Cornell University
\and
RICCARDO PUCELLA\\
Northeastern University}
            
\begin{abstract} 
Expectation is a central notion in probability theory.
The notion of expectation also makes sense for other notions of
uncertainty.  
We introduce a propositional logic for reasoning about expectation,
where the semantics depends on the underlying representation of
uncertainty.
We give sound and complete axiomatizations for the logic in the case
that the underlying representation is (a) probability, (b) sets of
probability measures, (c) belief functions, and (d) possibility
measures.  
We show that this logic is more expressive than the corresponding
logic for reasoning about likelihood in the case of sets of
probability measures, but equi-expressive in the case of probability,
belief, and possibility. 
Finally, we show that satisfiability for these logics is NP-complete,
no harder than satisfiability for propositional logic. 
\end{abstract}

            
            

\begin{document}

\begin{bottomstuff} 
A preliminary version of this paper appeared in the \emph{Proceedings
of the Eighteenth Conference on Uncertainty in Artificial
Intelligence} \cite{HalPuc02:UAI}.
We thank the UAI and JACM reviewers for their comments.
This work was supported in part by 
NSF under grant 
IIS-0090145, by ONR under grants N00014-00-1-0341,
N00014-01-1-0511,  and N00014-02-1-0455,
by the DoD Multidisciplinary University Research
Initiative (MURI) program administered by the ONR under
grants N00014-97-0505 and N00014-01-1-0795.  In addition, 
while on sabbatical in 2001-02, 
Halpern
was supported by a Guggenheim and a
Fulbright Fellowship. Sabbatical support from CWI and the Hebrew
University of Jerusalem is also gratefully acknowledged.
This work was done while Pucella was at Cornell University. 
Authors' address: J. Y. Halpern, Department of
Computer Science, Cornell University, Ithaca, NY 14853,
email: \texttt{halpern@cs.cornell.edu}, home page:
\texttt{http://www.cs.cornell.edu/home/halpern},  R. Pucella, College of Computer and Information Science, Northeastern University, Boston, MA 02115, email: \texttt{riccardo@ccs.neu.edu}. 
\end{bottomstuff}

\maketitle

\section{Introduction}

One of the most important notions in probability theory is that of
{\em expectation}.  The expected value of a random variable is, in a
sense, the single number that best describes the random variable.
While probability is certainly still the most dominant approach to
representing uncertainty, in the past two decades there has been a
great deal of interest in alternative representations of uncertainty,
both from a normative and descriptive point of view.
This may seem somewhat surprising to those familiar with the many
arguments that have been made showing that probability is the only
rational approach to representing uncertainty (see, for example,
\citeN{Cox}, \citeN{Ram}, \citeN{DeFinetti31}, \citeN{Savage}).  However, all these arguments depend on
assumptions, perhaps the most controversial of which is that the
uncertainty of an event can be completely characterized by a single
number.  (See Walley \citeyear{Walley91} for a good summary of the
arguments for the need to occasionally go beyond probabilistic expectation.)
Some alternatives to probability in the literature include sets of
probability measure 
\cite{Huber81,Walley91}, \emph{Dempster-Shafer belief functions} \cite{Shaf}
and the closely related \emph{nonadditive measures} \cite{Schmeidler89}, 
and possibility measures \cite{DuboisPrade88}.

In this paper, we consider the notion of expectation
for all these representations of uncertainty.  We do not take a stand
here on what the ``right'' way is to represent uncertainty; we simply
investigate characterizations of expectation and reasoning about
expectation, both for probability and for other representations of
uncertainty.  

It is well known that a probability
measure determines a unique expectation function that is linear
(i.e., $E(aX + bY) = aE(X) + bE(Y)$),
monotone 
(i.e., $X \le Y$ implies $E(X) \le E(Y)$),
and maps constant functions to their value.  Conversely, given an
expectation function $E$ (that is, a function from random variables to the
reals) that is linear, monotone, and maps constant functions to their
value, there is a unique probability measure $\mu$ such that $E =
E_\mu$.    That is, there is a 1-1 mapping from probability measures to
(probabilistic) expectation functions.
One of the goals of this paper is to provide similar characterizations
of expectation for other representations of uncertainty.

Some work along these lines has already been done, particulary with
regard to sets of probability measures \cite{Huber81,Walley91,Walley}.%
\footnote{Walley \citeyear{Walley91} actually characterizes lower and upper
\emph{previsions}; but these are  essentially lower and upper 
expectations with respect to sets of probability measures.}
However, there seems to be surprisingly
little work on 
characterizing
expectation in the context of other measures of uncertainty, such as  
belief functions \cite{Shaf} and possibility measures
\cite{DuboisPrade88}.
We provide characterizations here.    

Having characterized expectation functions, we then turn to the
problem of reasoning about 
them.
We define a logic similar in spirit to that introduced by \citeN{FHM} 
(FHM from now on)
for reasoning about likelihood expressed as either probability or belief.
The same logic is used by \citeN{HalPuc00}
(HP from now on)
for reasoning about upper probabilities.
The logic for reasoning about expectation is strictly more expressive
than its counterpart for reasoning about likelihood if the underlying
semantics is given in terms of sets of probability measures (so that
upper probabilities and upper expectations are used, respectively); it
turns out to be equi-expressive in the case of probability, belief
functions, and possibility measures.  This is somewhat surprising,
especially in the case of belief functions.  In all cases, the fact that
expectations are at least as expressive is immediate, since the
expectation of $\phi$ (viewed as an indicator function, that is, the
random variable that is 1 in worlds where $\phi$ is true and 0 otherwise)
is equal to its likelihood.  However, it is not 
always
obvious how to
express
the expectation of a random variable in terms of likelihood.

We then provide 
a sound and complete axiomatization for the logic with respect to each
of the interpretations of expectation that we consider, 
using our characterization of expectation.
Finally, we show that, just as in the case of the corresponding logic
for reasoning about likelihood,
the complexity of the satisfiability problem is NP-complete. 
This is clear when the underlying semantics is given in terms
of probability measures, belief functions, or possibility measures,
but it is 
perhaps surprising that, despite the added expressiveness in the case
of sets of probability measures, reasoning in the logic remains
NP-complete. 

To the best of our knowledge, there is only one previous attempt to
express properties of expectation in a logical setting.  Wilson and
Moral \citeyear{Wilson94} take as 
their starting point Walley's notion of lower and upper previsions.
They consider when acceptance of one set of gambles implies
acceptance of another gamble.  This is a notion that is easily
expressible in our logic when the original set of gambles 
is finite,
so our logic subsumes theirs in the finite case.

This paper is organized as follows. In the next section, 
the characterizations of expectation for probability measures and sets of
probability measures are reviewed, and the characterizations of
expectation for belief functions and possibility measures are provided.
In Section~\ref{s:logics}, we introduce a logic for reasoning about 
expectation with respect to all these representations of uncertainty.
In Section~\ref{sec:expressive}, we compare the
expressive power of our expectation logic to that of the logic for reasoning
about likelihood. In Section~\ref{sec:axiomatization}, we derive sound 
and complete axiomatizations for the logic in Section~\ref{s:logics},
with respect to different representations of uncertainty.  
In Section~\ref{s:decision}, we
prove that the decision problem for the expectation logic is
NP-complete
for each of the representations of uncertainty we consider. 
Finally, in Section~\ref{s:gambleineq}, we discuss an axiomatization
of gamble inequalities, which is assumed by the axiomatizations given
in Section~\ref{sec:axiomatization}.
The proofs of the more technical results  are given in the appendix.

\section{Expectation Functions}\label{s:exp}

Recall that a {\em random variable\/} $X$ on a sample
space (set
of possible worlds) $W$ is a
function from $W$ to some range. 
Let $\V(X)$ denote 
the image of $X$, that is, 
the possible values
of $X$.  
A {\em gamble\/} is a
random variable whose range is the reals.
In this paper, we focus on the expectation of gambles. Additionally,
we restrict to finite sample spaces; most of 
the issues of interest already arise in the finite sample space
setting. 
(Most of the results in this 
section extend in a straightforward way to the infinite sample space
setting, by adding suitable continuity assumptions on the measures
defined. See \citeN{Hal31} for more detail.) Note that if the
sample space is finite, the range $\V(X)$ of a gamble $X$ is
finite. This allows us to define expectation using summation rather
than integration.

\subsection{Expectation for Probability Measures}

Given a finite sample space $W$, and a probability measure $\mu$ and 
gamble $X$ over $W$, the {\em expected value of $X$\/} (or the {\em
expectation of $X$\/}) with respect to $\mu$, denoted $E_\mu(X)$, is
just 
\begin{equation}\label{expdef1} 
\sum_{w \in W} \mu(w) X(w).
\end{equation}
Thus, the expected value of a
gamble is essentially the ``average'' value of the variable.

Actually, (\ref{expdef1}) makes sense only if every singleton is
measurable (i.e., in the domain of $\mu$).
If singletons are not necessarily measurable, the standard assumption is
that $X$ is {\em measurable\/} with respect
to the algebra $\F$ on which $\mu$ is defined; that is, for each
value $x \in \V(X)$, the set of worlds $X=x$ where $X$ takes on value
$x$ is measurable.%
\footnote{Recall that an algebra $\F$ over a sample space $W$ is a set of
subsets of $W$ that includes $W$ itself and is closed under
complementation and union, so that if $U, V \in \F$, then so is
$\overline{U}$ and $U \union V$.}
(In general, a function $f: W \rightarrow W'$ is measurable
with respect to $\F$ if $f^{-1}(w') \in \F$ for all $w' \in W'$.)
Then
\begin{equation}\label{expdef2}
E_\mu(X) = \sum_{x \in \V(X)} x \mu(X=x).
\end{equation}
Note that this definition makes sense even if $W$ is not finite, so
long as $\V(X)$ is finite. It is easy to check that (\ref{expdef1})
and (\ref{expdef2}) are equivalent if $W$ is finite and all singletons
are measurable. 

As is well known, probabilistic expectation functions can be
characterized by a small collection of properties.
If $X$ and $Y$ are gambles on $W$ and $a$ and
$b$ are real numbers, define the gamble $aX +b Y$ on $W$ in the
obvious way: $(aX + bY)(w) = aX(w) + bY(w)$.  Say that $X \le Y$ if
$X(w) \le Y(w)$ for all $w \in W$.
Let $\tilde{c}$ denote the
constant function which always returns $c$; that is, $\tilde{c}(w) = c$.
Let $\mu$ be a probability measure on $W$.

\pro\label{p:probexp-props}
The function $E_\mu$ has the following properties for all measurable
gambles $X$ and~$Y$.
\begin{itemize}
\item[(a)] $E_\mu$ is {\em additive\/}: $E_\mu(X + Y) = E_\mu(X) + E_\mu(Y)$.
\item[(b)] $E_\mu$ is {\em affinely homogeneous\/}: $E_\mu(a
X + \tilde{b}) =  a E_\mu(X) + b$ for all
$a, b \in \R$.
\item[(c)] $E_\mu$ is {\em monotone}: if $X \le Y$, then $E_\mu(X)
\le E_\mu(Y)$.
\end{itemize}
\epro

\prf See any standard text in probability or discrete 
mathematics \cite{Billingsley95}
\eprf

The next result shows that the properties in Proposition~\ref{p:probexp-props}
essentially characterize probabilistic expectation functions.
It too is well known.
We provide the proof here, just to show how the assumptions are used.
In the proof (and throughout the paper) we make use of a special type of
random variable. Let $X_U$ denote the gamble such that $X_U(w) =
1$ if $w \in U$ and $X_U(w) = 0$ if $w \notin U$.  A gamble of the form
$X_U$ is traditionally called an {\em indicator function}.

\thm\label{t:probexp-char}
Suppose that $E$ maps gambles measurable with respect to some
algebra $\F$ 
to $\R$ and $E$ is additive, affinely homogeneous, and monotone.
Then there is a (necessarily unique) probability measure $\mu$ on $\F$ such
that $E = E_\mu$. \ethm

\prf 
Define $\mu(U) = E(X_U)$.  Note that $X_W = \tilde{1}$, so $\mu(W) = 1$,
since $E$ is affinely homogeneous.
Since $X_\emptyset$ is $\tilde{0}$ and $E$ is affinely
homogeneous, it follows that $\mu(\emptyset) =
E(X_\emptyset) = 0$.  $X_\emptyset \le X_U
\le X_W$ for all $U \subseteq W$; since $E$ is monotone, it follows that $0 =
E(X_\emptyset) \le E(X_U) = \mu(U) \le E(X_W) = 1$.  If $U$ and $V$
are disjoint, then it is easy to see that  $X_{U \union V} = X_U +
X_V$.  By additivity, $$\mu(U \union V) = E(X_{U \union V}) =
E(X_U) + E(X_V) = \mu(U) + \mu(V).$$
Thus, $\mu$ is indeed a
probability measure.

To see that $E = E_\mu$, note that it is immediate from (\ref{expdef2})
that $\mu(U) = E_\mu(X_U)$ for $U \in \F$.  Thus, $E_\mu$ and $E$ agree on all
measurable indicator functions.  Every measurable gamble $X$ can be
written as a linear combination of measurable indicator functions.  For
each $a \in \V(X)$, let $U_{X,a} = \{w: X(w) = a\}$.  Since $X$ is a
measurable gamble, $U_{X,a}$ must be in $\F$.  Moreover,
$X = \sum_{a \in \V(X)} a X_{U_{X,a}}$.  By
additivity and
affine homogeneity,
$E_\mu(X) = \sum_{a \in \V(X)} a E(X_{U_{X,a}})$.
By Proposition~\ref{p:probexp-props}, $E_\mu(X) = \sum_{a \in \V(X)} a
E_\mu(X_{U_{X,a}})$. Since $E$ and $E_\mu$ agree on measurable indicator
functions, it follows that $E(X) = E_\mu(X)$.  Thus, $E = E_\mu$ as
desired.

Clearly, if $\mu(U) \ne \mu'(U)$, then $E_\mu(X_U) \ne E_{\mu'}(X_U)$.
Thus, $\mu$ is the unique probability measure on $\F$ such that $E =
E_\mu$.
\eprf

\subsection{Expectation for Sets of Probability Measures}

If $\P$ is a set of probability measures on a space $W$, 
define 
$$\begin{array}{l}
\P_*(U) = \inf\{\mu(U): \mu \in \P\} \mbox{ and }\\
\P^*(U) = \sup\{\mu(U): \mu \in \P\}.
\end{array}$$
$\P_*(U)$ is called the {\em lower
probability\/} of $U$ and $\P^*(U)$ is
called the {\em upper probability\/} of $U$.  Lower and upper
probabilities have been well studied in the literature (see, for example,
\citeN{Borel43}, \citeN{Smith}).

There are straightforward analogues of lower and upper probability
in the context of expectation.
If $\P$ is a set of probability measures such that $X$ is measurable
with respect to each probability measure $\mu \in \P$, then define
$E_\P(X) = \{E_\mu(X): \mu \in \P\}$.  $E_\P(X)$ is a set of
numbers.
Define the {\em lower expectation\/} and {\em upper
expectation\/}
of $X$ with respect to $\P$, denoted $\lE_\P(X)$ and
$\uE_\P(X)$, as the $\inf$ and $\sup$ of the set $E_\P(X)$,
respectively.   

The properties of $\lE_\P$ and $\uE_\P$ are not so
different from those of probabilistic expectation functions.
Note that $E_\mu(X_U) = \mu(U)$.  
Similarly, it is easy to see that $\P_*(U) = \lE_\P(X_U)$ and
$\P^*(U) = \uE_\P(X_U)$.  
Moreover, we have the following analogue of
Propositions~\ref{p:probexp-props}.

\pro\label{p:lpexp-props}
The functions $\lE_\P$ and $\uE_\P$ have the following properties for
all gambles $X$ and~$Y$.
\begin{itemize}
\item[(a)] $\lE_\P(X + Y) \ge
\lE_\P(X) + \lE_\P(Y)$ ({\em superadditivity});\\ 
$\uE_\P(X + Y) \le
\uE_\P(X) + \uE_\P(Y)$ ({\em subadditivity}).
\item[(b)] $\lE_\P$ and $\uE_\P$ are both {\em positively affinely
homogeneous}:
$\lE_\P(a X + \tilde{b}) = a \lE_\P(X) +b$ and
$\uE_\P(a X + \tilde{b}) = a \uE_\P(X) +b$ if $a,b \in \R$, $a \ge 0$.
\item[(c)] $\lE_\P$ and $\uE_\P$ are monotone.
\item[(d)] $\uE_\P(X) = - \lE_\P(-X)$.
\end{itemize}
\epro

\prf This result is also well-known; see \citeN[Section 2.6.1]{Walley91}.
(Walley proves the result for what he calls coherent lower and upper
previsions, but then proves \cite[Section 3.3.4]{Walley91} that these
are equivalent to lower and upper expectations, respectively.)
\eprf

Superadditivity (\respc subadditivity), positive affine
homogeneity, and
monotonicity in fact characterize $\lE_\P$ (\respc $\uE_\P$).

\thm\label{t:lpexp-char} {\rm \cite{Huber81}}
Suppose that $E$ maps gambles  measurable with respect to $\F$
to $\R$ and is superadditive (\respc subadditive), positively affinely homogeneous,
and monotone.  Then there is a set $\P$ of probability measures on $\F$ such
that $E = \lE_\P$ (\respc $E = \uE_\P$).%
\footnote{There is an equivalent characterization of $\lE_\P$, due to
Walley 
\citeyear{Walley91}. 
He shows that $E = \lE_\P$ for some set $\P$ of
probability measures iff $E$ is superadditive, $E(cX) = cE(X)$, and 
$E(X) \ge \inf\{X(w): w \in W\}$.  
There is an analogous characterization of $\uE_\P$.}
\ethm

The set $\P$ constructed in Theorem~\ref{t:lpexp-char} is not unique.
It is not hard to construct sets $\P$ and $\P'$ such that $\P \ne \P'$
but $\lE_{\P} = \lE_{\P'}$.  However, there is a
canonical largest set $\P$ such that $E = \lE_{\P}$; $\P$ consists of
all probability measures $\mu$
such that $E_\mu(X) \ge E(X)$ for all gambles $X$.
This set $\P$ can be shown to be closed and convex.  Indeed, it easily
follows 
that Theorem~\ref{t:lpexp-char} actually provides a 1-1
mapping from closed, convex sets of probability measures to lower/upper
expectations.  Moreover, in a precise sense, this is the best we can
do.  If $\P$ and $\P'$ have the same convex closure (where the {\em convex
closure\/} of a set is the smallest closed, convex set containing it),
then $\lE_{\P} = \lE_{\P'}$.

As Walley \citeyear{Walley91} shows, 
what he calls {\em coherent lower/upper previsions\/} are also lower/upper
expectations with respect to some set of probability measures.
Thus, lower/upper previsions can be identified with closed,
convex sets of probability measures.

\subsection{Expectation for Belief Functions}\label{s:belief-exp}

As is well known, a belief function \cite{Shaf} $\Bel$ is a function
from subsets of a state space $W$ to $[0,1]$ satisfying the following
three properties:
\begin{axiomlist}
\item[B1.] $\Bel(\emptyset) = 0$.
\item[B2.] $\Bel(W) = 1$.
\item[B3.]
For $n = 1, 2, 3, \ldots$, \\ \mbox{}\quad
$\Bel(\bigcup_{i=1}^n U_i) \ge \sum_{i=1}^n \sum_{\{I \subseteq \{1,
\ldots, n\}: |I| = i\}} (-1)^{i+1} \Bel(\bigcap_{j \in I} U_j).$ 
\end{axiomlist}

Given a belief function $\Bel$, there is a corresponding {\em
plausibility function\/} $\Plaus$, where $\Plaus(U) = 1 -
\Bel(\overline{U})$.  It follows easily from B3 that $\Bel(U) \le
\Plaus(U)$ for all $U \subseteq W$.  $\Bel(U)$ can be
thought of as a lower bound of a set of probabilities and $\Plaus(U)$
can be thought of as the corresponding upper bound.  This intuition is
made precise in the following well-known result.

\thm\label{t:bel-as-lp} {\rm \cite{Demp1}}
Given a belief function $\Bel$ defined on $W$, let $\P_{\Bel} = \{\mu:
\mu(U) \ge \Bel(U) \mbox{ for all } U \subseteq W\}$. Then
$\Bel = (\P_{\Bel})_*$.%
\footnote{Dempster \citeyear{Demp1} defines 
$\P_{\Bel}$ as $\{\mu: \Plaus(U) \ge \mu(U) \ge \Bel(U) \mbox{ for all } U
\subseteq W\}$, but his definition  is easily seen to be equivalent to
that given here.  For if $\mu(U) \ge
\Bel(U)$ for all $U \subseteq W$ then, in particular, $\mu(\overline{U}) \ge
Bel(\overline{U})$, so $\Plaus(U) = 1 - \Bel(\overline{U}) \ge
1 - \mu(\overline{U}) = \mu(U)$.}
\ethm

Similarly, we can check that $\Plaus(U)=1-\Bel(\overline{U})=(\P_{\Bel})^*$. 
There is an obvious way to define a notion of
expectation based on belief functions, using the identification of
$\Bel$ with $(\P_{\Bel})_*$.
Given a belief function $\Bel$, define $E_{\Bel} = \lE_{\P_{\Bel}}$.
Similarly, for the corresponding plausibility function $\Plaus$, define
$E_{\Plaus} = \uE_{\P_{\Bel}}$.  (These definitions are in fact used by
Dempster~\citeyear{Demp1}).

This is well defined, but it
seems more natural to get a notion of expectation for belief functions
that is defined purely in terms of belief functions, without
reverting to probability.  One way of doing so is due to Choquet
\citeyear{Choq}.%
\footnote{Choquet actually talked about $k$-monotone capacities, which
are essentially functions that satisfy B3 where $n = 1, \ldots, k$.
Belief functions are infinitely monotone
capacities.}  

\begin{figure*}[t]
\begin{center}
\includegraphics[width=3in]{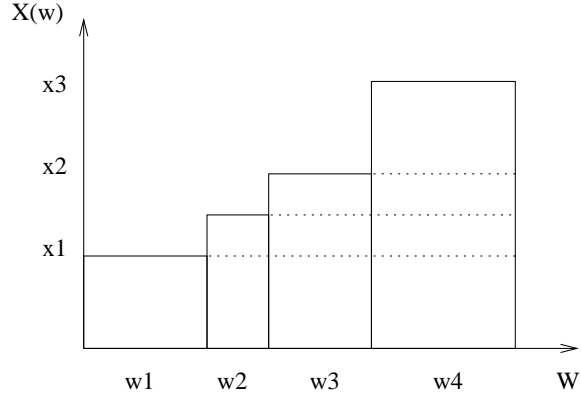}
\end{center}
\caption{Two equivalent definitions of probabilistic expectation.}
\label{f:alternate-definition}
\end{figure*}

It takes as its point of departure
the following alternate definition of expectation in the case of
probability.  Suppose 
that
$X$ is a gamble such that  $\V(X) = \{x_1,
\ldots, x_n\}$, with $x_1 < \ldots < x_n$.
\pro\label{Choqueteq}
$E_\mu(X) = x_1 + (x_2-x_1)\mu(X > x_1) +
\cdots +  (x_n - x_{n-1})\mu(X > x_{n-1}).$
\epro

\prf Figure~\ref{f:alternate-definition} should help make clear why this result is true,
where we assume for simplicity that the worlds $W=\{w_1,w_2,\ldots\}$
are ordered such that if $X(w_i) < X(w_j)$, then $i < j$.
If we assume that the probability of a world $w_i$ is the width of the
vertical rectangle over $w_i$, then it should be clear that the total
area of the rectangles represent the expectation of $X$. 
Notice that the vertical rectangles determine the expectation using the
standard definition (\ref{expdef2}), while the horizontal rectangles
determine the expectation using the formula in this proposition.

For a more formal proof, suppose that $\V(X) = \{x_1, \ldots, x_n\}$,
where $x_1 < \ldots < x_n$.
We proceed by induction on $n$.  If $n=1$, the result is trivial, since
clearly $E_\mu(X) = x_1$.  If $n > 1$, then note that $X = X_1 + X_2$,
where $X_1(w) = X(w)$ if $X(w) \ne x_{n}$, and $X_1(w) = x_{n-1}$ if
$X(w) = x_n$, and $X_2(w) = 0$ if $X(w) \ne x_n$, and $X(w) = x_n -
x_{n-1}$ if $X(w) = x_n$.  Clearly 
$X = X_1 + X_2$, so  by additivity, $E(X) = E(X_1) + E(X_2)$.  Note that
$\V(X_1) = \{x_1, \ldots, x_{n-1}\}$, so by the induction  hypothesis,
$E_\mu(X_1) = x_1 + (x_2-x_1)\mu(X > x_1) +
\cdots +  (x_{n-1} - x_{n-2})\mu(X > x_{n-2}).$  Finally, it is
immediate from the definition of $X_2$ that $E_\mu(X_2) = (x_n -
x_{n-1}) \mu(X > x_{n-1})$.  The result follows immediately.
\eprf

Define
\begin{equation}\label{Choqueteq'}
\begin{array}{ll}
E_{\Bel}'(X) = x_1 + (x_2-x_1)\Bel(X > x_1) + \cdots +  (x_n -
x_{n-1})\Bel(X > x_{n-1}). 
\end{array}
\end{equation}
An analogous definition holds for plausibility:
\begin{equation}\label{plausex}
\begin{array}{ll}
E_{\Plaus}'(X) 
= x_1 + (x_2-x_1)\Plaus(X > x_1) + \cdots +  (x_n - x_{n-1})\Plaus(X >
x_{n-1}). 
\end{array}
\end{equation}

\pro\label{p:schmeidler}
{\rm \cite{Schmeidler89}} $E_{\Bel} = E_{\Bel}'$ and
$E_{\Plaus} = E_{\Plaus}'$.
\epro

Schmeidler \citeyear{Schmeidler86,Schmeidler89} actually used Choquet's
definition to define a notion of expectation for what he called {\em
nonadditive probabilities}, where a nonadditive probability $\nu$
maps subsets of a space $W$ to $[0,1]$ such that $\nu(\emptyset) = 0$,
$\nu(W) = 1$, and $\nu(U) \le \nu(V)$ if $U \subseteq V$.  He proved an
analogue of Proposition~\ref{p:schmeidler} for arbitrary nonadditive
probabilities.  Since belief functions and plausibility functions are
both nonadditive probabilities in Schmeidler's sense,
Proposition~\ref{p:schmeidler} is actually a special case of
Schmeidler's result.

Proposition~\ref{p:schmeidler} shows that (\ref{Choqueteq'}) gives a way
of defining expectation for 
belief functions without referring to probability.
There is yet a third way of defining expectation for belief functions,
which also does not use probability; see the proof of
Theorem~\ref{t:smallmodel2} in Appendix~\ref{appa6}.

Since $E_{\Bel}$ can be viewed as a special case of the lower
expectation $\lE_{\P}$ (taking $\P = \P_{\Bel}$), it is immediate from
Proposition~\ref{p:lpexp-props} that $E_{\Bel}$ is superadditive,
positively affinely homogeneous, and monotone.  (Similar remarks hold
for $E_{\Plaus}$, except that it is subadditive.  For ease of
exposition, we focus on $E_{\Bel}$ in the remainder of this section,
although analogous remarks hold for $E_{\Plaus}$.)

Since it is immediate from
the definition that $E_{\Bel}(X_U) = \Bel(U)$, the
inclusion-exclusion
property B3 of belief
functions can be expressed in terms of expectation (just  by replacing
all instances of $\Bel(V)$ in B3 by $E_{\Bel}(X_V)$).  Moreover, it does
not follow from the other properties, since it can be shown not to hold for
arbitrary lower probabilities.
This restatement of the inclusion-exclusion property
applies only to the expectation of indicator functions.  But, in fact, a
more general 
inclusion-exclusion property holds for $E_{\Bel}$.  
Given gambles $X$
and $Y$, define the gambles $X \land Y$ and $X \lor Y$ as the minimum
and maximum of $X$ and $Y$, respectively; that is, $(X \land Y)(w) =
\min(X(w),Y(w))$ and $(X \lor Y)(w) = \max(X(w),Y(w))$.
Consider the
following inclusion-exclusion property for expectation
functions:
\begin{equation}\label{inex}
E(\lor_{i=1}^n X_i) \ge \sum_{i=1}^n \sum_{\{I \subseteq \{1, \ldots,
n\}: |I| = i\}} (-1)^{i+1} E(\land_{j \in I} X_j). 
\end{equation}
Since it is immediate that $X_{U \union V} = X_U \lor X_V$ and $X_{U
\inter V} = X_U \land X_V$, (\ref{inex}) generalizes B3.
We recover B3 from (\ref{inex}) by taking $X_i$ to be the indicator
function $X_{U_i}$.

There is yet another property satisfied by expectation functions based
on belief functions.  Two gambles $X$ and $Y$ are said to be
\emph{comonotonic} if it is not the case that one increases while the
other decreases. Formally, this means that there do not exist worlds
$w$ and $w'$ such that $(X(w)-X(w'))(Y(w)-Y(w'))<0$. The property
satisfied by expectation based on belief functions is called
\emph{comonotonic additivity}: 
\begin{equation}
\label{belexprop2}
\begin{array}{l}
\mbox{If $X$ and $Y$ are comonotonic, then $E(X+Y)=E(X)+E(Y)$.}
\end{array}
\end{equation}
The fact that $E_{\Bel}$ satisfies this property was essentially
recognized by Dellacherie \citeyear{Dellacherie70}.

\pro\label{p:belexp-props}
The function $E_{\Bel}$ is superadditive, positively affinely homogeneous,
 monotone, and satisfies
(\ref{inex}) and (\ref{belexprop2}).%
\footnote{
In a preliminary version of this paper
\cite{HalPuc02:UAI} we used a different characterization of
expectation functions, based on belief functions. The current
characterization is much simpler.}
\epro

\prf See Appendix~\ref{appa2}. \eprf

\thm\label{t:belexp-char}
Suppose that $E$ is an expectation function
that is positively affinely homogeneous,
monotone, and satisfies (\ref{inex}) and (\ref{belexprop2}).
Then there is a (necessarily unique) belief function $\Bel$
such that $E = E_{\Bel}$. \ethm

\prf See Appendix~\ref{appa2}.  \eprf

Schmeidler~\citeyear{Schmeidler86} proved that expectation functions for
nonadditive measures are characterized by positive affine homogeneity,
monotonicity, and comonotonicity.  Theorem~\ref{t:belexp-char} shows
that the inclusion-exclusion property
(\ref{inex}) is what distinguishes expectation for belief functions from
expectation for arbitrary nonadditive measures.
Note that superadditivity was not assumed in the statement of
Theorem~\ref{t:belexp-char}.  Indeed, it is a consequence of
Theorem~\ref{t:belexp-char} that superadditivity follows from the other
properties. In fact, the full stength of positive affine homogeneity
is not needed either in Theorem~\ref{t:belexp-char}. It suffices to assume
that $E(\tilde{b})=b$.

\cor\label{cor:belunique} Given a belief function $\Bel$, $E_{\Bel}$ is the
unique expectation function
$E$ that is superadditive, positively affinely homogeneous, monotone,
and satisfies (\ref{inex}) and (\ref{belexprop2}) such that $E(X_U) =
\Bel(U)$ for all $U \subseteq W$.
\ecor
\prf
Proposition~\ref{p:belexp-props} shows that $E_{\Bel}$ has 
the required properties.  If $E'$ is
an expectation function that has these properties, by
Theorem~\ref{t:belexp-char}, $E' = E_{\Bel'}$ for some belief function
$\Bel'$.  Since $E'(X_U) = \Bel'(U) = \Bel(U)$ for all $U \subseteq W$,
it follows that $\Bel = \Bel'$.
\eprf

Corollary~\ref{cor:belunique} is somewhat surprising.  While it is almost
immediate that an additive, affinely homogeneous expectation function
(the type that arises from a probability measure) is determined by its
behavior on indicator functions, it is not at all obvious that a
superadditive, positively affine homogeneous expectation function should
be determined by its behavior on indicator functions.  In fact, in
general it is not; the
inclusion-exclusion property is essential.
Corollary~\ref{cor:belunique}
says that $\Bel$ and $E_{\Bel}$ contain the same
information.  Thus, so do $(\P_{\Bel})_*$ and $\lE_{\P_{\Bel}}$
(since $\Bel = (\P_{\Bel})_*$ and $E_{\Bel} = \lE_{\P_{\Bel}}$).
However, this is not true for
arbitrary sets $\P$ of probability measures, as the following example shows.
\xam\label{xamdetermination} 
Let $W = \{1,2,3\}$.
A probability measure $\mu$ on $W$ can be characterized by a triple
 $(a_1,a_2,a_3)$, where $\mu(i) = a_i$.  Let $\P$ consist of the three
 probability measures $(0,3/8,5/8)$, $(5/8,0,3/8)$, and $(3/8,5/8,0)$.
It is almost immediate that $\P_*$ is 0 on singleton
subsets of $W$ and $\P_* = 3/8$ for doubleton subsets.
Let $\P' = \P \union \{\mu_4\}$, where $\mu_4 = (5/8,3/8,0)$.
It is easy to check that $\P'_* = \P_*$. However, $\lE_\P \ne
\lE_{\P'}$.  In particular, let $X$ be the gamble such that $X(1) = 1$,
$X(2) = 2$, and $X(3) = 3$.  Then $\lE_\P(X) = 13/8$ but $\lE_{\P'}(X) =
11/8$.  Thus, although $\lE_\P$ and $\lE_{\P'}$ agree on indicator
functions, they do not agree on all gambles.  In light of
the discussion above, it should be no surprise that $\P_*$
is not a belief function. 
\exam

\subsection{Expectation for Possibility Measures}\label{s:possexp}

A {\em possibility measure\/} $\Poss$ is a function from subsets of $W$
to $[0,1]$ such that 
\begin{axiomlist}
\item[Poss1.] $\Poss(\emptyset) = 0$.
\item[Poss2.] $\Poss(W) = 1$.
\item[{\rm Poss3.}] $\Poss(U \union V) = \max(\Poss(U),\Poss(V))$ if
$U$ and $V$ are disjoint.
\end{axiomlist}
It is not hard to show that Poss3 implies that $\Poss(U \union V) =
\max(\Poss(U), \Poss(V))$ even when $U$ and $V$ are not disjoint.

It is well known  \cite{DuboisPrade82} that possibility measures are
special cases of plausibility functions.  Thus, (\ref{plausex}) can be
used to define a notion of possibilistic expectation; indeed, this has
been done in the literature \cite{DP87}.  
It is also straightforward to see from Poss3 that the expectation
function $E_{\Poss}$ defined from a possibility measure $\Poss$ in
this way satisfies the following \emph{max property} defined in terms
of indicator functions:
\begin{equation}\label{maxprop}
E_{\Poss}(X_{U \union V}) = \max (E_{\Poss}(X_{U}),
E_{\Poss}(X_{V}). 
\end{equation}

\pro\label{p:possexp-props}
The function $E_{\Poss}$ is positively affinely  homogeneous,
monotone, and satisfies (\ref{belexprop2}) and (\ref{maxprop}).
\epro

\prf See Appendix~\ref{appa2}. \eprf

\thm\label{t:possexp-char}
Suppose that $E$ is an expectation function
that is positively affinely homogeneous,
monotone, and satisfies (\ref{belexprop2}) and (\ref{maxprop}).
Then there is a (necessarily unique) possibility measure $\Poss$
such that $E = E_{\Poss}$. \ethm

\prf See Appendix~\ref{appa2}. \eprf

Note that, although $\Poss$ is a plausibility measure, and thus satisfies
the analogue of (\ref{inex}) with $\ge$ replaced by $\le$, there is no
need to state (\ref{inex}) explicitly; it follows from
(\ref{maxprop}).
Moreover, just as with expectation for belief functions,
it follows from the other properties that $E_{\Poss}$ is subadditive.
(Since a possibility measure is a plausibility function, not a belief
function, the corresponding expectation function is subadditive rather
than superadditive.)

\section{A Logic for Reasoning about Expectation}\label{s:logics}
We now consider a logic for reasoning about expectation.  To set the
stage, we briefly review the FHM logic for reasoning
about likelihood.  

\subsection{Reasoning about Likelihood}\label{s:likelihood}

\renewcommand{\theta}{a}
\renewcommand{\alpha}{b}
The syntax of the FHM logic is straightforward.  Fix a set
$\Phi_0=\{p_1,p_2,\ldots\}$ of \emph{primitive propositions}. 
The choice of primitive propositions is application dependent.  The set
of primitive propositions could include statements such as ``the patient has
cancer'' and ``the patient has fever'' if we are reasoning in a medical
domain, or statements such as ``the price of IBM stock is over \$80'' if
we are considering a financial domain.  The set
$\Phi$ of \emph{propositional formulas} is the closure of $\Phi_0$ under 
$\land$ and $\neg$. 
We can define $\lor$ and $\rimp$ in the usual way; we use the operators
freely throughout the paper.
We assume a special propositional formula
\truep, and abbreviate $\neg\truep$ as
\falsep. 
A \emph{basic likelihood formula} has the form $\theta_1 \mi(\phi_1) + \cdots +
\theta_k \mi(\phi_k) \ge \alpha$, where $\theta_1, \ldots, \theta_k, 
\alpha$ are 
integers
and $\phi_1, \ldots, \phi_k$ are propositional formulas.%
\footnote{As observed in FHM, we gain no further generality by allowing
the coefficients $\theta_1, \ldots, \theta_k$ to be rational numbers,
since for any formula with rational coefficients, we can easily find an
equivalent formula with coefficients that are integers by clearing the
dominator.  There is no difficulty giving semantics to formulas where
the coefficients are arbitrary real numbers
(and, indeed, this is what we did in a preliminary version of the paper
\cite{HalPuc02:UAI}), 
but allowing real numbers
causes problems in the complexity results.
(In the preliminary version, we restricted to integer coefficients for
the complexity results.)}
The
$\mi$ stands for \emph{likelihood}.  
Thus, a basic likelihood formula
talks about a linear combination of likelihood terms of the form
$\mi_i(\phi)$.  A \emph{likelihood formula} is a 
Boolean combination of
basic likelihood formulas.  
Let $\LPP$ be the language consisting of likelihood formulas.
(The QU stands for {\em quantitative uncertainty}.  The name for the
logic is taken from \citeN{Hal31}.)

We use standard abbreviations such as $-\mi(\phi)$ for
$(-1)\mi(\phi)$, and formulas $\mi(\phi_1)\ge\mi(\phi_2)$ for
$\mi(\phi_1)-\mi(\phi_2)\ge 0$. 
In addition, 
we write $a_1\mi(\phi_1)+\ldots+a_k\mi(\phi_k)\le\alpha$ for
$-a_1\mi(\phi_1)-\ldots-a_k\mi(\phi_k)\ge-\alpha$,
$a_1\mi(\phi_1)+\ldots+a_k\mi(\phi_k)>\alpha$ for
$\neg(a_1\mi(\phi_1)+\ldots+a_k\mi(\phi_k)\le\alpha)$,  
$a_1\mi(\phi_1)+\ldots+a_k\mi(\phi_k)<\alpha$ for
$-a_1\mi(\phi_1)-\ldots-a_k\mi(\phi_k)>-\alpha$, and 
$a_1\mi(\phi_1)+\ldots+a_k\mi(\phi_k)=\alpha$ for
$(a_1\mi(\phi_1)+\ldots+a_k\mi(\phi_k)\ge\alpha) \land 
 (a_1\mi(\phi_1)+\ldots+a_k\mi(\phi_k)\le\alpha)$.

The semantics of $\LPP$ depends on how $\mi$ is
interpreted.  In FHM, it is interpreted as a probability measure
and as a belief function; 
in HP, it is interpreted as an
upper probability (determined by a set of probability measures).
Depending on the interpretation, $\mi(\phi)$ is the probability of
$\phi$ (i.e., more 
precisely, the probability of the set of worlds where $\phi$ is true),
the belief in $\phi$, etc.  For example, in the case of probability,
define a \emph{probability structure} to be 
a tuple $M=(\Worlds,\F,\mu,\pi)$, where $\Worlds$ is a (possibly infinite)
set of worlds, $\mu$ is a probability measure whose domain is the algebra $\F$
of subsets 
of $\Worlds$, 
and $\pi$ 
is an \emph{interpretation}, which
associates with each state (or world) in
$\Worlds$ a truth assignment on the primitive propositions in
$\Phi_0$. Thus, $\pi(s)(p)\in\{\true,\false\}$ for
$s\in\Worlds$ and $p\in\Phi_0$. 
We require that primitive propositions be measurable, that is,
that $\{s\in\Worlds \sep \pi(s)(p)=\true\} \in \F$ for all $p \in
\Phi_0$.  
Extend
$\pi(s)$ to a truth assignment on all propositional formulas in 
the
standard way, and associate with each propositional formula the set
$\eventM{\phi}=\{s\in\Worlds \sep \pi(s)(\phi)=\true\}$. 
Note that $\eventM{\phi}$ is measurable for all $\phi$ since $\F$ is an
algebra.  
Then 
\begin{itemize}
\item[] $M\sat \theta_1
\ell(\phi_1)+\cdots+\theta_n\ell(\phi_n)\geq\alpha$ iff
$\theta_1 \mu(\eventM{\phi_1}) + \cdots + \theta_n
\mu(\eventM{\phi_n})\geq\alpha.$
\end{itemize}
The semantics of Boolean combinations of basic likelihood formulas is 
given in the obvious way.

We can similarly give semantics to $\mi$ using lower (or upper) probability.  
Define a \emph{lower probability structure} to be a tuple
$M=(\Worlds,\F,\cP,\pi)$, $\Worlds$, $\F$ and $\pi$ are, as before, a
(possibly infinite) set of worlds, an algebra of subsets of $\Worlds$, and
an interpretation that makes each primitive proposition measurable, and
$\cP$ is a set of probability  
measures over $\F$.
Likelihood is interpreted as lower probability in lower probability
structures:%
\footnote{In HP, we interpreted likelihood as upper
probability.  We interpret it here as lower probability to bring out the
connections to belief, which is an instance of lower probability.  It is
easy to translate from upper probabilities to lower probabilities and
vice versa, since $\P_*(U) = 1 - \P^*(\overline{U})$.}
\begin{itemize}
\item[] $M\models \theta_1 \ell(\phi_1)+\cdots+\theta_n \ell(\phi_n)\geq\alpha$
iff $ \theta_1 \P_*(\eventM{\phi_1}) + \cdots + \theta_n
\P_*(\eventM{\phi_n})\geq\alpha$.
\end{itemize}

A \emph{belief structure} has the form $M = (\Worlds, \Bel, \pi)$,
where
$\Bel$ is a belief function.  We can interpret likelihood formulas with
respect to belief structures in the obvious way. 
Similarly, a \emph{possibility structure} has the form $M = (\Worlds,
\Poss, \pi)$, where 
$\Poss$ is a possibility measure. Again, we
interpret likelihood formulas with respect to possibility structures
in the obvious way. 

Let $\MProb$, $\MLP$, $\MBel$, and $\MPoss$ denote 
the set of all probability structures, lower probability structures, 
belief structures, and possibility structures, respectively.

\subsection{Reasoning about Expectation}

Our logic for reasoning about expectation is similar in spirit to $\LPP$.
The idea is to interpret a propositional formula $\phi$ as the indicator
function $X_{\eventM{\phi}}$, which is 1 in worlds where $\phi$ is
true, and 0 otherwise.  
We can then take linear combinations of such gambles.  
Formally, we again start with a set $\Phi_0$ of primitive propositions.
A \emph{(linear) propositional gamble} has the form 
$\alpha_1\phi_1 + \cdots + \alpha_n\phi_n$,
where $\alpha_1,\ldots,\alpha_n$ are integers and $\phi_1, \ldots,
\phi_n$ are propositional formulas that mention only the primitive
propositions in $\Phi_0$.
\newcommand{\gam}{\gamma}
We use $\gam$ to represent propositional gambles. 
An \emph{expectation inequality} is a statement of the form $\theta_1
e(\gam_1) + \cdots + \theta_k e(\gam_k)\geq\alpha$, where
$\theta_1\ldots,\theta_k$ are integers, $k\geq 1$, and $\alpha$ is 
an integer. An \emph{expectation formula} is 
a Boolean combination of 
expectation inequalities.  We use $f$ and 
$g$ to represent expectation formulas.  
We define $\le$, $>$, $<$, and $=$ just as 
in Section~\ref{s:likelihood}, 
Let $\LE$ be the language consisting of expectation formulas.
Given a model $M$, we associate with 
a propositional gamble $\gam$
the gamble $\gambM{\gam}$, where
$\gambM{b_1\phi_1+\cdots+b_n\phi_n}=b_1 X_{\eventM{\phi_1}} + \cdots
+ b_n X_{\eventM{\phi_n}}$.  
Of course, the intention is to interpret $e(\gam)$ 
in $M$
as the expected value
of the gamble $\gambM{\gam}$, 
where the notion of ``expected value'' depends on the underlying
semantics.  In the case of probability structures, it is 
probabilistic expectation; in the case of belief structures, it is
expected belief; in the case of lower probability structures, it is
lower expectation; and so on.
For example, if $M \in \MProb$, then 
\begin{itemize}
\item[] $M \models \theta_1 e(\gam_1) + \cdots + \theta_k
     e(\gam_k)\geq\alpha$ iff $ \theta_1 E_\mu(\gambM{\gam_1}) +
\cdots + \theta_k E_\mu(\gambM{\gam_k})\geq\alpha$.
\end{itemize}

Again, Boolean combinations are defined in the obvious way.  We leave
the obvious semantic definitions in the case of belief structures and
lower probability structures to the reader.

\section{Expressive Power}\label{sec:expressive}

It is easy to see that $\LE$ is at least as expressive as $\LPP$.
Since the expected value of an indicator function is its likelihood, for
all the notions of likelihood we are considering, replacing all
occurrences of $\mi(\phi)$ in a formula in $\LPP$ by $e(\phi)$ gives an
equivalent formula in $\LE$.  Is $\LE$ strictly more expressive than
$\LPP$?  That depends on the underlying semantics.

In the case of probability, it is easy to see that it is not.  Using 
additivity and affine homogeneity, it is easy to take an arbitrary
formula $f \in \LE$ and find a formula $f' \in \LE$ that is equivalent to $f$
(with respect to structures in $\MProb$) such that $e$ is applied only
to propositional formulas.  Then using the equivalence of $e(\phi)$
and $\mi(\phi)$, we can find a formula $f^T \in \LPP$ equivalent to $f$
with respect to structures in $\MProb$.
It should be clear that the translation $f$ to $f^T$ causes at most a
linear blowup in the size of the formula. 

The same is true if we interpret formulas with respect to $\MBel$
and $\MPoss$.  In both cases, given a formula $f \in \LE$, 
we can use (\ref{belexprop2}) to find a formula $f' \in \LE$
equivalent to $f$ such that $e$ is applied only to propositional
formulas (see Lemma~\ref{l:transf2} in the appendix).
It is then easy to find a formula $f^T \in \LPP$ equivalent to $f'$ with
respect to structures in $\MBel$ 
and $\MPoss$.
However, now
the translation from $f$ to $f^T$
can cause an exponential blowup in the size of the formula;
we do not know if there is a shorter translation.

What about lower expectation/probability?  In this case, $\LE$ is
strictly more expressive than $\LPP$.  It is 
not hard to
construct two structures in $\MLP$ that agree on all formulas in $\LPP$ but
disagree on formulas in $\LE$ such as $e(p + q) > 1/2$.  That means
that there cannot be a formula in $\LPP$ equivalent to $e(p+q) > 1/2$.

The following theorem summarizes this discussion.

\thm\label{thm:expressive} $\LE$ and $\LPP$ are equivalent in expressive
power with respect to 
$\MProb$, $\MBel$, and $\MPoss$.  $\LE$ is strictly more expressive than $\LPP$
with respect to $\MLP$.
\ethm

\prf See Appendix~\ref{appa4}. \eprf

\section{Axiomatizing Expectation}\label{sec:axiomatization}

In FHM, a sound and complete axiomatization is provided for
$\LPP$ both with respect to $\MProb$ and $\MBel$; in 
HP,
a sound and complete axiomatization is provided for $\LPP$ with respect
to $\MLP$.  Here we provide a sound and complete axiomatization for
$\LE$ with respect to these structures, as well as with respect to
$\MPoss$. 

The axiomatization for $\LPP$ given in FHM splits into three parts, dealing
respectively with propositional reasoning, reasoning about linear
inequalities, and reasoning about likelihood.  We follow the same
pattern here.  The following axioms
characterize propositional reasoning: 
\begin{axiomlist}
\item[\axiom{Taut}.] All instances of propositional tautologies
in the language $\LE$.
\item[\axiom{MP}.] From $f$ and $f\rimp g$ infer $g$.
\end{axiomlist}
Instances of \axiom{Taut} include all formulas of the form $f\lor\neg
f$, where $f$ is an expectation formula.  We could replace \axiom{Taut}
by a simple collection of axioms that characterize propositional
reasoning (see, for example, \citeN{Men}), but we have chosen to focus
on aspects of reasoning about expectations.

The following axiom characterizes reasoning about linear inequalities:
\begin{axiomlist}
\item[\axiom{Ineq}.] All instances 
in $\LE$ 
of valid formulas about linear inequalities.
\end{axiomlist}
This axiom is taken from FHM.  There, 
an inequality formula is taken to be a 
Boolean combination of 
formulas of the form $a_1
x_1 + \dots + a_n x_n \geq c$, over variables $x_1,\ldots,x_n$.  
Such a formula is valid
if the resulting inequality holds 
under every possible assignment of real
numbers to variables. To get an instance of \axiom{Ineq}, we 
replace each variable $x_i$ that occurs in a valid formula about
linear inequalities by a primitive expectation term of the form
$e(\gam_i)$ (naturally each occurrence of the variable $x_i$ must be
replaced by the same primitive 
expectation
term $e(\gam_i)$). As with \axiom{Taut}, we can replace \axiom{Ineq}
by a sound and complete axiomatization for Boolean combinations of
linear inequalities. One such axiomatization is given in
FHM.
It is described in Section~\ref{s:gambleineq}; the details do not matter
for the discussion in this section.

The following axioms characterize probabilistic expectation in terms of
the properties described in Proposition~\ref{p:probexp-props}.
\begin{axiomlist}
\item[\axiom{E1}.] $e(\gam_1+\gam_2)=e(\gam_1)+e(\gam_2)$,
\item[\axiom{E2}.] $e(a\phi)=a e(\phi)$ for all $a\in\R$,
\item[\axiom{E3}.] $e(\falsep)=0$,
\item[\axiom{E4}.] $e(\truep)=1$,
\item[\axiom{E5}.] $e(\gam_1)\leq e(\gam_2)$ if $\gam_1\leq\gam_2$ is
an instance of 
a valid formula about propositional gamble inequality (see below).
\end{axiomlist}
Axiom \axiom{E1} is simply additivity of expectations. Axioms
\axiom{E2}, \axiom{E3}, and \axiom{E4}, in conjunction with
additivity, capture affine homogeneity. Axiom \axiom{E5} captures
monotonicity.  
A propositional gamble inequality
is a formula of the form $\gam_1\leq\gam_2$, where $\gam_1$
and $\gam_2$ are propositional gambles. 
Examples of valid propositional gamble inequalities are $p = p\land q +
p \land \neg q$,  $\phi \le \phi + \psi$, and $\phi \le \phi \lor \psi$.  
We define the semantics of gamble inequalities more carefully in
Section~\ref{s:gambleineq}, where we provide a complete axiomatization
for them.
As in the case of \axiom{Ineq},
we can replace \axiom{E5} by a sound and complete axiomatization for Boolean
combinations of gamble inequalities.%
\footnote{We could have taken a more 
complex language that contains both expectation formulas and gamble
inequalities. We could then merge the axiomatizations for expectation
formulas and gamble inequalities.
For simplicity, and to clarify the relationship between reasoning 
about expectation versus reasoning about likelihood (see
Section~\ref{sec:expressive}), we consider only the restricted
language in this paper.}

Let $\AXProb$ be the axiomatization $\{\axiom{Taut}, \axiom{MP},
\axiom{Ineq}, \axiom{E1}, \axiom{E2}, \axiom{E3}, \axiom{E4},
\axiom{E5}\}$.  
As usual, 
given an axiom system $\ax$,
we say that a formula $f$ is \emph{$\ax$-provable}
if it can be proved 
in finitely many steps
using the
axioms and rules of inferences of $\ax$.
$\ax$ is sound with respect to a class $\M$ of structures 
if every $\ax$-provable formula is valid in $\M$.  
$\ax$ is \emph{complete} with respect to $\M$ if every formula that
is valid in $\M$ is $\ax$-provable.

\thm\label{t:soundcomplprob}
$\AXProb$ is a sound and complete axiomatization 
of $\LE$ with respect to $\MProb$.
\ethm

\prf See Appendix~\ref{appa5}. \eprf

Despite the fact that we allow structures that have infinitely many
worlds, we do not have axioms capturing the continuity properties of
expectations that hold for expectations over infinite
spaces. 
Roughly speaking, this is because the logic cannot distinguish between
infinite sample spaces and finite ones. 
We make this intuition precise in Section~\ref{s:decision}. 

The characterizations of Theorems~\ref{t:lpexp-char} and~\ref{t:belexp-char}
suggest the appropriate axioms for reasoning about 
lower
expectations and expected beliefs.
The following axioms capture the properties specified in
Proposition~\ref{p:lpexp-props}:
\begin{axiomlist}
\item[\axiom{E6}.] $e(\gam_1+\gam_2)\geq e(\gam_1)+e(\gam_2)$,
\item[\axiom{E7}.] $e(a\gam+b~\truep)=a e(\gam) + b$,  for all $a,b\in\R$,
$a\geq 0$,
\item[\axiom{E8}.] $e(a\gam+b~\falsep)=a e(\gam)$, for all $a,b\in\R$,
$a\geq 0$.
\end{axiomlist}
Axiom \axiom{E6} is superadditivity of the expectation. Axioms
\axiom{E7} and \axiom{E8} capture positive affine homogeneity. Note
that because we do not have additivity, we cannot get away with
simpler axioms as in the case of probability. Monotonicity is
captured, as in the case of probability measures, by axiom
\axiom{E5}.  Let $\AXLP$ be the axiomatization $\{\axiom{Taut},
\axiom{MP},
\axiom{Ineq},\axiom{E5},\axiom{E6},\axiom{E7},\axiom{E8}\}$. 

\thm\label{t:soundcompllp}
$\AXLP$ is a sound and complete axiomatization 
of $\LE$ with respect to $\MLP$.
\ethm

\prf See Appendix~\ref{appa5}. \eprf

Although it would seem that Theorem~\ref{t:soundcompllp} should follow
easily from Proposition~\ref{p:lpexp-props}, this is, unfortunately, not the
case.  As usual, soundness is straightforward, and to prove completeness,
it suffices to show that if a formula $f$ is consistent with $\AXLP$, it is
satisfiable in a structure in $\MLP$.  Indeed, it suffices to consider 
formulas $f$ that are conjunctions of expectation inequalities and their
negations.  However, the usual approach for proving completeness in modal
logic, which involves considering maximal consistent sets and canonical
structures does not work.  The problem is that there are maximal
consistent sets of formulas that are not satisfiable.  For example,
there  is a maximal consistent set of formulas that includes
$e(\gamma) > 0$ and $e(\gamma) \le 1/n$ for $n = 1, 2, \ldots$; this
is clearly unsatisfiable.  A similar problem arises in 
the completeness proofs for $\LPP$ given in FHM and HP, but the
techniques used there do 
not seem to suffice for dealing with expectations.

Of course, it is the case that any expectation function that satisfies
the constraints in the formula $f$ and also every instance of
axioms \axiom{E6}, \axiom{E7}, and \axiom{E8} must be a lower
expectation, by Theorem~\ref{t:lpexp-char}.  The problem is that,
\emph{a priori}, there are infinitely many relevant instances of the
axioms.  To get completeness, we must reduce this to a finite number of
instances of  these axioms.  It turns out that this can be done, 
using techniques from linear programming and Walley's \citeyear{Walley91}
notion of \emph{natural extension}.  

It is also worth noting that, although $\LE$ is a more expressive
language than $\LPP$ in the case of lower probability/expectation, the
axiomatization for $\LE$ in this case is much more elegant than the
corresponding axiomatization for $\LPP$ given in 
HP.

We next consider
expectation with respect to belief.
As expected, 
the axioms capturing the interpretation of belief expectation rely on
the properties pointed out in 
Proposition~\ref{p:belexp-props}. Stating these properties in the logic
requires a way to express the max and min of two propositional
gambles. It turns out that we can view the notation $\gam_1\lor\gam_2$ 
as an abbreviation for a more complex expression.
Given a propositional gamble $\gam = b_1 \phi_1 + \cdots + b_n \phi_n$,
we construct an equivalent gamble $\gam'$ as follows.
First define a family $\rho_A$ of propositional formulas indexed by
$A\subseteq\{1,\ldots,n\}$ 
by taking $\rho_A = \bigwedge_{i\in A}\phi_i\land
(\bigwedge_{j \notin A} \neg \phi_j)$.  Thus, $\rho_A$ is true 
exactly if the $\phi_i$'s for $i \in A$ are true, and the other $\phi_j$'s
are false.  Note that the formulas $\rho_A$ are mutually exclusive.
Define $\alpha_A$ for $A \subseteq \{1, \ldots, n\}$ 
by taking $\alpha_A=\sum_{i\in A}\alpha_i$. 
Define $\gam' = \sum_{A\subseteq\{1,\ldots,n\}}\alpha_A\rho_A$. 
It is easy to check that the propositional gambles $\gam$ and $\gam'$
are equal. 
Given two propositional gambles, say $\gam_1$ and $\gam_2$, we can
assume without loss of generality that the involve the same primitive
propositions $\phi_1, \ldots, \phi_n$.  (If not, we can always add
``dummy'' terms of the form $0 \psi$.)  Form the gambles $\gam_1'$ and
$\gam_2'$ as above.  Since all the formulas mentioned in $\gam_1'$ and
$\gam_2'$ are mutually exclusive, 
it follows that $\max(\gam_1',\gam_2') = \sum_{A \subseteq \{1, \ldots,
n\}} \max(b_A,b_A') \rho_A$.   
We take $\gam_1\lor\gam_2$ to be an abbreviation for this gamble. 
(Note that if $\gam_1$ and $\gam_2$ are propositional formulas, then
$\gam_1 \lor \gam_2$ really
is a gamble equivalent to the propositional formula $\gam_1 \lor \gam_2$, 
given our identification of propositional formulas with indicator functions,
so the use of $\lor$ is justified here.)  Of course,
we can similarly define $\gam_1\land\gam_2$, simply by 
taking $\min$ instead of $\max$. 

With these definitions, the following axiom accounts for
property (\ref{inex}):
\begin{axiomlist}
\item[\axiom{E9}.] $e(\gam_1\lor\cdots\lor\gam_n) \ge
 \sum_{i=1}^{n}\sum_{\{I\subseteq\{1,\ldots,n\}:|I|=i\}}(-1)^{i+1} 
e(\bigwedge_{j\in I}\gam_j)$.
\end{axiomlist}
To deal with the comonotonic additivity property (\ref{belexprop2}),
it seems that comonotonicity must be expresssed in the logic. It turns
out that it suffices to capture only a restricted form of
comonotonicity. Note that if $\phi_1,\ldots,\phi_m$ are pairwise
mutually exclusive, $a_1\leq\ldots\leq a_m$, 
$b_1\leq\ldots \leq b_m$, $\gam_1=a_1\phi_1+\cdots+a_m\phi_m$, and
$\gam_2=b_1\phi_1+\cdots+b_m\phi_m$, then in all structures $M$, the
gambles $\gambM{\gam_1}$ and $\gambM{\gam_2}$ are comonotonic. Thus,
by (\ref{belexprop2}), it follows that
$E_{\Bel}(\gambM{\gam_1+\gam_2})=E_{\Bel}(\gambM{\gam_1})+E_{\Bel}(\gambM{\gam_2})$.
The proof that $E_{\Bel}$ satisfies comonotonic additivity (see the
proof of Theorem~\ref{t:belexp-char} in the appendix) shows that it
suffices to restrict to gambles of this form. These observations lead
to the following axiom:
\begin{axiomlist}
\item[\axiom{E10}.] $e(\gam_1+\gam_2)=e(\gam_1)+e(\gam_2)$ if
$\gam_1=a_1\phi_1+\cdots+a_m\phi_m$,
$\gam_2=b_1\phi_1+\cdots+b_m\phi_m$, $a_1\leq\ldots\leq a_m$,
$b_1\leq\ldots\leq b_m$, and $\neg(\phi_i\land\phi_j)$ is a
propositional tautology for all $i\neq j$.
\end{axiomlist}
Let $\AXBel$ be the axiomatization $\{\axiom{Taut},
\axiom{MP}, \axiom{Ineq}, \axiom{E5},\axiom{E7},\axiom{E8},
\axiom{E9}, \axiom{E10}\}$. 

\thm\label{t:soundcomplbel}
$\AXBel$ is a sound and complete axiomatization 
of $\LE$ with respect to $\MBel$.
\ethm

\prf See Appendix~\ref{appa5}. \eprf

Finally, we consider expectation with respect to possibility. The axioms
capturing 
the interpretation of possibilistic expectation $E_{\Poss}$ rely on
the properties pointed out in Proposition~\ref{p:possexp-props}.
The max property (\ref{maxprop}) is 
captured by the following axiom:
\begin{axiomlist}
 \item[\axiom{E11}.]
$(e(\phi_1)\geq e(\phi_2))\rimp(e(\phi_1\lor\phi_2)=e(\phi_1))$. 
\end{axiomlist}
\axiom{E11} 
essentially says that $e(\phi_1 \lor \phi_2) =
\max(e(\phi_1),e(\phi_2))$.

Let $\AXPoss$ be the axiomatization $\{\axiom{Taut}, 
\axiom{MP}, \axiom{Ineq}, \axiom{E5},\axiom{E7},\axiom{E8}, \axiom{E10},
\axiom{E11}\}$.

\thm\label{t:soundcomplposs}
$\AXPoss$ is a sound and complete axiomatization of $\LE$ with
respect to $\MPoss$.
\ethm

\prf See Appendix~\ref{appa5}. \eprf

\section{Decision Procedures}\label{s:decision}

In FHM, it was shown that the satisfiability problem for
$\LPP$ is NP-complete, both with respect to $\MProb$ and $\MBel$; in
HP, NP-completeness was also shown with respect to
$\MLP$.  
Here we prove similar results for the language $\LE$.  In the
case of $\MProb$, this is not at all surprising, given
Theorem~\ref{thm:expressive} and the fact that the translation from
$\LE$ to $\LPP$ causes only a linear blowup in the case of $\MProb$.
However, we cannot get the result for $\MBel$ or $\MPoss$ from
Theorem~\ref{thm:expressive}, since the translation causes an
exponential blowup.  Of course, in the case of $\MLP$, no translation
exists at all.  Nevertheless, in all these cases, we can get
NP-completeness using techniques very much in the spirit of the linear
programming techniques used in FHM.

We define $|f|$ to be the
length of $f$, that is, the number of symbols required to write $f$,
where each coefficient is counted as one symbol. Define $||f||$ to be
the length of the longest coefficient appearing in $f$, when written
in binary. The size of a rational number $\frac{a}{b}$, denoted
$||\frac{a}{b}||$, where $a$ and $b$ are relatively prime, is defined
to be $||a||+||b||$. 

The next theorems, helpful to establish the complexity of the decision 
procedures, show that the logic is not expressive enough to distinguish
between infinite sample spaces and infinite ones;
if a formula is satisfiable at all, it is satisfiable in a
(relatively small) finite structure.
In other words, we could have restricted to structures with only
finitely many worlds
without loss of generality.
We prove this result for $\MLP$, $\MBel$, and $\MPoss$; it follows from
\citeN[Theorem~2.4]{FHM} (which shows that a formula in the language
$\LPP$ is satisfiable in $\MProb$ iff it is satisfiable in a finite
structure in $\MProb$) and Theorem~\ref{thm:expressive} (which shows that
every formula in $\LE$ is equivalent to a formula in $\LPP$ for
structures in $\MProb$) that the result holds for $\MProb$ as well.

\thm
\label{t:smallmodel1}
Suppose that $f \in \LE$ is satisfied in some structure in $\MLP$.  Then
$f$ is satisfied in a structure $(\Worlds,\P,\pi)$ such that  $|\Worlds|
\le |f|^2$, $|\P|\leq|f|$, $\mu(w)$ is a rational number such that
$||\mu(w)||$ is $O(|f|^2||f||+|f|^2\log(|f|))$ for every world $w \in
\Worlds$ and $\mu \in \P$, and $\pi(w)(p) = \false$ for every world $w
\in \Worlds$ and every primitive proposition $p$ not appearing in $f$.
\ethm

\prf See Appendix~\ref{appa6}. \eprf

To prove a small-model theorem for belief functions (and
possibility measures), we need a representation of belief functions
that is particularly compact. Given a 
set $W$ of worlds, a \emph{mass
function} $m$ is a function from $2^W$ to $[0,1]$ such that
$m(\emptyset)=0$, and $\sum_{U\subseteq W} m(U)=1$
\cite{Shaf}. 
As is well known, in finite spaces, 
there is a one-to-one correspondence between belief
functions and mass functions \cite{Shaf}.
Given a mass function $m$, let $\Bel_m$ be the set function defined by
setting $\Bel_m(U)=\sum_{V\subseteq U} m(V)$. It can be shown that
$\Bel_m$ is a belief function (in particular, it satisfies B3). 
Conversely, given a belief
function $\Bel$, consider the set function $m_\Bel$ defined by
setting $m_\Bel(\emptyset)=0$, and 
$m_\Bel(U)=\Bel(U)-\sum_{V\subset U} m_\Bel(V)$. 
It can be shown that $m_\Bel$ is a mass function (in particular, 
$m_\Bel(U) \ge 0$ for all $U$)
and that $\Bel_{m_\Bel} =
\Bel$.  It easily follows that the mappings $\Bel \mapsto m_{\Bel}$
and $m \mapsto \Bel_m$ give a one-to-one correspondence between
belief function and mass functions.
\thm\label{t:smallmodel2}
Suppose that $f \in \LE$ is satisfied in some structure in $\MBel$
(resp., $\MPoss$). Then $f$ is satisfied in a structure
$(\Worlds,\nu,\pi)$ such that $|\Worlds|\le |f|^2$, $\nu$ is a belief
function (resp., possibility measure) whose corresponding mass
function is positive on at most $|f|$ subsets of $\Worlds$ and the mass
of each of these $|f|$ sets is a rational number of size 
$O(|f|\,||f||+|f|\log(|f|))$, 
and $\pi(w)(p) = \false$ for every world $w \in \Worlds$ and every
primitive proposition $p$ not appearing in $f$.  
\ethm

\prf See Appendix~\ref{appa6}. \eprf

Theorems~\ref{t:smallmodel1} and \ref{t:smallmodel2} yield the
following complexity result for the decision problem for $\LE$.

\thm\label{t:decproc3}
The problem of deciding whether a formula in $\LE$ is
satisfiable 
in $\MProb$ (\respc $\MLP$, $\MBel$, $\MPoss$) is NP-complete.
\ethm

\prf See Appendix~\ref{appa6}. \eprf

\section{Reasoning about Gamble Inequalities}\label{s:gambleineq}
\axiom{Ineq} presumes an oracle for all valid formulas about linear
inequalities; as we said earlier, this oracle can be replaced by the
complete axiomatization of linear inequalities provided in FHM.
Similarly, \axiom{E5} assumes an oracle for valid formulas about gamble
inequalities.  In this section, we provide an axiomatization that can
replace that oracle.  One of the axioms involves reasoning about linear
inequalities over real-valued functions, so we axiomatize this as well.

Let $\Lg$ consist of all Boolean combinations of gamble inequalities
$\gam \ge \tilde{c}$, where $\gam$ is a propositional gamble, as defined in
Section~\ref{s:logics}, and $c$ is an integer.
As in Section~\ref{s:likelihood}, we write $\gam_1 \le \gam_2$ as
an abbreviation for $\gam_1 - \gam_2 \le \tilde{0}$, which is the form
of gamble inequality used in axiom~\axiom{E5}.  
We also use the abbreviations we defined in Section~\ref{s:likelihood} for
$\le$, $<$, and $=$.  However, we now take $\gam_1 > \gam_2$ to be an
abbreviation for $\gam_1 \ge \gam_2 \land \neg(\gam_1 \le \gam_2)$.  The
analogue of the conjunct $\gam_1 \ge \gam_2$ is not necessary when
reasoning about likelihood or expectation, since likelihood and
expectation terms are interpreted as real numbers, and thus are
totally ordered.  (For real numbers $\alpha_1$ and $\alpha_2$,
$\neg(\alpha_1 \le \alpha_2)$ implies $\alpha_1 > \alpha_2)$.
However, this is not the case for gamble expressions, which are
interpreted as real-valued functions.

We can provide a semantics for gamble formulas by considering structures
$M=(W,\pi)$ where $W$ is a 
nonempty 
set of worlds and $\pi$ associates with each world in $W$ a truth
assignment on the primitive propositions. 
Let $\Mg$ be the class all such structures. (Clearly, 
every structure in $\MProb$, $\MLP$, $\MBel$, and $\MPoss$ can be
interpreted as a structure in $\Mg$
by simply ``forgetting'' the uncertainty measure over the worlds.
Note that $W$ is nonempty for such structures, since
$\mu(W)\ne\mu(\emptyset)$ for all the measures of uncertainty considered.)
Then
\[
  M\models \gam \geq \tilde{c} \mbox{~~iff~~}
  \mbox{for all $w\in W$,~}
     \gambM{\gam}(w)\geq c.
\]
Again, Boolean combinations are given semantics in the obvious way.

We can characterize gamble formulas axiomatically as follows.
As before, we have \axiom{Taut} and \axiom{MP} (although
now we consider the instances of valid propositional tautologies in
the language $\Lg$). Instead of \axiom{Ineq}, we have a similar axiom
that captures reasoning about linear inequalities interpreted over
\emph{real-valued functions} with nonempty domains:
\begin{axiomlist}
\item[\axiom{IneqF.}] All instances in $\L^g$ of valid formulas
about linear inequalities over real-valued functions with nonempty
domains. 
\end{axiomlist}
A formula about linear inequalities over real-valued functions 
is a Boolean
combination of formulas of the form $a_1 v_1 + \cdots a_n v_n\ge 
\tilde{c}$, where $v_1, \ldots, v_n$ are variables ranging over
real-valued functions.  This formula is valid if, for every domain $X$
and every assignment for real-valued functions with domain $X$ to $v_1,
\ldots, v_n$, the resulting inequality  holds when $\tilde{c}$ is
interpreted as the constant function with domain $X$ that always returns
$c \in \R$ and $\ge$ is the pointwise ordering on functions.
To get an instance
of \axiom{IneqF}, we replace each function variable $v_i$ that
occurs in a valid formula about linear inequalities by a propositional
formula $\phi_i$. (Again, each occurence of the 
function variable
$v_i$ must be replaced by the same propositional formula $\phi_i$.) As
with \axiom{Ineq}, we can replace \axiom{IneqF} by a sound and
complete axiomatization for Boolean combinations of linear
inequalities over real-valued functions. 
We provide one such
axiomatization at the end of this section.\footnote{In a preliminary
version of this paper
\cite{HalPuc02:UAI}, we mistakenly used an axiomatization that relied on
\axiom{Ineq}. \axiom{Ineq} is inappropriate for reasoning about linear 
inequalities over real-valued functions because, intuitively,
\axiom{Ineq} captures properties of inequalities where $\ge$ is a
linear order. However, the pointwise ordering on functions is only a
partial order. Furthermore, we did not have axioms \axiom{G2} and \axiom{G3}
in that version of the paper; they seem to be necessary.} We also
consider the following axioms, 
which capture properties of indicator functions, such as: for any
$w\in W$, $X_U(w)+X_V(w) = X_{U\cup V}(w)$ if $U\cap V=\emptyset$, and
$X_U(w)\leq X_V(w)$ if $U\subseteq V$.
\begin{axiomlist}
\item[\axiom{G1}.] $a((\phi \land \psi) \lor(\phi \land \neg
\psi))=a(\phi \land \psi) +a(\phi \land \neg \psi)$, 
\item[\axiom{G2}.] $a(\phi \land \psi) +b(\phi \land \neg \psi)
\ge\tilde{0}\riff a(\phi \land \psi)\ge\tilde{0}\land b(\phi \land \neg
\psi)\ge\tilde{0}$  
\item[\axiom{G3}.] $\truep=\tilde{1}$,
\item[\axiom{G4}.] $\phi - \psi = \tilde{0}$ if $\phi \dimp \psi$ is a
propositional tautology.
\end{axiomlist}
Intuitively, \axiom{G1} and \axiom{G3} say that propositional
formulas act like indicator functions, and \axiom{G2} says that the
$\ge$ order is the pointwise order on gambles. Axiom \axiom{G4} simply
says that equivalent formulas yield the same indicator functions. 
Let $\AXg$ be the axiomatization $\{\axiom{Taut}, \axiom{MP},
\axiom{IneqF},\axiom{G1},\axiom{G2},\axiom{G3},\axiom{G4}\}$.
\thm\label{t:gamblecompl}
$\AXg$ is a sound and complete axiomatization 
of $\Lg$ with respect to $\Mg$.
\ethm

\prf See Appendix~\ref{appa7}. \eprf

Finally, we consider the complexity of the decision procedure for $\Lg$. As 
we did in Section~\ref{s:decision}, let $\Lg_1$ be the restriction of
$\Lg$ to inequalities with integer coefficients. 
Perhaps not surprisingly, the logic for reasoning about gamble
inequalities is NP-complete.
\thm\label{t:decprocLg} The problem of deciding whether a formula of
$\Lg_1$ is satisfiable in $\Mg$ is NP-complete.
\ethm

\prf See Appendix~\ref{appa7}. \eprf

We now provide a sound and complete axiomatization for reasoning about 
linear inequalities over real-valued functions. 
As before, start with \axiom{Taut} and \axiom{MP}, and consider the
following axioms:
\begin{axiomlist}
\item[\axiom{F1.}] $v-v\ge \tilde{0}$,
\item[\axiom{F2.}] $(a_1 v_1 + \cdots + a_k v_k\ge \tilde{c})\riff (a_1v_1 +
\cdots + a_kv_k + 0v_{k+1}\ge \tilde{c})$,
\item[\axiom{F3.}] $(a_1 v_1 + \cdots + a_k v_k\ge \tilde{c})\riff (a_{j_1}
x_{j_1} + \cdots + a_{j_k} v_{j_k}\ge \tilde{c})$, if $j_1,\ldots,j_k$ is a
permutation of $1,\ldots,k$,
\item[\axiom{F4.}] $(a_1v_1 + \cdots + a_kv_k\ge
\tilde{c})\land(a_1'v_1+ \cdots+a_k'v_k\ge
\tilde{c}')$\\
$\rimp (a_1+a_1')v_1+ \cdots+(a_k+a_k')v_k\ge (\widetilde{c+c'})$,
\item[\axiom{F5.}] $(a_1v_1 + \cdots  +a_kv_k\ge \tilde{c}) \riff (da_1v_1 +
\cdots + da_kv_k\ge \widetilde{dc})$, if $d>0$,
\item[\axiom{F6.}] $(a_1v_1 + \cdots + a_kv_k\ge \tilde{c})\rimp
(a_1v_1+ \cdots+a_kv_k>\tilde{d})$, if $c>d$.
\end{axiomlist}
The axiomatization for Ineq for reasoning about linear inequalities over
\emph{reals} presented in FHM is identical to this axiomatization,
except that it includes one additional axiom: $x \ge c \lor x
\le c$.  (Note that now $c$ represents a real number, as opposed to
$\tilde{c}$, which represents a constant real-valued function.)  The
absence of this axiom is not surprising;
although $\ge$ is a total order over the reals, 
it is only a partial order over real-valued functions. 

Let $\AXf$ be the axiomatization
$\{\axiom{Taut},\axiom{MP},\axiom{F1},\axiom{F2},\axiom{F3},\axiom{F4},\axiom{F5},\axiom{F6}\}$.
\thm\label{t:ineqfuncompl}
$\AXf$ is sound and complete for reasoning about formulas about linear 
inequalities over real-valued functions with nonempty domain.
\ethm

\prf See Appendix~\ref{appa7}. \eprf

We can now replace axiom \axiom{IneqF} in $\AXg$ by 
\axiom{F1}--\axiom{F6} 
(since $\AXg$ already has \axiom{Taut} and
\axiom{MP}). Following the discussion above, this means that we
replace the $v$'s appearing in 
\axiom{F1}--\axiom{F6} 
by propositional formulas $\phi$'s. Note that axiom \axiom{F1} becomes
redundant in 
$\AXg$, since it becomes $\falsep\ge\falsep$, which follows from
\axiom{G2}. 

The complexity of reasoning about linear inequalities over real-valued 
functions (with integer coefficients, following what was said in
Section~\ref{s:decision}) is NP-complete, just like reasoning about linear
inequalities over the reals.
\thm\label{t:decprocLineqf} The problem of deciding whether a formula
about linear inequalities (with integer coefficients) 
is satisfiable
over real-valued
functions with nonempty domain is NP-complete.
\ethm

\prf See Appendix~\ref{appa7}. \eprf

\section{Conclusion}
The notion of expectation is critical in many areas, ranging from
physics to decision theory (where much of modern theory of decision
making under risk is based on the notion of maximizing expected
utility \cite{Savage}), and theoretical computer science (where it
arises, for example, when
establishing the correctness and efficiency of randomized algorithms
\cite{Motwani95}).   For the most part, expectation has been considered
in the context of probability.  However, it has been considered for
other representations of uncertainty.   For example, various notions of
decision making are based on taking expectations with respect to other
representations of uncertainty \cite{Schmeidler86,Walley91}.  Indeed, it
is possible to consider almost all decision rules as instances of a
generalized expected utility, for different representations of
uncertainty \cite{CH03a,CH03b}. 
After reviewing characterizations of expectation functions with respect
to various notions of uncertainty (and, in a few cases, providing
characterizations when none existed), we considered a propositional
logic for reasoning about uncertainty.  Such a logic should prove useful
to reason about decision making with respect to various notions of
uncertainty.  
Our logic is at least as expressive as the corresponding logic for
reasoning about uncertainty for all representations we considered, and
in some cases it is more expressive.
For each
representation of uncertainty, we provided a sound and complete
axiomatization for the logic, and showed that the satisfiability
problem is NP-complete, as it is for reasoning about probabilibity,
lower probabilities, and belief functions. Thus, the added
expressiveness of being able to reason about expectation of sets of
probability measures comes at no computational-complexity cost. 
Moreover, the axiomatization is, in some cases, much more elegant.

\appendix

\section{Proofs}

\subsection{Proofs for Section~\protect{\ref{s:exp}}}\label{appa2}

\commentout{
\opro{p:probexp-props}
The function $E_\mu$ has the following properties for all measurable
gambles $X$ and $Y$.
\begin{itemize}
\item[(a)] $E_\mu$ is {\em additive\/}: $E_\mu(X + Y) = E_\mu(X) + E_\mu(Y)$.
\item[(b)] $E_\mu$ is {\em affinely homogeneous\/}: $E_\mu(a
X + \tilde{b}) =  a E_\mu(X) + b$ for all
$a, b \in \R$.
\item[(c)] $E_\mu$ is {\em monotone}: if $X \le Y$, then $E_\mu(X)
\le E_\mu(Y)$.
\end{itemize}
\eopro
\prf
The proof is well known and straightforward; we give it here for
completeness. 

(a) Let $X,Y$ be gambles, and $\mu$ a probability measure. We
can check that:
\begin{eqnarray*}
E_\mu(X+Y) 
 & = & \sum_{z\in\V(X+Y)} z\mu(X+Y=z)\\
 & = & \sum_{x\in\V(X)}\sum_{y\in\V(Y)} (x+y)\mu(X=x)\mu(Y=y)\\
 & = & (\sum_{x\in\V(X)}\sum_{y\in\V(Y)} x\mu(X=x)\mu(Y=y)) + 
       (\sum_{y\in\V(Y)}\sum_{x\in\V(X)} y\mu(X=x)\mu(Y=y))\\
 & = & (\sum_{x\in\V(X)} (x\mu(X=x)\sum_{y\in\V(Y)} \mu(Y=y))) + 
       (\sum_{y\in\V(Y)} (y\mu(Y=y)\sum_{x\in\V(X)} \mu(X=x)))\\
 & = & (\sum_{x\in\V(X)} x\mu(X=x)) + (\sum_{y\in\V(Y)} y\mu(Y=y))\\
 & = & E_\mu(X)+E_\mu(Y)
\end{eqnarray*}

(b) Let $X$ be a gamble, $\mu$ a probability measure, and
$a,b\in\R$. Consider first the case $a=0$. Clearly,
$E_\mu(aX+\tilde{b})=E_\mu(\tilde{b})=b\mu(\tilde{b}=b)=b$. Consider
next the case $a\not=0$:
\begin{eqnarray*}
E_\mu(aX+\tilde{b}) 
 & = & \sum_{x\in\V(aX+\tilde{b})} x\mu(aX+\tilde{b})\\
 & = & \sum_{z\in\V(X)} (az+b)\mu(aX+\tilde{b}=az+b)\\
 & = & \sum_{z\in\V(X)} (az+b)\mu(X=z)\\
 & = & a(\sum_{z\in\V(X)} z\mu(X=z))+b\\
 & = & a E_\mu(X)+b
\end{eqnarray*}

(c) To show monotonicity, it is sufficient to show that if $X\ge 0$,
then $E_\mu(X)\ge 0$, for a gamble $X$ and a probability measure
$\mu$. If $X\ge 0$, then all $x\in\V(X)$ are such that $x\ge
0$. Therefore, $E_\mu(X)=\sum_{x\in\V(X)}x\mu(X=x)\ge 0$, since
$x\mu(X=x)\ge 0$ for all $x\in\V(X)$. To show that this implies
monotonicity, consider gambles $X,Y$ such that $X\le Y$. Then, $Y-X\ge
0$, and by the previous result, $E_\mu(Y-X)\ge 0$. By additivity and
linearity, $E_\mu(Y-X)=E_\mu(Y)-E_\mu(X)\ge 0$, so that $E_\mu(X)\le
E_\mu(Y)$. 
\eprf
}
\commentout{
\othm{t:probexp-char}
Suppose that $E$ maps gambles measurable with respect to some
algebra $\F$ to $\R$ and $E$ is additive, affinely homogenous, and
monotone. Then there is a (necessarily unique) probability measure
$\mu$ on $\F$ such that $E = E_\mu$. 
\eothm
\prf 
}
\commentout{
\opro{p:lpexp-props}
The functions $\lE_\P$ and $\uE_\P$ have the following properties for
all gambles $X$ and $Y$.
\begin{itemize}
\item[(a)] $\lE_\P(X + Y) \ge
\lE_\P(X) + \lE_\P(Y)$ ({\em superadditivity});\\ 
$\uE_\P(X + Y) \le
\uE_\P(X) + \uE_\P(Y)$ ({\em subadditivity}).
\item[(b)] $\lE_\P$ and $\uE_\P$ are both {\em positively affinely
homogeneous}:
$\lE_\P(a X + \tilde{b}) = a \lE_\P(X) +b$ and
$\uE_\P(a X + \tilde{b}) = a \uE_\P(X) +b$ if $a,b \in \R$, $a \ge 0$.
\item[(c)] $\lE_\P$ and $\uE_\P$ are monotone.
\item[(d)] $\uE_\P(X) = - \lE_\P(-X)$.
\end{itemize}
\eopro
\prf
The proof relies on properties of $\inf$ and $\sup$, that can be found
in any textbook on real analysis.

(a) Let $X,Y$ be gambles, and $\P$ be a set of probability
measures.
$\lE_\P(X+Y) = \inf\{E_\mu(X+Y)~:~\mu\in\P\} =
\inf\{E_\mu(X)+E_\mu(Y)~:~\mu\in\P\} \ge
\inf\{E_\mu(X)~:~\mu\in\P\}+\inf\{E_\mu(Y)~:~\mu\in\P\} =
\lE_\P(X)+\lE_\P(Y)$. Similarly, $\uE_\P(X+Y) =
\sup\{E_\mu(X+Y)~:~\mu\in\P\} = \sup\{E_\mu(X)+E_\mu(Y)~:~\mu\in\P\}
\le \sup\{E_\mu(X)~:~\mu\in\P\}+\sup\{E_\mu(Y)~:~\mu\in\P\} =
\uE_\P(X)+\lE_\P(Y)$. 

(b) Let $X$ be a gamble, $\P$ a set of probability measures, and
$a,b\in\R$, with $a\ge 0$. First, consider the case $a=0$. Clearly,
$\lE_\P(aX+\tilde{b}) = \lE_\P(\tilde{b}) =
\inf\{E_\mu(\tilde{b})~:~\mu\in\P\} = \inf\{b~:~\mu\in\P\}=b$.
If $a>0$, then $\lE_\P(aX+\tilde{b}) =
\inf\{E_\mu(aX+\tilde{b}~:~\mu\in\P\} = \inf\{a
E_\mu(X)+b~:~\mu\in\P\} = a\inf\{E_\mu(X)~:~\mu\in\P\}+b =
a\lE_\P(X)+b$. Similarly for $\uE_\P(aX+\tilde{b})$. 

(c) Let $X,Y$ be gambles such that $X\le Y$. Let $\P$ be a set of
probability measures. By monotonicity of probabilistic expectation,
for any probability measure $\mu\in\P$, $E_\mu(X)\le
E_\mu(Y)$. Therefore, $\inf\{E_\mu(X)~:~\mu\in\P\}\le E_\mu(Y)$ for
all $\mu\in\P$. So, $\inf\{E_\mu(X)~:~\mu\in\P\}$ is a lower bound for
the set $\{E_\mu(Y)~:~\mu\in\P\}$, and hence $\lE_\P(X) =
\inf\{E_\mu(X)~:~\mu\in\P\} \le
\inf\{E_\mu(Y)~:~\mu\in\P\} = \lE_\P(Y)$. Similarly, $E_\mu(X)\le
\sup\{E_\mu(Y)~:~\mu\in\P\}$ for all $\mu\in\P$. So,
$\sup\{E_\mu(Y)~:~\mu\in\P\}$ is an upper bound for the set
$\{E_\mu(X)~:~\mu\in\P\}$, and hence $\uE_\P(X) =
\sup\{E_\mu(X)~:~\mu\in\P\} \le \sup\{E_\mu(Y)~:~\mu\in\P\} =
\uE_\P(Y)$. 

(d) Let $X$ be a gamble, and $\P$ a set of probability
measures. First, we can check that for any probability measure
$\mu\in\P$, $E_\mu(X)=E_\mu(--X)=-E_\mu(-X)$. Thus, we have
$\lE_\P(X)=\inf\{E_\mu(X)~:~\mu\in\P\} = \inf\{-
E_\mu(-X)~:~\mu\in\P\} = -\sup\{E_\mu(-X)~:~\mu\in\P\} =
-\uE_\P(-X)$. 
\eprf
}
\commentout{
\opro{Choqueteq}
$E_\mu(X) = x_1 + (x_2-x_1)\mu(X > x_1) +
\cdots +  (x_n - x_{n-1})\mu(X > x_{n-1}).$
\eopro
\prf
Let $x_1,\ldots,x_n$ be the element of $\V(X)$, ordered from least to
greatest. 
By definition,
$E_\mu(X)=\sum_{i=1}^{n}x_i\mu(X=x_i)$. This can be
written as the telescoping sum 
$$(x_1+\sum_{j=2}^{i}(x_j-x_{j-1}))\mu(x_i) = \sum_{i=1}^{n} (x_1
\mu(X=x_i)+\sum_{j=2}^{i}(x_j-x_{j-1})\mu(X=x_i)).$$ Rearranging terms,
we can write this sum as $$x_1(\sum_{i=1}^{n}\mu(X=x_i)) +
\sum_{i=2}^{n}(x_i-x_{i-1})(\sum_{j=i}^{n}\mu(X=x_j)).$$  But by our
choice of $x_1,\ldots,x_n$, this is equivalent to $$x_1\mu(X\ge
x_1)+\sum_{i=2}^{n}(x_i-x_{i-1})\mu(X\ge x_i) = x_1 +
\sum_{i=2}^{n}(x_i-x_{i-1})\mu(X> x_{i-1}).$$
\eprf
}
Proposition~\ref{Choqueteq} defines the expectation of $X$ in terms of
the values $\{x_1, \ldots, x_n\}$ that make up $\V(X)$.  The following
lemma, which turns out to be useful in a number of technical results,
shows that we can use the same calculation for expectation for any set
of values that includes $\V(X)$.  The lemma
holds for all the
representations of uncertainty $\nu$ considered in this paper. It
uses only the fact that $\nu(W)=1$ and $\nu(\emptyset)=0$ 
for all representations that we are considering.
\lem\label{l:choquetprop}
Let $X$ be a gamble, let $s_1\le\ldots\le s_m$ be reals
such that 
$\V(X)\subseteq\{s_1,\ldots,s_m\}$,
and let $\nu$ be either a probability measure, belief function, or
possibility measure.
Then $E_\nu(X) =
s_1+(s_2-s_1)\nu(X>s_1)+ \cdots+(s_m-s_{m-1})\nu(X>s_{m-1})$. 
\elem
\prf
Clearly, it suffices to show that the result holds for values
such that $s_1<\ldots<s_m$. Indeed, if $s_i=s_{i+1}$ for some $i$,
then clearly the term $(s_{i+1}-s_i)\nu(X>s_i)$ does not contribute to
the sum. For a set $S=\{s_1,\ldots,s_m\}$ such that $S \supseteq \V(X)$
and $s_1<\ldots<s_m$, 
define 
$$E^S_\nu(X) =
s_1+(s_2-s_1)\nu(X>s_1)+ \cdots+(s_m-s_{m-1})\nu(X>s_{m-1}).$$ 
We proceed by induction on $|S - \V(X)|$.  If $|S - \V(X)| = 0$, then $S
= \V(X)$.
Observe that $E_\nu(X)$ is just $E^{\V(X)}_\nu(X)$, 
so the base case is immediate.  For the inductive step, 
suppose that $S \supseteq \V(X)$, $|S - \V(X)| = k+1$, and the claim
holds for all sets $S'$ such that $S' \supseteq \V(X)$ and $|S' - \V(X)|
= k$.  
Choose $s \in S - \V(X)$  and let $S'=S-\{s\}$.  
We show that $E^S_\nu(X)=E^{S'}_\nu(X)$. There are three cases.
\begin{enumerate}
\item $s = \min(S)$:  
Since $s\not\in\V(X)$, we must have $(X>s)=W$.
Thus,
$$\begin{array}{ll}
&E^S_\nu(X)\\
=
&s+(s_1-s)\nu(X>s)+(s_2-s_1)\nu(X>s_1)+ \cdots+(s_n-s_{n-1})\nu(X>s_{n-1})\\
= &s + (s_1-s) + (s_2-s_1)\nu(X>s_1) + \cdots + (s_n-s_{n-1})\nu(X>s_{n-1})\\
= &s_1 + (s_2-s_1)\nu(X>s_1) + \cdots + (s_n-s_{n-1})\nu(X>s_{n-1}) \\
= &E^{S'}_\nu(X). 
\end{array}
$$
\item 
$s = \max(S)$.  Let $s_n = \max(S')$.  
Since $s\not\in\V(X)$, $(X>s_n)=\emptyset$. Thus,
$$\begin{array}{ll}
&E^S_\nu(X)\\
=
&s_1+(s_2-s_1)\nu(X>s_1)+ \cdots+(s_n-s_{n-1})\nu(X>s_{n-1})+(s-s_n)\nu(X>s_n)\\ 
= &s_1+(s_2-s_1)\nu(X>s_1)+ \cdots+(s_n-s_{n-1})\nu(X>s_{n-1})\\
= &E^{S'}_\nu(X). 
\end{array}$$
\item $S=\{s_1,\ldots,s_k,s,s_{k+1},\ldots,s_n\}$ for some
$s_1,\ldots,s_n$, such that
$s_1<\ldots<s_k<s<s_{k+1}<\ldots<s_n$. Since $s\not\in\V(X)$ and
$s>s_k$, we have $(X>s)=(X>s_k)$. Thus,
$$\begin{array}{l}

(s_{k+1} - s_{k})\mu(X > s_{k}) = 
(s_{k+1} - s)\mu(X > s_{k}) + (s- s_{k})\mu(X > s_{k})\\
 =  (s_{k+1} - s)\mu(X > s) + (s- s_{k})\mu(X > s_{k}).
\end{array}$$
It is immediate that $E^S_\nu(X)=E^{S'}_\nu(X)$.
\end{enumerate}
This completes the proof of the inductive step.
\eprf

\lem\label{l:bel-plaus-exp}
The functions $E_{\Bel}$ and $E_{\Plaus}$ satisfy (\ref{belexprop2}). 
\elem
\prf
Suppose that $X$ and $Y$ are comonotonic gambles.  We claim that
there exist pairwise disjoint sets $U_1, \ldots, U_n$ 
and reals $a_1, \ldots, a_n, b_1, \ldots, b_n$
such that
\begin{itemize}
\item[(a)] $X = a_1X_{U_1}+ \cdots+a_nX_{U_n}$, 
\item[(b)] $Y = b_1X_{U_1}+ \cdots+b_nX_{U_n}$, 
\item[(c)] $a_1\le\ldots\le a_n$, and 
\item[(d)] $b_1\le\ldots\le b_n$.
\end{itemize}
To see this, suppose that $\V(X)=\{a_1',\ldots,a_m'\}$, where
$a_1'<\ldots<a_m'$, and 
$\V(Y)=\{b_1',\ldots,b_{m'}'\}$, where $b_1'<\ldots<b_{m'}'$.  Define 
$U'_i = X^{-1}(a_i')$ for $i = 1, \ldots, m$, and $U''_j = Y^{-1}(b_j')$
for $j = 1, \ldots, m'$.  Then clearly
(a) $X = a_1'X_{U'_1}+ \cdots+a_m'X_{U'_m}$, (b) $Y = b_1'X_{U''_1}+
\cdots+b_{m'}'X_{U''_{m'}}$, (c) the sets $U_1', \ldots, U_m'$ are
pairwise disjoint, (d) the sets $U_1'', \ldots, U_{m'}''$ are
pairwise disjoint, and (e)  $W = \union_{i=1}^{m} U'_i = 
\union_{i=1}^{m'} U''_i$.
Let $V_{ij}=U'_i\cap U''_j$. 
Note that if $V_{ij}$ and $V_{i'j'}$ are both nonempty and $i <
i'$, then $j \le j'$.  For suppose not; then $i < i'$ and $j > j'$.  
Thus, if $w \in V_{ij}$ and $w' \in V_{i'j'}$, then $(X(w) -
X(w'))(Y(w) - Y(w')) = (a_{i}' - a_{i'}')(b_{j}' - b_{j'}') < 0$,
contradicting the comonotonicity of $X$ and $Y$.  It follows that we can
take the claimed sets $U_1, \ldots, U_n$ to be the nonempty sets
$V_{ij}$ ordered lexicographically (so that if $U_k = V_{ij}$,
$U_{k'} = V_{i'j'}$, and $k < k'$, then either $i < i'$ or $i=i'$ and $j
< j'$).  
Moreover, if $U_k = V_{ij}$, take $a_k = a_i'$ and $b_k = b_j'$. 
It is clear that these choices have the required properties.
For this choice of $U_1, \ldots, U_n$, it follows that 
$X+Y =
(a_1+b_1)X_{U_1}+ \cdots+(a_n+b_n)X_{U_n}$, with $a_1+b_1\le\ldots\le
a_n+b_n$. By construction, we have
$\V(X+Y) = \{a_1+b_1,\ldots,a_n+b_n\}$. By
Lemma~\ref{l:choquetprop}, 
we have the following, where $\nu$ is either $\Bel$ or $\Plaus$:
$$\begin{array}{ll}
&E_{\nu}(X+Y) \\
= & 
   (a_1+b_1)+(a_2+b_2-a_1-b_1)\nu(X+Y>a_1+b_1)+ \cdots+\\
  & \quad (a_n+b_n-a_{n-1}-b_{n-1})\nu(X+Y>a_{n-1}+b_{n-1})\\
= & (a_1+b_1)+(a_2+b_2-a_1-b_1)\nu(U_2\cup\ldots\cup
   U_n)+ \cdots+\\
 & \quad (a_n+b_n-a_{n-1}-b_{n-1})\nu(U_n)\\
= & a_1+(a_2-a_1)\nu(U_2\cup\ldots\cup U_n)+ \cdots+(a_n-a_{n-1})\nu(U_n)+\\
     & \quad b_1+(b_2-b_1)\nu(U_2\cup\ldots\cup U_n)+ \cdots+(b_n-b_{n-1})\nu(U_n)\\
= & a_1+(a_2-a_1)\nu(X>a_1)+ \cdots+(a_n-a_{n-1})\nu(X>a_{n-1})+\\
  & \quad b_1+(b_2-b_1)\nu(Y > b_1)+ \cdots+(b_n-b_{n-1})\nu(Y > b_{n-1})\\
= & E_{\nu}(X)+E_{\nu}(Y),
\end{array}$$
where the last equality follows from Lemma~\ref{l:choquetprop} and the
fact that $\V(X) = \{a_1,\ldots,a_n\}$ and
$\V(Y) = \{b_1,\ldots,b_n\}$. 
\eprf

\opro{p:belexp-props}
The function $E_{\Bel}$ is superadditive, positively affinely homogeneous,
monotone, and satisfies 
(\ref{inex}) and (\ref{belexprop2}).
\eopro
\prf 
The fact that $E_{\Bel}$ is superadditive, positively affinely
homogeneous, and monotone follows immediately from
Theorem~\ref{t:bel-as-lp} and Proposition~\ref{p:lpexp-props}. 
The fact that $E_{\Bel}$ satisfies (\ref{inex}) follows essentially
from property B3 of belief functions and Proposition~\ref{Choqueteq}.  
First, it is easily checked that for
gambles $X$ and $Y$, $X\lor Y>x=(X>x)\cup(Y>x)$ and $X\land
Y>x=(X>x)\cap(Y>x)$. Consider the gamble $X_1\lor\ldots\lor X_n$. Let
$S=\V(X_1)\cup\ldots\cup\V(X_n)=\{s_1,\ldots,s_m\}$,
with $s_1<\ldots<s_m$. Let $s_0$ be an arbitrary real less than every
number in $S$. Clearly, $\V(X_1\lor\ldots\lor X_n)\subseteq
S$. Moreover, for every subset $I\subseteq\{1,\ldots,n\}$,
$\V(\land_{i\in I}X_i)\subseteq S$. By Proposition~\ref{Choqueteq}
and Lemma~\ref{l:choquetprop}, we have:
$$\begin{array}{ll}
&E_{\Bel}(X_1\lor\ldots\lor X_n)\\
= & s_1 + \sum_{k=1}^{m} (s_k-s_{k-1})\Bel(X_1\lor\ldots\lor
            X_n>s_{k-1}) \\
= & s_1 \Bel(X_1\lor\ldots\lor X_n>s_0) + \sum_{k=1}^{m} 
          (s_k-s_{k-1})\Bel(X_1\lor\ldots\lor X_n>s_{k-1}) \\ 
= & s_1 \Bel(X_1>s_0 \cup \ldots \cup X_n>s_0) + \\
 & \quad \sum_{k=1}^{m}
          (s_k-s_{k-1})\Bel(X_1>s_{k-1}\cup\ldots\cup X_n>s_{k-1})\\
\ge & s_1 \sum_{i=1}^{n} \sum_{I\subseteq\{1,\ldots,n\},|I|=i}
          (-1)^{i+1} \Bel(\cap_{j\in I}X_j>s_0) +\\
& \quad \sum_{k=1}^{m}
          (s_k-s_{k-1}) \sum_{i=1}^{n} \sum_{I\subseteq\{1,\ldots,n\},|I|=i} 
          (-1)^{i+1} \Bel(\cap_{j\in I}X_j>s_{k-1})\\
= & \sum_{i=1}^{n} \sum_{I\subseteq\{1,\ldots,n\},|I|=i} (-1)^{i+1}
          (s_1 \Bel(\cap_{j\in I}X_j>s_0) +\\
 & \quad \sum_{k=1}^{m}
          (s_k-s_{k-1}) \Bel(\cap_{j\in I}X_j>s_{k-1})) \\
= & \sum_{i=1}^{n} \sum_{I\subseteq\{1,\ldots,n\},|I|=i} (-1)^{i+1} 
          (s_1 + \sum_{k=1}^{m} (s_k-s_{k-1}) \Bel(\land_{j\in
          I}X_j>s_{k-1})) \\ 
= & \sum_{i=1}^{n} \sum_{I\subseteq\{1,\ldots,n\},|I|=i} (-1)^{i+1} 
          E_{\Bel}(\land_{j\in I}X_j).
\end{array}
$$

Finally, the fact that $E_{\Bel}$ satisfies (\ref{belexprop2}) follows
from Lemma~\ref{l:bel-plaus-exp}.   
\eprf

The following property turns out to be useful in various proofs
involving expectation for belief functions. 

\lem\label{l:bel-property}
If $E$ is an expectation function that satisfies (\ref{belexprop2})
and positive affine homogeneity, then it satisfies the following property on
indicator functions:   
\begin{equation}
\label{belexprop3}
\begin{array}{l}
\mbox{If $U_1 \supseteq U_2 \supseteq \ldots \supseteq U_n$ and
$a_1, \ldots, a_n \ge 0$},\\
\mbox{then}~E(a_1 X_{U_1} + \cdots + a_n X_{U_n}) = 
a_1 E(X_{U_1}) + \cdots + a_n
E(X_{U_n}).
\end{array}
\end{equation}
\elem
\prf
Define $S_k=a_1X_{U_1}+ \cdots+a_k X_{U_k}$, for $k=1,\ldots,n$. 
Since $a_i>0$
for $i = 1, \ldots, n$, it is not hard to show that 
$S_k$ and $a_{k+1} X_{U_{k+1}}$ are comonotonic for all $k$.
For if $S_k(w) > S_k(w')$, since $U_1 \supseteq \ldots \supseteq U_k$,
there must be some $j \le k$ such that 
$w \in U_j$ and $w' \notin U_j$.  Since $U_k \supseteq U_{k+1}$, it must
be the case that $w' \notin U_{k+1}$.  Hence $a_{k+1} X_{U_{k+1}}(w') =
0$ and $a_{k+1} X_{U_{k+1}}(w) \ge 0$.  Thus, we have comonotonicity.
Since $E$ satisfies (\ref{belexprop2}), it
follows that 
$$\begin{array}{ll}
&E(a_1 X_{U_1} + \cdots + a_n X_{U_n})\\
= &E(S_n) = E(S_{n-1}+a_n X_{U_n}) = E(S_{n-1})+ E(a_n X_{U_n}) = \ldots\\
= & E(a_1 X_{U_1}) + \cdots + E(a_n X_{U_n})
\end{array}
$$
The result is now immediate, since $E$ satisfies positive homogeneity.
\eprf

\othm{t:belexp-char}
Suppose that $E$ is an expectation function
that is positively affinely homogeneous,
monotone, and satisfies (\ref{inex}) and (\ref{belexprop2}).
Then there is a (necessarily unique) belief function $\Bel$
such that $E = E_{\Bel}$. \eothm
\prf
Define $\Bel(U) = E(X_U)$.  Just as in the case of probability,
it follows from positive affine homogeneity and monotonicity that
$\Bel(\emptyset) = 0$, $\Bel(W) = 1$, and $0 \le \Bel(U) \le 1$ for all
$U \subseteq W$.  By (\ref{inex}) (specialized to indicator functions),
it follows that $\Bel$ satisfies B3.  Thus, $\Bel$ is a belief function.
Now if $X$ is a gamble such that $\V(X) = \{x_1, \ldots, x_n\}$ and $x_1
< x_2 < \ldots < x_n$, then it is easy to check that
$$X = x_1 X_W + (x_2 - x_1) X_{X > x_1} + \cdots + (x_n - x_{n-1}) X_{X
> x_{n-1}}.$$

Clearly $W \supseteq (X > x_1) \supseteq
\ldots \supseteq (X > x_{n-1})$.  Thus, by Lemma~\ref{l:bel-property},
$$\begin{array}{lll}
E(X) &= &x_1 E(X_W) + (x_2 - x_1) E(X_{X > x_1}) + \cdots + (x_n -
x_{n-1})E(X_{X 
> x_{n-1}})\\
&= &x_1 + (x_2 - x_1)\Bel(X > x_1) + \cdots + (x_n - x_{n-1}) \Bel(X >
x_{n-1})\\
&= &E_{\Bel}(X). 
\end{array}$$

For uniqueness, observe that if $E_{\Bel} = E_{\Bel'}$, then
$\Bel(U) = E_{\Bel}(X_U) = E_{\Bel'}(X_U) = \Bel'(U)$ for all $U
\subseteq W$.
\eprf

\opro{p:possexp-props}
The function $E_{\Poss}$ is positively affinely  homogeneous,
monotone, and satisfies (\ref{belexprop2}) and (\ref{maxprop}).
\eopro

\prf
Recall from Section~\ref{s:possexp} that a possibility measure is just
a plausibility function. Therefore, by Theorem~\ref{t:bel-as-lp} and
Proposition~\ref{p:schmeidler}, we have $E_{\Poss}=\uE_{\P_{\Bel}}$, where
$\Bel$ is the belief function whose corresponding plausibility
function is $\Poss$. By Proposition~\ref{p:lpexp-props}, we 
immediately have that $E_{\Poss}$ is positively affinely homogeneous,
and monotone. By Lemma~\ref{l:bel-plaus-exp}, we have that $E_{\Poss}$
satisfies (\ref{belexprop2}). Showing that $E_{\Poss}$ satsifies
(\ref{maxprop}) is straightforward: $E_{\Poss}(X_{\cup_i\, U_i}) =
\Poss(\cup_i\,U_i) = \max_i \Poss(U_i) = \max_i E_{\Poss}(X_{U_i})$. 
\eprf

\othm{t:possexp-char}
Suppose that $E$ is an expectation function
that is positively affinely homogeneous,
monotone, and satisfies (\ref{belexprop2}) and (\ref{maxprop}).
Then there is a (necessarily unique) possibility measure $\Poss$
such that $E = E_{\Poss}$. 
\eothm
\prf
Define $\Poss(U)=E(X_U)$. Just as in the case of probability and
belief functions, it follows from positive affine homogeneity and
monotonicity that $\Poss(\emptyset)=0$, $\Poss(W)=1$, and
$0\le\Poss(U)\le 1$ for all $U\subseteq W$. By (\ref{maxprop}), it
follows that $\Poss$ satisfies Poss3. Thus, $\Poss$ is a possibility
measure. The proof that $E=E_{\Poss}$ is the same as that of
Theorem~\ref{t:belexp-char}, which is not surprising since $\Poss$ is
a plausibility function. More precisely, let $X$ be a gamble such that
$\V(X)=\{x_1,\ldots,x_n\}$ with $x_1<\ldots<x_n$, and write $X$ as
$x_1 X_W + (x_2-x_1)X_{X>x_1} + \cdots +
(x_n-x_{n-1})X_{X>x_{n-1}}$. Since $W\supseteq
(X>x_1)\supseteq\ldots\supseteq (X>x_{n-1})$, we can apply
Lemma~\ref{l:bel-property} to get
$$\begin{array}{lll}
E(X) &= &x_1 E(X_W) + (x_2 - x_1) E(X_{X > x_1}) + \cdots + (x_n -
x_{n-1})E(X_{X 
> x_{n-1}})\\
&= &x_1 + (x_2 - x_1)\Poss(X > x_1) + \cdots + (x_n - x_{n-1}) \Poss(X >
x_{n-1})\\
&= &E_{\Poss}(X).
\end{array}$$
\eprf

\subsection{Proofs for Section~\protect{\ref{sec:expressive}}}\label{appa4}

The following transformations on formulas of $\LE$ will be used in the 
proofs of Theorems~\ref{thm:expressive}, \ref{t:soundcomplprob},
\ref{t:soundcomplbel}, and \ref{t:soundcomplposs}. We prove here that
these transformations preserve satisfiability of the formulas, with
respect to the appropriate structures. 

First, define for a formula $f$ of $\LE$, the transformation $f^{T_1}$
inductively on the structure of $f$, taking $(a_1 e(\gamma_1)+\cdot+
a_n e(\gamma_n)\geq b)^{T_1}$ to be $t_1 + \cdots + t_n\geq b$, where
$t_i$ is obtained from $a_i$ and $\gamma_i=c_{1,i}\phi_{1,i}+\cdots
c_{k,i}\phi_{k,i}$ as $a_i c_{1,i} e(\phi_{1,i}) + \cdots + a_i
c_{k,i} e(\phi_{k,i})$. We define $(f_1\land f_2)^{T_1}$ and $(\neg
f_1)^{T_1}$ as $f_1^{T_1}\land f_2^{T_1}$ and $\neg(f_1^{T_1})$,
respectively.

\lem\label{l:transf1} 
If $M\in\MProb$, then $M\sat f$ if and only if $M\sat f^{T_1}$. 
\elem
\prf
We proceed by induction on the structure of $f$. 
For the base case,
\[\begin{array}{ll}
&M\sat a_1 e(\gamma_1) + \cdots \geq b\\
\mbox{iff~} &a_1 E_\mu(X_{\gamma_1}) + \cdots \geq b\\
\mbox{iff~} &a_1
  E_\mu(c_{1,1}X_{\eventM{\phi_{1,1}}}+\cdots+c_{k,1}X_{\eventM{\phi_{k,1}}})
  + \cdots \geq b\\
\mbox{iff~} &a_1 c_{1,1} E_\mu(X_{\eventM{\phi_{1,1}}}) + \cdots
+ a_1 c_{k,1} E_\mu(X_{\eventM{\phi_{k,1}}}) + \cdots \geq b\\
\mbox{iff~} &M\sat a_1 c_{1,1} e(\phi_{1,1}) + \cdots + a_1
c_{k,1} e(\phi_{k,1}) + \cdots \geq b.
  \end{array}\]
The inductive cases are immediate. 
\eprf

The second transformation is a bit more complicated. As before, we
define a transformation taking a formula $f$ of $\LE$ to a formula
$f^{T_2}$ by induction on the structure of $f$. 
Intuitively, we translate every expectation term $e(\gamma)$ appearing 
in $f$ into an expectation term of the form given by
Proposition~\ref{Choqueteq}, namely
$d_0+(d_1-d_0)e(\psi_1)+\ldots+(d_m-d_{m-1})e(\psi_m)$ for formulas
$\psi_1,\ldots,\psi_m$. 
As before, take $(f_1\land f_2)^{T_2}$ and $(\neg f_1)^{T_2}$ to be
$f_1^{T_2}\land f_2^{T_2}$ and $\neg(f_1^{T_2})$, respectively.
Take $(a_1
e(\gamma_1)+\cdots+ a_n e(\gamma_n)\geq b)^{T_2}$ to be $t_1 + \cdots +
t_n\geq b$, where $t_i$ is obtained from $a_i$ and
$\gamma_i=c_{1,i}\phi_{1,i}+\cdots+c_{k,i}\phi_{k,i}$ as follows.  Let
$p_1,\ldots,p_N$ be the primitive propositions that appear in $f$.
Let an \emph{atom over $p_1, \ldots, p_N$} be a formula of the form
$q_1 \land \ldots \land q_N$, where $q_i$ is either $p_i$ or $\neg
p_i$.  There are clearly $2^N$ atoms over $p_1, \ldots, p_N$. Let
$\delta_1,\ldots,\delta_{2^N}$ be an arbitrary ordering of these atoms. It
is easy to see that atoms are pairwise disjoint. It is also easy
to see that any formula over $p_1, \ldots, p_N$ can be written in a
unique way as a disjunction of atoms. Using these atoms, we construct
a gamble $\gamma'_i$ equivalent to $\gamma_i$
as follows.
For $j\in\{1,\ldots,2^N\}$, let $c_j = \sum_{\{l ~:~
\text{$\delta_j\rimp\phi_{l,i}$ is a prop. tautology}\}}c_{l,i}$. 
It is straightforward to check that $\gamma_i = \sum_{j=1}^{2^n} c_j
\delta_j$.  
Let $d_0,\ldots,d_m$ be the distinct 
elements of $\{c_j\}_{j\in\{1,\ldots,2^N\}}$, with $d_0 < \ldots <
d_m$. For $j\in\{1,\ldots,m\}$, let $\psi_j =
\bor_{l\in\{1,\ldots,2^N\}, c_l\geq d_j}\delta_l$. Note
that $\psi_0$ is logically equivalent to $\truep$; moreover,
$\psi_{j+1}\rimp\psi_j$ is a propositional tautology for all
$j\in\{0,\ldots,m-1\}$. Finally, let $\gamma'_i=d_0 + (d_1-d_0) \psi_1 +
\cdots + (d_m-d_{m-1}) \psi_m$. 
One can check that $\gamma_i$ and $\gamma'_i$ are equivalent gambles. 
Indeed, $\gamma'_i$ is provably equivalent to
$d_0(\sum_{l=1}^{2^N}\delta_l)+(d_1-d_0)(\sum_{c_l\ge
d_1}\delta_l)+\ldots+(d_m-d_{m-1})(\sum_{c_l\ge d_m} \delta_l)$. 
Distributing
the coefficients and putting like terms together, we see that the
coefficient of $\delta_l$ is just
$d_0+(d_1-d_0)+\ldots+(d_{r_\ell}-d_{{r_\ell}-1})=d_{r_\ell}$, where
$r_\ell=\max\{r : c_l\ge d_r\}$. By definition of $d_1,\ldots,d_m$,
$d_{r_\ell}=c_l$, and therefore $\gamma_i$ and $\gamma'_i$ are equivalent.
Note that, by Lemma~\ref{l:bel-property}, if $E$ is an expectation
function that satisfies comonotonicity and positive homogeneity, then
$E(X_{\gamma_i'}) = d_1 + (d_{1}-d_{0}) E(X_{\eventM{\psi_{1}}})
+ \cdots + (d_{m}-d_{m-1}) E(X_{\eventM{\psi_{m}}})$.
Let $t_i$ be $a_i d_0 + a_i 
(d_1-d_0) e(\psi_1) + \cdots + a_i (d_m-d_{m-1}) e(\psi_m)$.  

\lem\label{l:transf2}
If $M\in\MProb\cup\MBel\cup\MPoss$, then $M\sat f$ if and
only if $M\sat f^{T_2}$.
\elem
\prf
We proceed by induction on the
structure of $f$, for $M\in\MBel$; the result follows for structures
in $\MProb$ and $\MPoss$, since these can be viewed as structures in
$\MBel$. 
For the base case, we have
\[\begin{array}{ll}
&M\sat a_1 e(\gamma_1) + \cdots a_n e(\gamma_n) \geq b\\
\mbox{iff~} &a_1 E_\Bel(X_{\gamma_1}) + \cdots a_n E_\Bel(X_{\gamma_n})
\geq b\\
\mbox{iff~} &a_1 E_\Bel(X_{\gamma'_1}) + \cdots  + a_n
E_\Bel(X_{\gamma_n'}) \geq b\\
\mbox{iff~} &a_1
  E_\Bel(d_{0,1} + (d_{1,1}-d_{0,1})
X_{\eventM{\psi_{1,1}}}+\cdots+(d_{m_1,1}-d_{m_1-1,1})X_{\eventM{\psi_{m_1,1}}}) 
  +\\
& \quad \cdots + a_n   E_\Bel(d_{0,n} + (d_{1,n}-d_{0,n})
X_{\eventM{\psi_{1,n}}}+\cdots+\\
& \qquad (d_{m_n,n}-d_{m_n-1,n})X_{\eventM{\psi_{m_n,n}}})
\geq b\\ 
\mbox{iff~} &a_1 d_{0,1} + a_1 (d_{1,1}-d_{0,1}) E_\Bel(X_{\eventM{\psi_{1,1}}})
+ \cdots +\\
& \quad  a_1 (d_{m_1,1}-d_{m_1-1,1}) E_\Bel(X_{\eventM{\psi_{m_1,1}}}) +
\cdots +\\
& \qquad a_n d_{0,n} + a_1 (d_{1,n}-d_{0,n}) E_\Bel(X_{\eventM{\psi_{n,1}}})
+ \cdots +\\
& \quad\qquad a_n (d_{m_n,n}-d_{m_n-1,n}) E_\Bel(X_{\eventM{\psi_{m_n,n}}})
\geq b\\ 
\mbox{iff~} &M\sat a_1 d_{0,1} + a_1 (d_{1,1}-d_{0,1}) e(\psi_{1,1}) +
\cdots + a_1 (d_{m,1}-d_{m-1,1}) e(\phi_{m,1}) + \cdots +\\
& \quad a_n d_{0,n} + a_n (d_{1,n}-d_{0,n}) e(\psi_{1,n}) +
\cdots
a_n (d_{m,n}-d_{m-1,n}) e(\phi_{m,n}) \geq b.
  \end{array}\]
The fact that we can  distribute
the expectation $E_\Bel$ of the gamble into a sum of expectation of
indicator functions follows from comonotonicity, using
Lemma~\ref{l:bel-property}. 

 The inductive cases are immediate.   
\eprf

\othm{thm:expressive} $\LE$ and $\LPP$ are equivalent in expressive
power with respect to $\MProb$, $\MBel$, and $\MPoss$.  $\LE$ is
strictly more expressive than $\LPP$ with respect to $\MLP$.  \eothm
\prf The proof follows the lines laid out in
Section~\ref{sec:expressive}. First, we show that $\LE$ is at least as
expressive as $\LPP$ with respect to $\MProb$, $\MBel$, $\MLP$, and
$\MPoss$. Formally, we show that every formula $f$ of $\LPP$ is
equivalent to the formula $f^T \in \LE$ that results by replacing
$\ell(\phi)$ by $e(\phi)$.  
Define $f^T$ by induction on the structure 
of $f$, taking $(a_1\ell(\phi_1)+\cdots+a_n\ell(\phi_n)\geq b)^T$ to
be $a_1 e(\phi_1)+\cdots+a_n\ell(\phi_n)\geq b$, $(f_1\land f_2)^T$ to
be $f_1^T\land f_2^T$, and $(\neg f_1)^T$ to be $\neg(f^T)$. 

It suffices to show that for all $M$ in $\MProb$, $\MBel$, $\MLP$, or
$\MPoss$, we have $M\sat f$ if and only if $M\sat f^T$. First, note
if $\nu$ is a probability measure, belief function, or set of
probability measures, then
$E_\nu(X_U)=\nu(U)$, where $X_U$ is the
indicator function of $U$. Let $M$ be an arbitrary structure in
$\MProb$, $\MBel$, $\MLP$, or $\MPoss$, with associated measure of
uncertainty $\mu$. 
Thus, if $\nu$ is the measure associated with structure $M$, we have
that $\nu(\eventM{\phi}) = E_\nu(\gambM{X_{\eventM{\phi}}})$.  
(If $\nu$ is the set $\P$  or probability measures, we take $E_\nu =
\lE_{\P}$.)
The result now follows immediately by a straightforward induction on the
structure of $f$.

We next show that $\LPP$ is as expressive as $\LE$ with respect to
$\MProb$, $\MBel$, and $\MPoss$. 
Let $\LEp$ be the sublanguage of $\LE$ where the 
expectation operator $e$ is applied only to propositional formulas, as
opposed to arbitrary gambles.  It is immediate from the construction
in the previous paragraph that if $f \in \LPP$, then $f^T \in \LEp$.
Moreover, it is easy to see that every formula $f' \in \LEp$ 
is of the form $f^T$ (where $f$ is the result of 
replacing all occurrences of $e(\phi)$ in $f'$ by $\ell(\phi)$).  
Since we showed above that $f$ is equivalent to $f^T$, to prove that
$\LPP$ is as expressive to $\LE$ with respect to $\MProb$, $\MBel$, and
$\MPoss$, it suffices to show that every formula in $\LE$ is equivalent
to a formula in $\LEp$ in structures in $\MProb \union \MBel \union
\MPoss$. But this is immediate from Lemma~\ref{l:transf2}, since the
formula $f^{T_2} \in \LEp$.  

Finally, we show that $\LE$ is strictly more expressive than $\LPP$
with respect to $\MLP$. To establish this, it is sufficient to exhibit
two structures in $\MLP$ that satisfy the same formulas in $\LPP$, but
can be distinguished by a formula $f$ in $\LE$. (This means that $f$
in $\LE$ cannot be equivalent to any formula in $\LPP$.) We use a
variant of Example~\ref{xamdetermination}. Consider the
two  $\MLP$ structures $M_1=(W,\P_1,\pi)$ and $M_2=(W,\P_2,\pi)$,
where $W=\{w_1,w_2,w_3\}$, $\pi$ is such that $p$ is true at $w_1$,
both $p$ and $q$ are true at $w_2$, and neither $p$ nor $q$ is true at
$w_3$, and the sets of probability measures $\P_1$ and $\P_2$ are
given as follows. We can describe a probability measure on $W$ by a
tuple $(a_1,a_2,a_3)$, where $a_i$ gives the probability at
$w_i$. Define $\P_1=\{(1/3,2/3,0),(0,1/3,2/3),(2/3,0,1/3)\}$, and
$\P_2=\P_1\cup\{(1/3,0,/2,3)\}$. It is easy to check that $(\P_1)_*$ and
$(\P_2)_*$ are both $0$ on singleton subsets of $W$ and $1/3$ on
doubleton subsets. Hence, $(\P_1)_*=(\P_2)_*$. Therefore, no formula
involving only lower probability can distinguish these two
structures, and $M_1$ and $M_2$ satisfy the same formulas of
$\LPP$. Now, consider the gamble $p+q$. It is easy to see that 
$\gamb{p+q}{M_1}=\gamb{p+q}{M_2}$ is the random
variable $X$ defined by $X(w_1)=1$, $X(w_2)=2$, and $X(w_3)=0$. It is
not hard to check  that $\lE_{\P_1}(X)=2/3$, and
$\lE_{\P_2}(X)=1/3$. Hence, $M_1\sat e(p+q)>1/2$, and $M_2\sat
e(p+q)<1/2$, so that $M_2\sat\neg e(p+q)>1/2$. 
\eprf

\subsection{Proofs for Section~\protect{\ref{sec:axiomatization}}}\label{appa5}
The proofs of Theorems~\ref{t:soundcomplprob}--\ref{t:soundcomplposs}
use an approach similar to the one taken in FHM. We first
need some definitions.  We say a formula $\sigma$ is \emph{consistent}
with an axiom system $\ax$ (or simply $\ax$-consistent) if
$\neg\sigma$ is not provable from $\ax$. 
To show that $\ax$ is a complete axiomatization with
respect to some class of structure $\cM$, we must show that every
formula that is valid in every structure in $\cM$ is provable in
$\ax$. This is equivalent to showing that every $\ax$-consistent
formula $\sigma$ is satisfiable in $\cM$.

The proofs of Theorems~\ref{t:soundcomplprob}--\ref{t:soundcomplposs}
are all structured as follows. Soundness is straightforward in all
cases. Completeness is obtained by showing that a consistent
formula $f$ is satisfiable in the appropriate set of structures
$\cM$. More specifically, assume that formula $f$ is not satisfiable
in a structure in $\cM$; we must show that $f$ is
inconsistent. We first reduce $f$ to a canonical form. Let
$g_1\lor\ldots\lor g_r$ be a disjunctive normal form expression for
$f$ (where each $g_i$ is a conjunction of expectation inequalities and
their negations). Using propositional reasoning, we can show that $f$
is provably equivalent to this disjunction. Since $f$ is
unsatisfiable, each $g_i$ must also be unsatisfiable. Thus, it is
sufficient to show that any unsatisfiable conjunction of expectation
inequalities and their negations is inconsistent. Let $f$ be such a
formula. To show that it is inconsistent, we essentially construct a
system of inequalities $\hat{f}$ from $f$ by replacing every term
$e(\gamma_i)$ in $f$ by a variable $x_i$, with the property that $f$
is satisfiable over structures in $\cM$ if and only if $\hat{f}$ has a 
solution over the reals. Since $f$ is unsatisfiable, $\hat{f}$ has no
solution, so that $\neg f$ must be an instance of \axiom{Ineq}. (That
$\hat{f}$ has no solution means that $\hat{f}$ is not satisfiable as a 
formula about linear inequalities, so that $\neg\hat{f}$ is a valid
formula about linear inequalities.) Therefore, $\neg f$ is provable,
and $f$ is inconsistent. The details of how to construct $\hat{f}$
differ for each representation of uncertainty.

\othm{t:soundcomplprob} 
$\AXProb$ is a sound and complete axiomatization of $\LE$ with respect
to $\MProb$.   
\eothm
\prf
Soundness is straightforward. For completeness, we proceed as above.
Without loss of generality, assume that $f$ is a conjunction of
expectation inequalities and their negations. Using axioms \axiom{E1}
and \axiom{E2}, we can convert $f$ into the equivalent formula
$f^{T_1}$ (Lemma~\ref{l:transf1}) where $e$ is applied only to
propositional formulas. For every propositional formula $\phi$ in
$f^{T_1}$, $\phi$ is equivalent to $\lor_{j=1}^k\delta_{i_j}$, where
$\delta_{i_1},\ldots,\delta_{i_k}$ are the atoms over the propositions
in $f$ such that $\delta_{i_j}\rimp\phi$ for all $1\leq j\leq
k$. Since $\neg(\delta_{i_j}\land\delta_{i_l})$ is a propositional
tautology for all $j\neq l$, $\phi=\delta_{i_1}+ \cdots+\delta_{i_k}$
is a valid formula about propositional gamble inequalities (see
Section~\ref{s:gambleineq}), and axiom \axiom{E5} yields
$e(\phi)=e(\delta_{i_1}+ \cdots+\delta_{i_k})$. This means that we can
find a formula $f'$ 
provably
equivalent to $f^{T_1}$, where 
$f'$ is formed by replacing  
each term $a e(\phi)$ of $f^{T_1}$ by the sum $a
e(\delta_{i_1}) + \cdots 
+ a e(\delta_{i_k})$, and then collecting like terms.  Let $f''$ be
obtained from $f'$ by adding as conjuncts to $f'$ all the
expectation inequality formulas $e(\delta_j)\geq 0$, for $1\leq j\leq
2^N$, $e(\delta_1)+ \cdots+e(\delta_{2^N})\geq 1$, and
$-e(\delta_1)-\cdots-e(\delta_{2^N})\geq -1$ (which together
essentially say that the sum of the probabilities of the atoms is
$1$). It is not hard to see that these formulas are provable, hence
$f''$ is provably equivalent to $f'$, and hence to $f$. We therefore
only have to show that $f''$ is satisfiable.

The negation of an expectation inequality $a_1 e(\gamma_1) + \cdots +
a_n e(\gamma_n)\geq b$ can be written $- a_1 e(\gamma_1) - \cdots -
a_n e(\gamma_n) > -b$. Thus, without loss of generality, we can assume 
that $f''$ is the conjunction of the formulas
\begin{align*}
e(\delta_1)+ \cdots+e(\delta_{2^N}) & \geq  1\\
-e(\delta_1)-\cdots-e(\delta_{2^N}) & \geq  -1\\
e(\delta_1) & \geq  0\\
 & \vdots  \\
e(\delta_{2^N}) & \geq  0\\
a_{1,1} e(\delta_1) + \cdots + a_{1,2^N} e(\delta_{2^N}) & \geq  b_1\\
& \vdots \\
a_{r,1} e(\delta_1) + \cdots + a_{r,2^N} e(\delta_{2^N}) & \geq  b_r\\
-a'_{1,1} e(\delta_1) - \cdots - a'_{1,2^N} e(\delta_{2^N}) & >  b'_1\\
& \vdots \\
-a'_{s,1} e(\delta_1) - \cdots - a'_{s,2^N} e(\delta_{2^N}) & >  b'_s.
\end{align*}
Consider the following system of inequations, obtained by replacing
$e(\delta_i)$ by the variable $x_i$:
\begin{align*}
x_1+ \cdots+x_{2^N} & \geq  1\\
-x_1-\cdots-x_{2^N} & \geq  -1\\
x_1 & \geq  0\\
 & \vdots  \\
x_{2^N} & \geq  0\\
a_{1,1} x_1 + \cdots + a_{1,2^N} x_{2^N} & \geq  b_1\\
& \vdots  \\
a_{r,1} x_1 + \cdots + a_{r,2^N} x_{2^N} & \geq  b_r\\
-a'_{1,1} x_1 - \cdots - a'_{1,2^N} x_{2^N} & >  b'_1\\
& \vdots  \\
-a'_{s,1} x_1 - \cdots - a'_{s,2^N} x_{2^N} & >  b'_s.
\end{align*}
Clearly if $f''$ is satisfiable, say in a structure $M$ with associated
probability measure $\mu$, then the system
is satisfiable, by taking $x_i = \mu(\eventM{\delta_i})$. Conversely, if
the system of inequations has a solution, then $f''$ is satisfied in a
structure $M$ with associated probability measure $\mu$ such that
$x_i = \mu(\eventM{\delta_i})$.

Returning to the main argument, suppose
by way of contradiction, that $f$ is not satisfiable. Then 
this system of inequations has no
solution.  It follows that $\neg f''$ is an instance of the axiom
\axiom{Ineq}. Since $f''$ is provably equivalent to $f$, it follows
that $\neg f$ is provable, so that $f$ is inconsistent, a
contradiction. 
\eprf

The
proofs of Theorems~\ref{t:soundcompllp}, \ref{t:soundcomplbel}, and
\ref{t:soundcomplposs} require some additional terminology. Let $f$
be a formula of $\LE$, and let $p_1,\ldots,p_N$ be the primitive
propositions that appear in $f$. Observe that there are $2^{2^N}$
inequivalent propositional formulas over $p_1,\ldots,p_N$. The
argument goes as follows. Recall the notion of an atom over
$p_1,\ldots,p_N$, introduced before Lemma~\ref{l:transf2}. It is easy
to see that any formula over $p_1,\ldots,p_N$ can be written in a
unique way as a disjunction of atoms. There are $2^{2^N}$ such
disjunctions, so the claim follows. Let $\rho_1,\ldots,\rho_{2^{2^N}}$
be some canonical listing of the inequivalent formulas over
$p_1,\ldots,p_N$. Without loss of generality, we assume that $\rho_1$
is equivalent to $\truep$, and $\rho_{2^{2^N}}$ is equivalent to
$\falsep$.
Following FHM, we call these formulas $\rho_i$, $i = 1, \ldots, 2^{2^N}$
{\em regions}.

\othm{t:soundcompllp}
$\AXLP$ is a sound and complete axiomatization 
of $\LE$ with respect to $\MLP$.
\eothm

\prf
Soundness is straightforward. For completeness, we proceed as above.
Again,
without loss of generality, assume that $f$ is a conjunction of
expectation inequalities and their 
negations. 

The first step of the proof is to find a formula $f'$ provably
equivalent to $f$, where the gambles in $f'$ are expressed in terms of
atoms $\delta_1,\ldots,\delta_{2^N}$. 
As observed above, every propositional formula $\phi$ in
$f$ is equivalent to some region $\rho_i$, $1\leq i\leq 2^{2^N}$,
which is of the form $\lor_{j=1}^k\delta_{i_j}$, where
$\delta_{i_1},\ldots,\delta_{i_k}$ are the atoms over the propositions
in $f$ such that $\delta_{i_j}\rimp\phi$ for all $1\leq j\leq
k$. Since $\delta_{i_j}\land\delta_{i_l}\riff\falsep$ is a
propositional tautology for all $j\neq l$,
$\phi=\delta_{i_1}+ \cdots+\delta_{i_k}$ is 
valid.
It easily follows that 
a propositional gamble $a_1\phi_1 + \cdots
+ a_n\phi_n$ is equal to a propositional gamble of the form
$a_1'\delta_1 + \cdots + a_{2^N}'\delta_{2^N}$. 
By axiom \axiom{E5},  $e(a_1\phi_1+ \cdots+a_n\phi_n)=e(a_1'\delta_1+
\cdots+a_{2^N}'\delta_{2^N})$
is provable.
This means we can find a formula $f'$
provably equivalent to $f$, where each term $a e(\gamma)$ of $f$ is
replaced by a term $a e(a_1'\delta_1+ \cdots+a_{2^N}'\delta_{2^N})$.
In fact, 
the following arguments shows that, without loss of generality, 
we can take the $a_i'$s to be nonnegative.
Given a gamble $a_1'\delta_1+\ldots+a_{2^N}'\delta_{2^N}$,
let $b=\min_{1\leq i\leq 2^N}(a_i')$. We can easily show that 
$(a_1'+b)\delta_1+\ldots+(a_{2^N}'+b)\delta_{2^N}-b =
a_1'\delta_1+\ldots+a_{2^N}'\delta_{2^N}$ is a valid 
gamble equality. Thus, every expectation inequality formula $c_1
e(\gamma_1)+\ldots+c_k e(\gamma_k)\ge d$ in $f'$ is provably
equivalent to a formula 
of the form
$c_1 e(\gamma_1')+c_1b_1+\ldots+c_k
e(\gamma_j')+c_k b_k\ge d$, 
which in turn is
provably equivalent to $c_1
e(\gamma_1')+\ldots+c_k e(\gamma_k')\ge d-c_1 b_1-\ldots-c_k b_k$. 
Thus, without loss of generality, we can take the coefficients
of the gambles in $f'$ to be nonnegative.

The next step of the proof
is to derive a finite system of inequations
$\hat{f}$ corresponding to $f'$ such that $f'$ is satisfiable if and
only $\hat{f}$ has a solution. 
Suppose that we have found such a formula $\hat{f}$ and that $f'$
is unsatisfiable. Then $\hat{f}$ has no solution. Thus, the formula
$\neg f'$ is an instance of \axiom{Ineq}. Since $f'$ is provably
equivalent to $f$, $\neg f$ is provable, and thus $f$ is inconsistent,
a contradiction. Therefore, $f'$, and hence $f$, is satisfiable. The
remainder of the proof consists of coming up with $\hat{f}$, and
showing that it has a solution if and only if $f'$ is satisfiable.

We would like $\hat{f}$ to
be a system of inequalities that yields a solution that can be used to
construct a lower expectation function that satisfies $f'$. 
We can force a solution to $\hat{f}$ to be a lower expectation function
by adding to $\hat{f}$ inequalities corresponding to
all the instances of axioms 
\axiom{E6}--\axiom{E8}. 
Unfortunately, there are infinitely many such instances. 
The aim now is therefore to derive a finite set of 
instances of 
\axiom{E6}--\axiom{E8}
that is sufficient to constrain a solution to $\hat{f}$ to be a lower
expectation function.  

To do this, let $\gamma_1, \ldots, \gamma_n$ be the propositional gambles
$\gamma$ such that $e(\gamma)$ appears in $f'$, together with the
propositional gambles $\falsep$ and $\truep$ (if they do not already
appear in $f'$). 
For ease of exposition, take 
$\gamma_{n-1} = \falsep$ and $\gamma_{n} = \truep$.
As we saw earlier, each $\gamma_i$ is provably
equivalent to a formula of the form
$a_{i,1} \delta_1 + \cdots + a_{i,2^N}\delta_{2^N}$, where
$\delta_1,\ldots,\delta_{2^N}$ are the atoms over the primitive
propositions appearing in $f$, and $a_{i,j}\geq 0$.
We now construct, for every
propositional gamble 
$\gamma_i$ ($1\le i \le n-1$), a finite set $B_i$ of vectors. (We do
not need a vector for the gamble $\gamma_n=\truep$.)
We give the construction of $B_1$, and then describe
the minor modifications needed to construct $B_i$ for $i = 2,
\ldots, n-1$.
We then show how to use these sets to identify a finite set of instances
of \axiom{E6}--\axiom{E8}.

Let $A_1$ be the $(n-1)\times 2^N$ matrix of real numbers 
whose $i$th row consists of the coefficients of
$\delta_1,\ldots,\delta_{2^N}$ in $\gamma_{i+1}$, 
$i = 1, \ldots, n-1$.
Let $B_1'$ be the set of vectors
$\vec{b}=\<b_2,\ldots,b_{n}\>$ such that $b_2,\ldots,b_{n-1}\geq 0$ 
($b_{n}$ can be negative) and 
\begin{equation}\label{e:b1}
\vec{b} A_1 \le \<a_{1,1},\ldots,a_{1,2^N}\>.
\end{equation}
It is easy to see that $B_1'$ is a closed convex set: it is clearly
closed, and if $\vec{b}_1, \vec{b}_2 \in B_1'$, then so is
$\theta\vec{b}_1 + (1-\theta)\vec{b}_2$, for all $\theta \in [0,1]$.
Thus, for each vector
$\vec{y}=\<y_1,\ldots,y_{n-1}\>$ such that $y_1,\ldots,y_{n-2}\ge 0$,
there exists a $\vec{b}_y\in B_1'$ such that $\vec{b}_y\cdot \vec{y}$ is
maximal (where 
$\cdot$ is the inner product). It is a consequence of the Krein-Milman
Theorem (see \citeN{Rudin91}, for instance) that the set of such $\vec{b}_y$
is a finite subset $B_1$ of $B_1'$.

\commentout{
Similarly, let $C_1'$ be the set of vectors $\vec{c}=\<c_2,\ldots,c_{n+1}\>$
such that $c_2,\ldots,c_{n}\geq 0$ ($c_{n+1}$ can be negative) and
\begin{equation}\label{e:c1}
c A_1 \le \<0,\ldots,0\>.
\end{equation}
Again, $C_1'$ is a closed convex set. For any choice of
$y=\<y_1,\ldots,y_{n}\>$ such that $y_1,\ldots,y_{n-1}\ge 0$,
there exists a $\vec{c}_y\in C_1'$ such that $\vec{c}_y\cdot y$ is
maximal.
It is again a consequence of the
Krein-Milman Theorem that the set of such $vec{c}_y$ is a finite subset
$C_1$ of $C_1'$.
}
We use the set $B_1$ to derive a finite number of instances
of \axiom{E6}--\axiom{E8} as follows.  Note that if 
$\vec{b}=\<b_2,\ldots,b_{n}\>$
satisfies (\ref{e:b1}), then $b_2
\gamma_2 + \cdots + b_{n} \gamma_{n} \leq \gamma_1$ is a valid formula
about gamble inequalities. Therefore, by \axiom{E5}, $e(b_2 \gamma_2 +
\cdots + b_{n} \gamma_{n})\leq e(\gamma_1)$. By applying \axiom{E6},
\axiom{E7} and \axiom{E8}, we can derive $b_2 e(\gamma_2) + \cdots
b_{n-1} e(\gamma_{n-1}) + b_{n} \le e(\gamma_1)$. Let $F_{\vec{b}}$ be the
conjunction of the instances of axioms \axiom{E6}, \axiom{E7}, and
\axiom{E8} that are used to perform this derivation. 
Let $f_1$ be the conjunction of $F_{\vec{b}}$ for all $\vec{b}\in B_1$.
This process defines a set $B_1$ of vectors, and from these a
formula $f_1$. Repeat this process for the gambles $\gamma_2, \ldots,
\gamma_{n-1}$,
by interchanging the role of $\gamma_1$ and $\gamma_i$,
for $2\leq i \leq n-1$.
After doing this for all gambles,
we obtain sets $B_1,\ldots,B_{n-1}$ of vectors,
and formulas $f_1,\ldots,f_{n-1}$.

Let $f''$ be the conjunction of $f_1$, \ldots, $f_{n-1}$, $f'$, 
$e(\gamma_{n-1}) \ge 0$, $-e(\gamma_{n-1}) \ge 0$, $e(\gamma_{n}) \ge 1$, and
$-e(\gamma_{n}) \ge -1$ (that is, $e(\falsep) =
0$ and $e(\truep) = 1$, since $\gamma_{n-1} = \falsep$ and $\gamma_{n} =
\truep$, by assumption). 
Let
$\hat{f}$ be the conjunction of inequalities obtained by replacing
every instance of a term of the form $e(\gamma)$ in $f''$ by the
variable $x_\gamma$.  Note that, besides $e(\gamma_1), \ldots,
e(\gamma_{n})$, 
there are other terms of the form $e(\gamma)$ that
arise in $f_1, \ldots, f_{n-1}$; the gambles $\gamma$ in these terms are
linear combinations of $\gamma_1, \ldots, \gamma_{n}$.
We then obtain a system of inequalities over the variables
$x_\gamma$. We now show that $\hat{f}$ is our required system
of inequalities, that is, $f'$ is satisfiable if and only if $\hat{f}$
has a solution.

It is straightforward to show that if $f'$ is satisfiable, then
$\hat{f}$ has a solution. If $f'$ is satisfiable, then there
exists a lower probability structure $(W,\cP,\pi)$ such that
$(W,\cP,\pi)\sat f'$, and hence $(W,\cP,\pi)\sat f''$ (since the
instances of \axiom{E6}--\axiom{E8} are valid in lower probability
structures). Clearly, taking $x_\gamma=\lE_\cP(\gambM{\gamma})$ gives a
solution to $\hat{f}$. 

The interesting direction is showing that if $\hat{f}$ has a solution,
then $f'$ is satisfiable in $\MLP$. Suppose that $\hat{f}$ has a
solution in which $x_{\gamma_1}, \ldots, x_{\gamma_{n}}$ taken on
values $x^*_1, \ldots, 
x^*_{n}$, respectively.  
(Note that by the construction 
of $\hat{f}$, we must have 
$x^*_{n-1} = 0$ and
$x^*_{n}=1$.) Take
$W=\{\delta_1,\ldots,\delta_{2^N}\}$, and let $\pi(\delta_i)(p)=\true$
if and only if $\delta_i\rimp p$ is a propositional tautology. We need
to show that there exists a set of probability measures $\cP$ such
that $(W,\cP,\pi)\sat f'$. To do this, we start with the solution to
$\hat{f}$, which intuitively gives us a 
lower expectation function defined on a subset of gambles
(those that appear in $f$),
and show that it can be extended to a lower expectation function
defined on all gambles. This lower expectation function will give us
the set $\cP$ of probability measures necessary for satisfiability.
More precisely, corresponding to the propositional gambles
$\gamma_1,\ldots,\gamma_{n}$ appearing in $f$ (along with $\falsep$ and 
$\truep$, as described above), let $X_i$ be the
gamble corresponding to
$\gamma_i=a_{i,1}\delta_1+\ldots+a_{i,2^N}\delta_{2^N}$ defined on $W$ 
by $X_i(\delta_j)=a_{i,j}$, for all $1\le j\le 2^N$. Define the
function $\unPs$ on $\{X_1,\ldots,X_n\}$ by
$\unPs(X_i)=x^*_i$. Our goal is to extend $\unPs$ to a lower expectation 
function $\unEs$ defined on all gambles, such that $\unEs(Y)=\unPs(Y)$
for $Y\in\{X_1,\ldots,X_n\}$. 

The first thing we have to check is that $\unPs$ is a potential
candidate to be a partial lower expectation function.  This is made
precise by Walley's \citeyear[p.72]{Walley91} notion of coherence.
A function $\unP$ defined on a set
$\{X_1,\ldots,X_n\}$ of gambles is {\em coherent\/} if and only if for
all reals $b_1,\ldots,b_n\ge 0$ and $i^* \in \{1, \ldots, n\}$, we
have%
\footnote{Walley has an apparently stronger requirement,
namely, 
that for all $Y_1,\ldots,Y_m\subseteq \{X_1,\ldots,X_n\}$ and all reals
$b_1, \ldots, b_m$, we have $\sup_{w \in W} \left\{ [(\sum_{j=2}^m b_j
(Y_j-\widetilde{\unPs(Y_j)})) - b_1 (Y_1-\widetilde{\unPs(Y_1)})](w)
\right\} \ge 0$.  However, 
we can assume that all the $Y_j$'s are distinct without loss of
generality (since if there is a repetition of $X_i$'s, we can simply add
the coefficients $b_i$), and if any $X_i$ does not appear among the
$Y_i$'s, we can add it, taking $b_i = 0$.}
\begin{equation}\label{e:sup}
\sup_{w \in W} \left\{ \left[(\sum_{j \ne i} b_j (X_j-\widetilde{\unP(X_j))})
- b_{i^*} 
(X_{i^*}-\widetilde{\unP(X_{i^*})})\right](w) \right\} \ge 0.
\end{equation}
If $E$ is a lower expectation function (that is, if $E =
\lE_\P$ for some set $\P$ of probability measures) then~(\ref{e:sup})
holds (with $\unP$ replaced by $E$).  To see this, first note that
for any gambles
$Y, Y_1, \ldots, Y_n$ and nonnegative reals $b, b_1, \ldots, b_n$, 
by Proposition~\ref{p:lpexp-props}, 
$E(b(Y - \widetilde{E(Y)})) = 0$, and 
$E(b_1(Y_1 - \widetilde{E(Y_1)}) + \cdots + E(Y_n - \widetilde{E(Y_n)}) \ge
0$.  Now suppose that 
\[\sup_{w \in W} \left\{ \left[ (\sum_{j=1}^n b_j
(Y_j-\widetilde{E(Y_j)})) - b E(Y-\widetilde{E(Y)})\right](w) \right\}
= c < 0.\] 
Then $\sum_{j=1}^n b_j (Y_j - \widetilde{E(Y_j)}) \le
b(E(Y-\widetilde{E(Y)}) + \tilde{c}$. 
Thus $$E(\sum_{j=1}^n b_j (Y_j - \widetilde{E(Y_j)})) \le E(b(Y-\widetilde{E(Y)}) + \tilde{c}) =
E(b(Y - \widetilde{E(Y)})) + c = c < 0.$$
But this is a contradiction.  It follows that (\ref{e:sup}) holds for
$E$.  

This shows that coherence, that is, (\ref{e:sup}), is a necessary condition
for $\unP$ to be extendible to a lower expectation.  Walley's {\em
Natural Extension Theorem\/}, which we now state, shows that it is
sufficient as well.

\thm\label{thm:natext} {\rm \cite[p.123]{Walley91}}  Suppose that
$\unP$ is a coherent 
real-valued function defined on a set $\mathcal{G}$ of gambles on
$W$. Define  the \emph{natural extension} $\unE$ of $\unP$ as
\[ \unE(Y)=\sup\{\alpha ~:~ Y-\tilde{\alpha} \ge \sum_{j=1}^{m}\lambda_j
(Y_j-\widetilde{\unP(Y_j)}), m\ge 0, Y_j\in\mathcal{G}, \lambda_j\ge 0,
\alpha\in\R\}.\]
Then $\unE$ is a lower expectation function (that is, $\unE = \lE_{\P}$
for some set $\P$ of probability measures on $W$) and $\unE$ agrees with
$\unP$ on the gambles in $\mathcal{G}$.
\ethm

It follows from Theorem~\ref{thm:natext} that to show that $\unPs$ can
be extended to a lower expectation, it suffices to show that it is
coherent.  This is done in the next lemma.

\lem\label{l:coherent}
The function $\unPs$ is coherent on $\{X_1,\ldots,X_n\}$. 
\elem
\prf
Observe that (\ref{e:sup}) is equivalent to the condition that,
for all nonnegative reals $b_1, \ldots, b_{n-1}$, reals $b_n$, and all $i^*$,
it is \emph{not} the case that
\begin{equation}\label{e:sum}
\sum_{j\ne i^*} b_j (X_j-\tilde{x}_j^*) < b_{i^*} (X_{i^*}-\tilde{x}_{i^*}^*).
\end{equation}
Suppose, by way of contradiction, that
there exist such nonnegative reals $b_1,\ldots,b_{n-1}$ and real $b_n$. 
Note that, without loss of generality, we can 
assume that $i^* \ne n$ and that $b_{i^*} = 1$.  
For if $b_{i^*} \ne 0$, then we can divide
both all coefficients on both sides by $b_{i^*}$ to get an equivalent
instance of (\ref{e:sum}) where $b_{i^*} = 1$.  Moreover,
observe that, since $\gamma_{n-1} = \falsep$ and $\gamma_{n} =
\truep$, we have that $X_{n-1} = \tilde{0}$, $x_{n-1}^* = 0$, $X_{n} =
\tilde{1}$, and $x_{n}^* = 1$.  Thus, if $i^* = n$ or if $b_{i^*} = 0$,
then we can take $i^* = n-1$ and $b_{i^*} = 1$, to get an equivalent
instance of (\ref{e:sum}).  Thus, if there exist $b_1, \ldots, b_{n}$
and $i^*$
such that (\ref{e:sum}) holds, then we can assume without loss of
generality that $i^* \ne n$ and $b_{i^*} = 1$.

For simplicity, in the remainder of the argument, 
we take $i^* = 1$ (the argument is the same for
every choice of $i^* \ne n$) and assume that $b_1 = 1$.
Since $X_{n} = \tilde{1}$ and $x_{n}^* = 1$, there exist 
nonnegative $b_2', \ldots, b_{n-1}'$ and 
$\epsilon>0$ such that  
\[ \sum_{j=2}^{n-1} b_j' (X_j-\tilde{x}_j^*) + \tilde{\epsilon} \le  X_1 -
\tilde{x}_1^*.\] 
Let $b'_{n} = \epsilon +x_1^* - \sum_{j=2}^{n-1} b'_j x^*_j$ ($b'_{n}$ can
be negative). Note that $\sum_{j=2}^{n} b'_j x^*_j = x^*_1 +
\epsilon$, since $x^*_{n}=1$.
Since $X_{n} - \tilde{x}_{n}^* = \tilde{0}$, it follows that
\[ \sum_{j=2}^{n} b_j' (X_j-\tilde{x}_j^*) + \tilde{\epsilon} \le  X_1 -
\tilde{x}_1^*.\] 
Thus, $\sum_{j=2}^{n} b'_j X_j \le X_1$.  By definition of $B_1$,
there exists $\<b_2^*,\ldots,b_{n}^*\> \in B_1$  such that
$\sum_{j=2}^{n} b_j^* X_j \le X_1$ and 
$\sum_{j=2}^{n} b_j^* x^*_j \ge \sum_{j=2}^{n} b'_j x^*_j = x^*_1 +
\epsilon$. In 
other words, we can assume without loss of generality that
$b'_2,\ldots,b'_{n}$ are in $B_1$, since otherwise, we can always
replace them by 
$b_2^*,\ldots,b_{n}^*$. 
By the above, we have that $E(b'_2 X_2+\ldots+b'_{n} X_n)\le E(X_1)$. 
By the constraints corresponding to $f_1$ in $\hat{f}$, we can
therefore derive $b'_2 E(X_2) + \cdots + b'_{n-1} E(X_{n-1}) + 
b'_{n} \le E(X_1)$. (The constraints corresponding to $f_1$ where
chosen so that exactly this derivation could be performed.)
However, replacing $b'_{n}$ by $\epsilon +x^*_1 - \sum_{j=2}^{n-1}
b'_j x^*_j$  and replacing $E(X_i)$ by $x^*_i$, we get that $\epsilon < 0$, a
contradiction. Thus, there cannot exist $b_1,\ldots,b_{n}$ of the
above form.  Hence, $\unPs$ is coherent, as desired.
\eprf
\commentout{
\item[] Case $b_1=0$:\\ 
Equation (\ref{e:sum}) becomes
\[ \sum_{j=2}^{n} b_j (X_j-x_j^*) < 0,\]
and thus, there exists an $\epsilon>0$ such that 
\[ \sum_{j=2}^{n} b_j (X_j-x_j^*) + \epsilon \le  0.\]
Let $b_{n+1} = \epsilon - \sum_{j=2}^{n} b_j x^*_j$ ($b_{n+1}$ can
be negative). Note that $\sum_{j=2}^{n+1} b_j x^*_j = \epsilon$, since
$x^*_{n+1}=1$.

Then $\sum_{j=2}^{n+1} b_j X_j \le 0$.  By construction, there exists
$\<c_2^*,\ldots,c_{n+1}^*\> \in C_1$  such that $\sum_{j=2}^{n+1} c_j^* X_j \le 0$ and
$\sum_{j=2}^{n+1} c_j^* x^*_j \ge \sum_{j=2}^{n+1} b_j x^*_j = \epsilon$. In
other words, we can assume without loss of generality that
$b_2,\ldots,b_{n+1}$ are in $C_1$, as we can always replace them by
$c_2^*,\ldots,c_{n+1}^*$. Now, by choice of the constraints corresponding
to $f_1$ in $\hat{f}$ , we have that
$E(b_2 X_2 + \cdots + b_{n+1} X_{n+1}) \le b_2 E(X_2) + \cdots + b_{n+1} \le
0$. However, replacing $b_{n+1}$ by $\epsilon - \sum_{j=2}^{n}
b_j x^*_j$  and replacing $E(X_i)$ by $x^*_i$, we get that $\epsilon < 0$, a
contradiction. Thus, there cannot exist $b_1,\ldots,b_{n}$ of the
above form. 
\end{description}

As indicated, the same argument holds for $i=2,\ldots,n$. Hence,
$\unPs$ is coherent.
\eprf
}

Continuing with the proof of Theorem~\ref{t:soundcompllp}, by
Lemma~\ref{l:coherent} and the Natural Extension Theorem, there exists
a set of probability measures $\cP$ such that $\lE_{\cP}$ is defined
on all gambles, and $\lE_{\cP}$ agrees with $\unPs$ on
$\{X_1,\ldots,X_{n}\}$. It is routine to check that $(W,\cP,\pi)\sat
f'$, and hence $(W,\cP,\pi)\sat f$, as required.
\eprf

\othm{t:soundcomplbel} 
$\AXBel$ is a sound and complete axiomatization
of $\LE$ with respect to $\MBel$.  
\eothm 
\prf 
Soundness is straightforward. For completeness, we proceed as in
Theorem~\ref{t:soundcomplprob}. Assume without loss of generality that
$f$ is a conjunction of expectation inequalities and their
negations. Using axioms \axiom{E7},
\axiom{E8}, and \axiom{E10} we can convert 
$f$ into the equivalent formula $f^{T_2} \in \LEp$
(Lemma~\ref{l:transf2}) where $e$ is applied only to 
propositional formulas.  Every propositional formula $\phi$ in
$f^{T_2}$ is provably equivalent to a region $\rho_i$ for some $1\leq
i\leq 2^{2^N}$. 
Since $\phi=\rho_i$ is valid,
$e(\phi)=e(\rho_i)$ is provable by \axiom{E5}. This means that we can
find a formula $f'$ provably equivalent to $f^{T_2}$, where $e$ is
applied only to formulas $\rho_1,\ldots,\rho_{2^{2^N}}$. Let $f''$ be
obtained from $f'$ by adding as conjuncts to $f'$ 
all the instances of \axiom{E9} involving the regions
$\rho_1,\ldots,\rho_{2^{2^N}}$, 
as well as the inequalities $e(\rho_i)\ge 0$
for all $1\le i\le 2^{2^N}$, $e(\rho_1)\ge 1$, $-e(\rho_1)\ge -1$ and
$-e(\rho_{2^{2^N}})\ge 0$. (Recall that by assumption, $\rho_1$ is
$\truep$ and $\rho_{2^{2^N}}$ is $\falsep$.) 
It is not hard to see that these formulas are provable, hence $f''$ is
provably equivalent to $f'$, and hence to $f$. 

As before, the negation of an expectation inequality $a_1 e(\gamma_1)
+ \cdots + a_n e(\gamma_n)\geq b$ can be written $- a_1 e(\gamma_1) -
\cdots - a_n e(\gamma_n) > -b$. Thus, without loss of generality, we
can assume that $f''$ is the conjunction of formulas of the form $a_1
e(\rho_1) + \cdots + a_{2^{2^N}} e(\rho_{2^{2^N}})\geq b$ and $-a_1'
e(\rho_1) - \cdots - a_{2^{2^N}}' e(\rho_{2^{2^N}}) > b'$. Let
$\hat{f}$ be the system of inequations obtained by replacing the terms
$e(\rho_i)$ by the variable $x_i$, for $1\leq i\leq 2^{2^N}$. 
We claim that $f''$ in satisfiable in
$\MBel$ iff the system $\hat{f}$ of equations has a solution.
One direction is straightforward. Suppose that $f''$ is
satisfiable. Thus, there exists a belief structure
$(W,\Bel,\pi)$ such that $(W,\Bel,\pi)\sat f''$.  Clearly, taking
$x_i = E_\Bel(\gambM{\rho_i}) = \Bel(\rho_i)$ gives a solution
to $\hat{f}$. Conversely, suppose that $\hat{f}$ has a
solution, say $x_1^*,\ldots,x_{2^{2^N}}^*$. Let $p_1,\ldots,p_N$ be
the primitive propositions appearing in $f$ (and, hence, in $f''$). 
Define the belief structure $M = (W,\Bel,\pi)$ as follows:
\begin{itemize}
\item $W = \{\delta_1,\ldots,\delta_{2^N}\}$, the set of atoms over $p_1,
\ldots, p_N$; 
\item $\Bel(\{\delta_{i_1},\ldots,\delta_{i_k}\})=x_i^*$, where $i$ is the
unique index such that $\delta_{i_1}\lor\ldots\lor\delta_{i_k}$ is
logically equivalent to 
the region $\rho_i$;
\item  $\pi(\delta_i)(p) = \true$ if
and only if $\delta_i\rimp p$ is a propositional tautology.
\end{itemize}
It is straightforward to show that 
$\Bel$ defined in this way is a
belief function. 
That $\Bel$
satisfies B1 and B2 follows from the fact that 
we must have $x_1^* = 0$ and $x_{2^{2^N}}^* = 1$.
Similarly, that $\Bel$ 
satisfies B3 follows from the 
observation that
any instance of B3 corresponds to an instance of \axiom{E9} that was added
as an inequality in $\hat{f}$, and thus is
satisfied by the solution
$x_1^*,\ldots,x_{2^{2^N}}^*$. 
Finally, that $M\sat f''$ holds also follows from the
construction of $\hat{f}$. Thus, $f''$ is satisfiable.
This completes the proof of the claim.

Returning to the proof of the theorem, we want to show that if $f$ is
consistent, then $f$ is satisfiable.  So suppose that $f$ is consistent
and, by way of contradicition, that it is not satisfiable.  Then $f''$
is not satisfiable.  Thus, by the claim, the system of equations
$\hat{f}$ has no solution.  Thus, $\neg f''$ is an instance of the axiom
\axiom{Ineq}. Since $f''$ is provably equivalent to $f$, it follows
that $\neg f$ is provable, so that $f$ is inconsistent, a
contradiction. 
\eprf

\othm{t:soundcomplposs}
$\AXPoss$ is a sound and complete axiomatization of $\LE$ with
respect to $\MPoss$.
\eothm
\prf
Soundness is straightforward. For completeness, we proceed almost
exactly as in Theorem~\ref{t:soundcomplbel}. 
\commentout{
Assume without loss of
generality that $f$ is a conjunction of expectation inequalities and
their negations. Using axioms \axiom{E7}, \axiom{E8}, and \axiom{E10}
we can convert $f$ into the equivalent formula $f^{T_2}$
(Lemma~\ref{l:transf2}) where $e$ is applied only to propositional
formulas.  Every propositional formula $\phi$ in $f^{T_2}$ is provably
equivalent to a formula $\rho_i$ for some $1\leq i\leq 2^{2^N}$. Since
$\phi=\rho_i$ is a valid formula about propositional gamble inequality
(see Section~\ref{s:gambleineq}), \axiom{E5} yields that
$e(\phi)=e(\rho_i)$. This means that we can find a formula $f'$
provably equivalent to $f^{T_2}$, where $e$ is applied only to
formulas $\rho_1,\ldots,\rho_{2^{2^N}}$. Let $f''$ be obtained from
$f'$ by adding as conjuncts to $f'$ all of the expectation
inequalities corresponding to \axiom{E11} on the propositional
formulas $\rho_1,\ldots,\rho_{2^{2^N}}$, as well as the inequalities
$e(\rho_i)\ge 0$ for all $1\le i\le 2^{2^N}$, $e(\rho_1)\ge 1$,
$-e(\rho_1)\ge -1$ and $-e(\rho_{2^{2^N}})\ge 0$. (Recall that by
assumption, $\rho_1$ is $\truep$ and $\rho_{2^{2^N}}$ is $\falsep$.)
It is not hard to see that these formulas are provable, hence $f''$ is
provably equivalent to $f'$, and hence to $f$. We therefore only have
to show that $f''$ is satisfiable.

As before, the negation of an expectation inequality $a_1 e(\gamma_1)
+ \cdots + a_n e(\gamma_n)\geq b$ can be written $- a_1 e(\gamma_1) -
\cdots - a_n e(\gamma_n) > -b$. Thus, without loss of generality, we
can assume that $f''$ is the conjunction of formulas of the form $a_1
e(\rho_1) + \cdots + a_{2^{2^N}} e(\rho_{2^{2^N}})\geq b$ and $-a_1'
e(\rho_1) - \cdots - a_{2^{2^N}}' e(\rho_{2^{2^N}}) > b'$. Let
$\hat{f}$ be the system of inequations obtained by replacing the terms
$e(\rho_i)$ by the variable $x_i$, for $1\leq i\leq 2^{2^N}$. We show
that $f''$ is satisfiable if and only if the system $\hat{f}$ has a
solution. With such a result, because the possibilistic expectation of
a propositional formula amount to the possibility of the propositional
formula, a solution to this system of inequations can be used to
construct a possibility measure on atoms. By way of contradiction,
assume that $f''$ is unsatisfiable. Then the set of corresponding
inequations is unsatisfiable. So, $\neg f''$ is an instance of the
axiom \axiom{Ineq}. Since $f''$ is provably equivalent to $f$, it
follows that $\neg f$ is provable, so that $f$ is inconsistent, a
contradiction. Thus, $f''$, and hence $f$, must be satisfiable.

It remains to show that $f''$ is satisfiable if and only if $\hat{f}$
has a solution. One direction is straightforward. Assume $f''$ is
satisfiable. In other words, there exists a belief structure
$(W,\Poss,\pi)$ such that $(W,\Poss,\pi)\sat f''$. It should be clear
that $E_\Poss(\gambM{\rho_1}),\ldots,E_\Poss(\gambM{\rho_{2^{2^N}}})$
forms a solution to $\hat{f}$. Conversely, assume $\hat{f}$ has a
solution, say $x_1^*,\ldots,x_{2^{2^N}}^*$. Let $p_1,\ldots,p_N$ be
the primitive propositions appearing in $f$ (and hence, in $f''$). Let
$W$ be the set of all atoms over $p_1,\ldots,p_N$ in the order in
which they were considered for the construction of $\hat{f}$,
$\delta_1,\ldots,\delta_{2^N}$. Define
$\Poss(\{\delta_{i_1},\ldots,\delta_{i_k}\})=x_i^*$, where $i$ is the
unique index such that $\delta_{i_1}\lor\ldots\lor\delta_{i_k}$ is
logically equivalent to $\rho_i$.  (Let us denote the $x_i^*$
corresponding to $S$ by $x_S^*$.) Define $\pi(\delta_i)(p) = \true$ if
and only if $\delta_i\rimp p$ is a propositional tautology. Let
$M=(W,\Poss,\pi)$. We first verify that $\Poss$ defined in this way is
a possibility measure. By definition, $\Poss(S)=x_S^*$, and by the
construction of $\hat{f}$, $x_S^*$ is nonnegative. That $\Poss$
satisfies Poss1 and Poss2 follows from the fact that $x_\emptyset^*=0$
and $x_W^*=1$, again by construction of $\hat{f}$. Similarly, that
$\Poss$ satisfies Poss3 follows from the construction of
$\hat{f}$. Intuitively, for any instance of Poss3, there corresponds
an inequality in $\hat{f}$ that is satisfied by the solution
$x_1^*,\ldots,x_{2^{2^N}}^*$. Therefore, $\Poss$ is a possibility
measure. Finally, that $M\sat f''$ holds also follows from the
construction of $\hat{f}$. Thus, $f''$ is satisfiable.}
As before, we can reduce to showing that a formula $f$ consistent with
$\AXPoss$ that is a conjunction of formulas of the form $a_1
e(\rho_1) + \cdots + a_{2^{2^N}} e(\rho_{2^{2^N}})\geq b$ and $-a_1'
e(\rho_1) - \cdots - a_{2^{2^N}}' e(\rho_{2^{2^N}}) > b'$ is
satisfiable.  We add as conjuncts to this formula
all the expectation
inequalities corresponding to \axiom{E11} involving the regions
$\rho_1,\ldots,\rho_{2^{2^N}}$, as well as the inequalities
$e(\rho_i)\ge 0$ for all $1\le i\le 2^{2^N}$, $e(\rho_1)\ge 1$,
$-e(\rho_1)\ge -1$ and $-e(\rho_{2^{2^N}})\ge 0$.  Again, let $\hat{f}$
be the system of inequations formed by replacing $e(\rho_i)$ by $x_i$.
Arguments similar in spirit to those of Theorem~\ref{t:soundcomplbel}
can be used to show that $\hat{f}$  has a solution iff there is a
structure in $\MPoss$ satisfying $f$; we leave details to the reader.
The proof is completed using \axiom{Ineq}, just as the earlier
completeness proofs.
\eprf

\subsection{Proofs for Section~\protect{\ref{s:decision}}}\label{appa6}
As in FHM and HP, the small model theorems we prove rely on the
following lemma, which can be derived from Cramer's rule
\cite{Shores99} and simple estimates on the size of the determinant
(see also \citeN{Chv} for a simpler variant): \lem
\label{l:chv}
If a system of $r$ linear equalities and/or inequalities with integer
coefficients, each of length at most $l$, has a nonnegative solution,
then it has a nonnegative solution with at most $r$ positive entries, 
where the size of each element of the solution is
$O(rl+r\log(r))$. 
\elem

Before getting to the small model theorems, we first establish a
\emph{finite model} theorem, that is, we show that,
for all the representations of uncertainty we consider in this paper,
if a formula in
$\LE$ is satisfiable, it is in fact satisfiable in a structure with
finitely many states. 
This is a consequence of the
completeness proofs in Section~\ref{sec:axiomatization}. 
\lem\label{l:finitemodel}
Suppose $f\in\LE$ is satisfied in some structure in $\MProb$ (resp.,
$\MLP$, $\MBel$, $\MPoss$). Then $f$ is satisfied in a structure in
$\MProb$ (resp., $\MLP$, $\MBel$, $\MPoss$) with finitely many states.
\elem
\prf
\commentout{
This results directly falls out of the particular proofs of
completeness of the axiomatizations $\AXProb$, $\AXLP$, $\AXBel$, and
$\AXPoss$. We first show how the result is derived for probabilities. 

Assume $f$ is satisfiable in $\MProb$. We know $f$ is provably
equivalent to its disjunctive normal form, and if $f$ is satisfiable,
one of the disjunct of this disjunctive normal is satisfiable. Thus,
without loss of generality, assume $f$ is a conjunction of
inequalities formulas and their negations. Consider the proof of
Theorem~\ref{t:soundcomplprob}. In the course of the proof, we derive,
corresponding to a formula $f$ that is a conjunction of inequality
formulas and their negation, a system $\hat{f}$ of inequations, with
the property that $\hat{f}$ has a solution if and only if $f$ is
satisfiable. Let $x_1^*,\ldots,x_{2^N}^*$ be a solution of
$\hat{f}$. Let $p_1,\ldots,p_N$ be the primitive propositions
appearing in $f$, and let $W=\{\delta_1,\ldots,\delta_{2^N}\}$ be the
atoms over $p_1,\ldots,p_N$, in the order considered during the
construction of $\hat{f}$. We define a probability measure $\mu$ over
$W$ by taking $\mu(\delta_i)=x_i^*$, and extending to sets of worlds
by additivity. Let $\pi(\delta_i)(p)=\true$ if and only if
$\delta_i\rimp p$ is a propositional tautology. Let
$M=(W,\mu,\pi)$. It is easy to check that $M\sat f$ (by construction
of $\hat{f}$), and moreover, it is clear that $W$ is finite. 

A similar argument can be applied to $\MLP$, $\MBel$, and
$\MPoss$. Indeed, all the proofs of completeness rely on constructing
a finite system of inequalities over finitely many variables that has
a solution if and only if a particular formula is satisfiable. This
system of inequalities can be used to construct a finite model. In
particular, the structures constructed at the end of the proofs of
Theorems~\ref{t:soundcomplbel} and \ref{t:soundcomplposs} are finite.}
In each of the proofs of
Theorems~\ref{t:soundcomplprob}--\ref{t:soundcomplposs}, we show that a  
formula is satisfiable iff a certain system of inequations has a
solution.  The system of inequations involves only finitely many
variables.  Our argument showing that if the system of inequations is
satisfiable, then there is a structure where the formula is satisfied
actually shows that the satisfying structure has only finitely many
states, with no more than one state per variable in the system.
\eprf

Note 
for future reference
that the satisfying structures in $\MProb$ and $\MLP$ constructed
in the proofs of Theorems~\ref{t:soundcomplprob} and
\ref{t:soundcompllp} are such that all subsets are measurable. 

\othm{t:smallmodel1}
Suppose that $f \in \LE$ is satisfied in some structure in $\MLP$.  Then
$f$ is satisfied in a structure $(\Worlds,\P,\pi)$ such that  $|\Worlds|
\le |f|^2$, $|\P|\leq|f|$, $\mu(w)$ is a rational number such that
$||\mu(w)||$ is $O(|f|^2||f||+|f|^2\log(|f|))$ for every world $w \in
\Worlds$ and $\mu \in \P$, and $\pi(w)(p) = \false$ for every world $w
\in \Worlds$ and every primitive proposition $p$ not appearing in $f$.
\eothm

\prf 
The first step in the proof involves showing that if $\cP$ is a
set of probability measures defined on a finite space $\Worlds$
(assuming all subsets are measurable), and if $X_1,\ldots,X_n$ are
gambles over $\Worlds$, then we can assume without loss of generality
that for each gamble $X_i$, there is a probability measure $\mu_{X_i}
\in \cP$ such that $E_{\mu_{X_i}}(X_i) = \lE_{\cP}(X_i)$ (rather than
$\lE_{\cP}(X_i)$ just being the inf of $E_\mu(X_i)$ for $\mu \in \cP$).
A similar result is proved in HP for upper probabilities.  More
specifically, in HP, it is
shown that, given $\P$, there exists a set $\P' \supseteq \P$ such that,
for each $S \subseteq W$, $\lE_{\P}(X_S) = \lE_{\P'}(X_S)$ and,
moreover, there exists some $\mu_S \in \P'$ such that $\mu_S(S) =
\lE_{\P}(X_S)$.  The next result can be viewed as a generalization of
this result to arbitrary gambles.  It is proved using essentially the
same technique as the corresponding result in HP.

\lem\label{achievemax}
Let $\cP$ be a set of probability measures defined on a finite set $W$ of
worlds, and let $X_1,\ldots,X_n$ be gambles over
$\Worlds$. Then there exists a set $\cP'$ of  probability measures
such that, for each gamble $X_i$, (a) $\lE_{\cP}(X_i) = 
\lE_{\cP'}(X_i)$, and (b) there is a probability measure $\mu_{X_i}
\in \cP'$ such that $E_{\mu_{X_i}}(X_i) = \lE_{\cP}(X_i)$.  
\elem

\prf To show that $\cP'$ exists,
it clearly suffices to show that, for each gamble
$X\in\{X_1,\ldots,X_n\}$, there is a probability measure $\mu_X$
such that 
$E_{\mu_X}(Y) \ge \lE_{\cP'}(Y)$ for all $Y\in\{X_1,\ldots,X_n\}$
and $E_{\mu_X}(X) = \lE_{\cP}(X)$.

Given a gamble $X\in\{X_1,\ldots,X_n\}$, if there exists $\mu \in \cP$
such that $E_{\mu}(X) = \lE_{\cP}(X)$, then we are done.  
\commentout{
Otherwise, we
construct a sequence $\mu_1, \mu_2, \ldots$ of probability measures in
$\cP$ 
that converges to a probability measure $\mu_X$ such that
$E_{\mu_X}(X)=\lE_{\cP}(X)$, and $E_{\mu_X}(Y)\ge\lE_{\cP}(Y)$ for
$Y\in\{X_1,\ldots,X_n\}$. 
Let $N$ be the size of $W=\{w_1,\ldots,w_N\}$, $U_1,\ldots,U_{2^N}$ be
an enumeration of the subsets of $W$, and 
let $Y_1, \ldots, Y_{n+2^N}$ be an enumeration of the gambles
$\{X_1,\ldots,X_n,X_{U_1},\ldots,X_{U_{2^N}}\}$, where $X_{U_i}$ is
the indicator gamble for the set $U_i$, and where $Y_1 = X$.  
We inductively construct a sequence of measures $\mu_{m1}, \mu_{m2},
\ldots$ in $\cP$ for 
$1\le m \le n+2^N$ 
such that $E_{\mu_{mi}}(Y_j)$ converges
to a value $v_{Y_j}\ge\lE_{\cP}(Y_j)$ for $j \le m$ and $\lim_{i
\rightarrow \infty} 
E_{\mu_{mi}}(X) = \lE_{\cP}(X)$.  For $m=1$, we know 
by the definition of $\lE_{\cP}(X)$ that 
there must be a sequence $\mu_{11},
\mu_{12}, \ldots$ of measures in $\cP$ such that 
$E_{\mu_{1i}}(X)$ converges to $\lE_{\cP}(X)$.  For the inductive
step, if 
$m < n+2^N$, suppose 
that
we have constructed an appropriate sequence
$\mu_{m1}, \mu_{m2}, \ldots$.  Consider the sequence of real numbers
$E_{\mu_{mi}}(Y_{m+1})$. 
Using the Bolzano-Weierstrass theorem
\cite{Rudin76} (which says that every sequence of real numbers has a
convergent subsequence), this sequence  has a convergent subsequence.  Let
$\mu_{(m+1)1}, \mu_{(m+1)2}, \ldots$ be the subsequence of $\mu_{m1},
\mu_{m2}, \ldots$ which generates this convergent subsequence.
This sequence of probability measures clearly has all the required
properties. 
In particular, $E_{\mu_{(m+1)i}}(Y_{m+1})$ converges to a value
$v_{Y_{m+1}}\ge\lE_{\cP}(Y_{m+1})$, since for each $i$, 
$E_{\mu_{(m+1)i}}(Y_{m+1})\ge\lE_{\cP}(Y_{m+1})$. 
This completes the inductive step. 

We can now define the desired probability measure $\mu_X$.  For a
subset $U\subseteq W$, let $\mu_X(U) = \lim_{i \rightarrow
\infty}\mu_{(n+2^N)i}(X_U)$. It is easy to check that $\mu_X$ is
indeed a probability measure. It also easy to check that for each
$Y\in\{X_1,\ldots,X_n\}$, we have
$E_{\mu_X}(Y)=\lim_{i\rightarrow\infty}
E_{\mu_{(n+2^N)i}}(Y)$. Thus,
$E_{\mu_X}(X)=\lim_{i\rightarrow\infty}E_{\mu_{(n+2^N)i}}(X)=\lE_{\cP}(X)$,
by construction. Similarly,
$E_{\mu_X}(Y)=v_Y\ge\lE_{\cP}(Y)$ for all
$Y\in\{X_1,\ldots,X_n\}$. Therefore,
$lE_{\cP}(Y)=\lE_{\cP\cup\{\mu_X\}}(Y)$ for all
$Y\in\{X_1,\ldots,X_n\}$. This shows that an appropriate set $\cP'$
exists, by taking $\cP'=\cP\cup\{\mu_{X_1},\ldots,\mu_{X_n}\}$. 
}
If not, there must be a sequence $\mu_1, \mu_2, \ldots$ of measures in
$\cP$ such that $\lim_{n \rightarrow \infty} E_{\mu_n}(X) =
\lE_{\cP}(X)$.  Suppose that $W = \{w_1, \ldots, w_n\}$.  
By the Bolzano-Weierstrass theorem
\cite{Rudin76} (which says that every sequence of real numbers has a
convergent subsequence), the sequence 
$\mu_1(w_1)$, $\mu_2(w_1)$, $\mu_3(w_1)$, $\ldots$
has a convergent subsequence. Suppose,
inductively, that we have found a subsequence $\mu_{j,1}, \mu_{j,2},
\ldots$ of $\mu_1, \mu_2, \ldots$ such that $\mu_{j,1}(w), \mu_{j,2}(w),
\mu_{j,3}(w), \ldots$ converges for $w \in \{w_1, \ldots, w_j\}$. 
By applying the Bolzano-Weierstrass Theorem again, there is a
subsequence $\mu_{j+1,1}, \mu_{j+2,1}, \ldots$ of 
$\mu_{j,1}, \mu_{j,2}, \ldots$ such that 
$\mu_{j+1,1}(w_{j+1})$, $\mu_{j+2,1}(w_{j+1})$, $\ldots$ converges.
It follows that $\mu_{j+1,1}(w), \mu_{j+2,w}), \ldots$ converges for all
$w \in \{w_1, \ldots, w_{j+1}\}$.  By induction, we can 
find a subsequence 
$\mu_{n,1}, \mu_{n,2}, \ldots$ of the original sequence
such that  $\mu_{n,1}(w), \mu_{n,2}(w), \ldots$ converges for all $w \in
W$.  Suppose that $\mu_{n,1}(w_i), \mu_{n,2}(w_i), \ldots$ converges to
$p_i$.  It is easy to see that $p_1 + \cdots + p_n = 1$, since 
$\sum_{i=1}^n \mu_{n,j}(w_i) = 1$ for all $j$.  Let $\mu_X$ be the
probability measure such that $\mu(w_i) = p_i$.  Since $\mu_{n,j}
\rightarrow \mu_X$, it must be the case that $E_{\mu_X}(X) =
\lE_{\cP}(X)$.  Moreover, since $\mu_{n,j} \in \cP$ for all $j$, 
it must be the case that $E_{\mu_{n,j}}(Y) \ge \lE_{\cP}(Y)$ for all
$j$.  Thus, $E_{\mu_X}(Y) \ge \lE_{\cP}(Y)$, as desired.

Now  let $\cP' = \cP \union \{\mu_{X_1}, \ldots, \mu_{X_n}\}$.  $\cP'$
clearly has the desired properties.
\eprf

Continuing with the proof of Theorem~\ref{t:smallmodel1}, 
suppose that $f$ is satisfiable in $\MLP$.  By 
Lemma~\ref{l:finitemodel}, $f$ is satisfied in 
a lower probability structure with a finite set $W$ of worlds.  Thus, by
Lemma~\ref{achievemax}, $f$ is satisfied in a structure
$M=(W,\cP,\pi)$ such that, for all $X\in\{X_1,\ldots,X_n\}$, there
exists $\mu_X \in \cP$ such that $E_{\mu_X}(X) = \lE_{\cP}(X)$.  

The rest of the proof also continues in much the same spirit as the
proof of the analogous result in HP.
A straightforward induction on structures shows that we can find a
formula equivalent to $f$ in disjunctive normal form, where each
disjunct has length at most $|f|$.  (The inductive hypothesis states
that we can find both a DNF formula equivalent to $f$ where each
disjunct has length at most $|f|$, and a CNF formula equivalent to $f$
where each conjunct has length at most $|f|$.)  Since a formula in DNF
is satisfiable iff one of its disjuncts is, we can thus assume
without loss of generality that $f$ is a conjunction of inequality
formulas and negations of inequality formulas.
Let $p_1,\ldots,p_N$ be the primitive formulas appearing in
$f$. Let $\delta_1,\ldots,\delta_{2^N}$ be the atoms over $p_1,
\ldots, p_N$.  As in the proof of completeness, we derive a system of
equalities and inequalities from $f$, but it is a slightly more
complicated system than that used in the completeness proof. Recall that
each propositional formula 
over $p_1, \ldots, p_N$ is a disjunction of atoms.  Let
$\gamma_1,\ldots,\gamma_k$ be the propositional gambles that appear in
$g$. 
Notice that $k < |f|$ (since there are some symbols in $f$, such as
the coefficients, that are not in the propositional formulas).  The
system of equations and inequalities we construct involve variables
$x_{ij}$, where $i = 1, \ldots, k$ and $j = 1,\ldots, 2^N$.
Intuitively, $x_{ij}$ represents
$\mu_{\gamma_i}(\eventM{\delta_j})$, where
$\mu_{\gamma_i} \in \cP$ is such that
$E_{\mu_{\gamma_i}}(\gambM{\gamma_i}) = \lE_{\cP}(\gambM{\gamma_i})$.
Thus, the system includes $k<|f|$ equations of the form
\[x_{i1} + \cdots + x_{i2^N} = 1,\] 
for $i = 1, \ldots, k$.  
Suppose that $\gamma_i$ is equivalent to
$b_{i1}\delta_1+ \cdots+b_{i2^N}\delta_{2^N}$.    Then 
$b_{i1}x_{i1}+ \cdots+b_{i2^N}x_{i2^N}$ represents
$E_{\mu_{\gamma_i}}(\gambM{\gamma_i})$.  Since
$E_{\mu_{\gamma_i}}(\gambM{\gamma_i}) \le E_{\mu}(\gambM{\gamma_i})$ for
all $\mu \in \cP$, the system 
includes $k2 - k$ inequalities of the form
\[ b_{i1}x_{i1}+ \cdots+b_{i2^N}x_{i2^N} \le b_{i1}x_{i'1}+ \cdots+b_{i2^N}x_{i'2^N}, \]
for each pair $i$, $i'$ such that $i \ne i'$.
For each conjunct in $g$ of the form $a_1 e(\gamma_1)+ \cdots+a_n
e(\gamma_k)\geq b$, there is a corresponding inequality where,
roughly speaking, we replace $e(\gamma_i)$ by
$E_{\mu_{\gamma_i}}(\gambM{\gamma_i})$.%
\footnote{For simplicity here, we are implicitly assuming that each of
the formulas $\gamma_i$ appears in each conjunct of $f$.  This is
without loss of generality, since if $\gamma_i$ does not appear, we can
put it in, taking $ a_i = 0$.}  Since
$E_{\mu_{\gamma_i}}{\gambM{\gamma_i}}$ corresponds to
$b_{i1}x_{i1}+ \cdots b_{i2^N}x_{i2^N}$, the appropriate inequality
is \[\sum_{i=1}^k  a_i(b_{i1}x_{i1}+ \cdots+b_{i2^N}x_{i2^N}) \ge b.\]
Negations of such formulas correspond to a negated inequality formula;
as before, this is equivalent to a formula of the form
\[-(\sum_{i=1}^k  a_i(b_{i1}x_{i1}+ \cdots+b_{i2^N}x_{i2^N}) > -b.\]
Notice that there are at most $|f|$ inequalities corresponding to the
conjuncts of $f$.  Thus, altogether, there are at most $k(k-1) + 2|f|
< |f|^2$ equations and inequalities in the system (since $k < |f|$).
We know that the system has a nonnegative solution (taking $x_j^i$ to
be $\mu_{\gamma_i}(\eventM{\delta_j})$).  It follows from Lemma
\ref{l:chv} that the system has a solution
$x^*=(x^*_{11},\ldots,x^*_{12^N},\ldots,x^*_{k1},\ldots,x^*_{k2^N})$
with $t\leq |f|^2$ entries positive, and with each entry of size
$O(|f|^2||f||+|f|^2\log(|f|))$.

We use this solution to construct a small structure satisfying the
formula $f$. Let $I=\{i \sep x^*_{ij}$ is positive, for some
$j\}$; suppose that $I = \{i_1, \ldots, i_{t'}\}$ for some $t' \le
t$.  Let $M=(W,\cP,\pi)$, where $W$ has $t'$ worlds, say
$s_1,\ldots,s_{t'}$. Let $\pi(s_h)$ be the truth assignment
corresponding to the formula $\delta_{i_h}$, that is, $\pi(s_h)(p) =
\true$ if and only if $\delta_{i_h}\rimp p$ is a propositional
tautology (and where $\pi(s_h)(p)=\false$ if $p$ does not appear in
$f$). Define $\cP=\{\mu_j\sep 1\leq i\leq k\}$, where $\mu_j(s_h) =
x^*_{i_hj}$.  It is clear from the construction that $M\models
f$. Since $|\cP|=k < |f|$, $|W|=t'\leq t\leq |f|^2$ and $\mu_j(s_h) =
x^*_{i_hj}$, where, by construction, the size of $x^*_{i_hj}$ is
$O(|f|^2||f||+|f|^2\log(|f|))$, the theorem follows.
\eprf

\othm{t:smallmodel2}
Suppose that $f \in \LE$ is satisfied in some structure in $\MBel$
(resp., $\MPoss$). Then $f$ is satisfied in a structure
$(\Worlds,\nu,\pi)$ such that $|\Worlds|\le |f|^2$, $\nu$ is a belief
function (resp., possibility measure) whose corresponding mass
function is positive on at most $|f|$ subsets of $\Worlds$ and the mass
of each of these $|f|$  is a rational number of size 
$O(|f|\,||f||+|f|\log(|f|))$, 
and $\pi(w)(p) = \false$ for every world $w \in \Worlds$ and every
primitive proposition $p$ not appearing in $f$.  
\eothm
\prf
First, consider expectation for belief functions.  
The proof is similar in spirit to the proof of the small model theorem
for reasoning about belief functions given in FHM.  In FHM, the
complexity result used the representation of belief functions in terms
of mass functions.  We do that here too; to do so, it is helpful to have
an alternate characterization of expectation for belief that uses mass
functions. 

Given a belief function $\Bel$
over a finite set $W$ of worlds, let $m$ be the corresponding mass
function. For a given gamble $X$, let 
$v_U = \min_{w\in U} X(w)$. 
\lem\label{l:altbelief-exp}
$E_\Bel(X)=\sum_{U\subseteq W} m(U)v_U$.
\elem
\prf
Suppose that $\Bel$ is a belief function on $W$ with corresponding mass
function $m$, and $X$ is a gamble on a finite set $W = 
\{w_1, \ldots, w_n\}$, where the elements of $W$ are ordered so 
that $i \le j$ implies $X(w_i) \le X(w_j)$.
For each $U \subseteq W$, let $i_U = \min\{i: w_i \in U\}$; note that
$v_U = X(w_{i_U})$.  
$$\lE_{\Bel}(X) = \sum_{U \subseteq W} m(U) v_U.$$
We want to show that  $\lE_{\Bel}(X) = E_{\Bel}(X)$.

Recall from Section~\ref{s:belief-exp} that $E_{\Bel} =
\lE_{\P_{\Bel}}$, where 
$\P_{\Bel}$ is the set $\{\mu:
\mu(U) \ge \Bel(U) \mbox{ for all } U \subseteq W\}$.
As a first step, let $\mu_0$ be a probability measure on $W$ such that
$\mu_0(X=x) = \sum_{\{U: v_U = x\}}m(U)$.  (Note that $\mu_0$ is a
probability measure, since $$\sum_{x \in \V(X)} \mu_0{X=x} = \sum_{x \in
\V(X)} \sum_{\{U: v_U = x\}}m(U) = \sum_{U \subseteq W} m(U) = 1,$$
since $m$ is a mass function.)  
Clearly 
\begin{align*}
E_{\mu_0}(X) & = \sum_{x \in \V(X)} x \mu_0(X=x) = \sum_{x \in
\V(X)} x \sum_{\{U: v_U = x\}}m(U) \\
& = \sum_{U \subseteq W} m(U)v_U =
\lE_{\Bel}(X).
\end{align*}
Moreover, for all $U \subseteq W$, since $\mu_0(U) = \sum_{\{V: w_{i_V}
\in U\}} m(V)$, and $w_{i_V} \in U$ for all $V \subseteq U$, it follows
that $\mu_0(U) \ge \sum_{\{V: w_{i_V} \in U\}} m(V) = \Bel(U)$.
It follows that $\mu_0 \in \P_{\Bel}$.  Thus, $$\lE_{\Bel}(X) =
E_{\mu_0}(X) \ge \inf_{\mu \in \P_{\Bel}} E_{\mu}(X) =
E_{\P_{\Bel}}(X) = E_{\Bel}(X).$$

\commentout{
\P_{\Bel}$ such that $E_{\mu_0}(X) 
= \lE_{\Bel}(X)$.  That will show that $\lE_{\P_\Bel}(X) \le
\lE_{\Bel}(X)$.
It clearly suffices to construct a probability measure $\mu_0 \in
\P_{\Bel}$ such that
\begin{equation}\label{eq1}
\mu_0(X=x) = \sum_{\{U: v_U = x\}}m(U)
\end{equation}
(since $E_{\mu_0}(X) = \lE_{\Bel}(X)$ by definition).
Define $\mu_0(w_i) = \sum_{\{U: w_i = w_{i_U}\}} m(U)$.
It is easy to see that $\mu_0$ satisfies (\ref{eq1}).  To
see that $\mu_0 \in \P_{\Bel}$, consider any set $U$.  We must show that
\begin{equation}\label{eq2}
\mu_0(U) \ge \Bel(U) = \sum_{V \subseteq U}m(V).
\end{equation}
But it easily
follows from the definition that $\mu_0(U) = \sum_{\{V: w_{i_V} \in U\}}
m(V)$ and for each $V \subseteq U$, we must have $w_{i_V} \in U$;
thus (\ref{eq2}) follows.
}

To show that $\lE_{\P_\Bel}(X) \ge \lE_{\Bel}(X)$,
let $x_1 \le \cdots \le x_k$ be the values of $X$ in increasing order,
and let $U_j = \union_{i = 1}^j (X=x_j) = \{w \in W: X(w) \le x_j\}$.  We
claim that $\mu_0(U_j) = 
\Plaus(U_j)$, $j = 1, \ldots, k$.  This is almost immediate from the
definition, since
$$\mu_0(U_j) = \sum_{\{U: v_U \le x_j\}} m(U) =
\sum_{\{U: U \inter U_j \ne \emptyset\}} m(U) = \Plaus(U_j).$$
It follows that if $\mu \in \P_{\Bel}$, then $\mu_0(U_j) \ge \mu(U_j)$,
for $j = 1, \ldots, k$ (since $\mu(U_j) \le \Plaus(U_j)$).  For
convenience, define $U_0 = \emptyset$.
Then we get that
$$\begin{array}{lll}
E_\mu(X) - E_{\mu_0}(X) & = &\sum_{i=i}^k x_i (\mu(X=x_i) - \mu_0(X=x_i))\\
&=&\sum_{i=1}^k x_i((\mu(U_i) - \mu(U_{i-1})) - (\mu_0(U_i) -
\mu_0(U_{i-1})))\\ 
&= &x_k(\mu(U_k) - \mu_0(U_k)) + \sum_{i=1}^{k-1} (x_{i+1}-x_i)(\mu_0(U_i)
- \mu(U_i))\\
&\ge &0.
\end{array}$$
The last inequality follows from the fact that $\mu(U_k) = \mu_0(U_k) =
\mu_0(W) = 1$, $x_{i+1} > x_i$, and $\mu_0(U_i) \ge \mu(U_i)$ for $i = 1, \ldots,
k-1$.  Thus, $E_{\mu_0}(X) = \lE_{\P_\Bel}(X)$.
It follows that $E_{\Bel}(X) = \lE_{\P_\Bel}(X)$.
\eprf

Continuing with the proof of Theorem~\ref{t:smallmodel2}, 
suppose that $f \in \LE$ is satisfiable in structure $M$.
As in the proof of Theorem~\ref{t:smallmodel1}, 
we can assume without loss of generality that $f$ is a conjunction of
expectation formulas and their negations.
Let $\delta_1,\ldots,\delta_{2^N}$ be the atoms
over $p_1, \ldots, p_N$. Let $\rho_1,\ldots,\rho_{2^{2^N}}$ be 
all the inequivalent propositional formulas over $p_1,\ldots,p_N$.

\commentout{
Furthermore, without loss of generality, we can take the set of worlds
in $M$ to be the atoms over $p_1,\ldots,p_N$. (Intuitively, this
ensures that every subset of $W$ corresponds to a different
propositional formula $\rho_i$.) To show this, assume that $M$ is a
model with $M\sat f$. We construct a model $M'=(W',\Bel',\pi')$ with
$M'\sat f$, and where the worlds of $M'$ are the atoms over
$p_1,\ldots,p_N$.  Accordingly, define
$W'=\{\delta_1,\ldots,\delta_{2^N}\}$, and take
$\Bel'(\{\delta_{i_1},\ldots,\delta_{i_k}\})=\Bel(\eventM{\delta_{i_1}\lor\ldots\lor\delta_{i_k}})$.
We define the interpretation $\pi'$ so that $\pi'(\delta_i)(p)=\true$ if
$p\rimp\delta_i$ is a propositional tautology. It is straightforward
to verify that $M'\sat f$. This follows essentially from the fact that
for all gambles $\gamma$ in $f$, $E_\Bel(\gambM{\gamma}) =
E_{\Bel'}(\gamb{\gamma}{M'})$, which we now show. Let $\gamma$ be a
gamble in $f$. The gamble $\gamma$ is provably equivalent to a gamble
of the form $a_1\delta_1+\ldots+a_{2^N}\delta_{2^N}$. Then
$\gambM{\gamma}=a_1 X_{\eventM{\delta_1}}+\ldots+a_{2^N}
X_{\eventM{\delta_{2^N}}}$, and $\gamb{\gamma}{M'}=a_1
X_{\event{\delta_1}{M'}}+\ldots+
a_{2^N}X_{\event{\delta_{2^N}}{M'}}=a_1
X_{\{\delta_1\}}+\ldots+a_{2^N} X_{\{\delta_{2^N}\}}$. Let
$i_1,\ldots,i_{2^N}$ be a permutation of $1,\ldots,2^N$
such that $a_{i_1}<\ldots<a_{i_{2^N}}$. We then have
\[\begin{array}{ll}
& E_\Bel(\gambM{\gamma})\\
= & a_{i_1}+(a_{i_2}-a_{i_1})\Bel(\gambM{\gamma}>a_{i_1})+\ldots+(a_{i_{2^N}}-a_{i_{2^N-1}})\Bel(\gambM{\gamma}>a_{i_{2^N-1}}) \\
= & a_{i_1}+(a_{i_2}-a_{i_1})\Bel(\eventM{\delta_{i_2}}\cup\ldots\cup\eventM{\delta_{i_{2^N}}})+\ldots+(a_{i_{2^N}}-a_{i_{2^N-1}})\Bel(\eventM{\delta_{i_{2^N}}}) \\
= & a_{i_1}+(a_{i_2}-a_{i_1})\Bel(\eventM{\delta_{i_2}\lor\ldots\lor\delta_{i_{2^N}}})+\ldots+(a_{i_{2^N}}-a_{i_{2^N-1}})\Bel(\eventM{\delta_{i_{2^N}}})
\\
= & a_{i_1}+(a_{i_2}-a_{i_1})\Bel'(\{\delta_{i_2},\ldots,\delta_{i_{2^N}}\})+\ldots+(a_{i_{2^N}}-a_{i_{2^N-1}})\Bel(\{\delta_{i_{2^N}}\})\\
= & a_{i_1}+(a_{i_2}-a_{i_1})\Bel'(\gamb{\gamma}{M'}>a_{i_1})+\ldots+(a_{i_{2^N}}-a_{i_{2^N-1}})\Bel(\gamb{\gamma}{M'}>a_{i_{2^N-1}})\\
= & E_{\Bel'}(\gamb{\gamma}{M'}).
\end{array}\]
Therefore, without loss of generality, the worlds in $M$ are the atoms.
}
As
in the 
proof of completeness, we derive a system of equalities and
inequalities from $f$.
First, note that every propositional formula appearing in the gambles
of $f$ is provably equivalent to
some region $\rho_i$. Therefore, we can replace every gamble in $f$ by an
equivalent gamble where the propositional formulas are 
the regions
$\rho_1,\ldots,\rho_{2^{2^N}}$. Let $f'$ be resulting
formula. Clearly, $M\sat f'$. Construct the following system of linear
inequalities over the variables $x_1,\ldots,x_{2^{2^N}}$, 
where, intuitively, 
the variable $x_i$ stands for the mass corresponding to
the set $\rho_i$. For
every gamble $\gamma$ that appears in $f'$ every $\rho_i$, we can
compute values $v_{\gamma,i}=\min_{w\in\eventM{\rho_i}}
(\gambM{\gamma}(w))$.  For every gamble $\gamma$, replace every
term
$e(\gamma)$ in $f'$ by $\sum_{i=1}^{2^{2^N}} x_i
v_{\gamma,i}$. Consider the system 
$\hat{f'}$ 
of at most $|f|$ inequalities (over unknowns $x_1,\ldots,x_{2^{2^N}}$)
resulting from this process, after putting together like terms, along
with the inequalities $x_1+ \cdots+x_{2^{2^N}}\ge 1$ and
$-x_1-\cdots-x_{2^{2^N}}\ge -1$. 

We now show that any nonnegative solution $x^*_1,\ldots,x^*_{2^{2^N}}$
of $\hat{f'}$ can be used to construct a belief function $\Bel^*$ such
that $(W,\Bel^*,\pi)\sat f'$. (Note that we keep the same worlds and
interpretation as in $M$.)  Let $x^*_1,\ldots,x^*_{2^{2^N}}$ be a
nonnegative solution of $\hat{f}$. Define the mass function
$m^*(\eventM{\rho_i})=x^*_i$,
and define $m^*(U) = 0$ for all $U \subseteq W$ such that $U$ is not of
the form $\eventM{\rho}$ for some region $\rho$.
Because
$x^*_1+\ldots+x^*_{2^{2^N}}=1$ (by choice of $\hat{f'}$), $m^*$ is
indeed a mass function.  Let $\Bel^*$ be the belief function
corresponding to $m^*$, and let $M^*=(W,\Bel^*,\pi)$. Since the
interpretation of propositional formulas and gambles depends only on
the set of worlds and the interpretation $\pi$, for all propositional
formulas $\rho_i$, we have $\eventM{\rho_i}=\event{\rho_i}{M^*}$, and
for all gambles $\gamma$ appearing in $f'$, we have
$\gambM{\gamma}=\gamb{\gamma}{M^*}$. We now show that $M^*\sat 
f'$. Let $a_1 e(\gamma_1)+\ldots+a_k e(\gamma_k)\ge\alpha$ be a
conjunct in $f'$. (A similar argument works for the negation of
expectation inequalities.) We know from the construction of $\hat{f'}$ 
and the fact that $x^*_1,\ldots,x^*_{2^{2^N}}$ is a solution of
$\hat{f'}$ that
\[a_1\sum_{i=1}^{2^{2^N}}x^*_iv_{\gamma_1,i}+\ldots+a_k\sum_{i=1}^{2^{2^N}}x^*_i
v_{\gamma_k,i}\ge \alpha.\]
By definition, this is just 
\begin{multline*}
a_1\sum_{i=1}^{2^{2^N}}m^*(\eventM{\rho_i})\min_{w\in\eventM{\rho_i}}(\gambM{\gamma_1}(w))+\ldots+\\
a_k\sum_{i=1}^{2^{2^N}}m^*(\eventM{\rho_i})\min_{w\in\eventM{\rho_i}}(\gambM{\gamma_k}(w))\ge
\alpha.
\end{multline*}
Since every subset of $W$ is of the form $\eventM{\rho_i}$
for some $i$, we have 
\[a_1\sum_{U\subseteq
W}m^*(U)\min_{w\in U}(\gambM{\gamma_1}(w))+\ldots+ a_k\sum_{U\subseteq
W}m^*(U)\min_{w\in U}(\gambM{\gamma_k}(w))\ge \alpha.\] But by
Lemma~\ref{l:altbelief-exp}, this is just $a_1
E_{\Bel^*}(\gambM{\gamma_1})+\ldots+a_k
E_{\Bel^*}(\gambM{\gamma_k})\ge\alpha$, which is equivalent to $a_1
E_{\Bel^*}(\gamb{\gamma_1}{M^*})+\ldots+a_k
E_{\Bel^*}(\gamb{\gamma_k}{M^*})\ge\alpha$, and thus $M^*\sat
a_1 e(\gamma_1)+\ldots+a_k e(\gamma_k)\ge\alpha$. It follows that
$M^*\sat f'$.

Because $f'$ is satisfiable in $M$, there is in fact a
nonnegative solution to the system $\hat{f'}$, where
$x_i = m(\eventM{\rho_i})$ and $m$ is the mass function corresponding
to the belief function in $M$. By Lemma~\ref{l:chv}, there is a small
nonnegative solution $x^*_1,\ldots,x^*_{2^{2^N}}$, that is, 
one with at most $|f|$ positive entries and each entry of size
$O((|f|)||f||+(|f|)\log(|f|))$.
By the argument above, $x^*_1,\ldots,x^*_{2^{2^N}}$ can be used to
construct a model $M^*=(W,\Bel^*,\pi)$ such that $M^*\sat f'$. We are
not quite done yet; while we have a small mass function, we still have
potentially too many worlds. We now show how to cut down the number of
worlds in the model to get a small enough structure $M'$.

Let $\{i_1,\ldots,i_{|f|}\} = \{i: m^*(\event{\rho_i}{M^*}) >0 \}$.
Let $\gamma_1,\ldots,\gamma_k$ be the
gambles in $f'$. For $i \in \{1,\ldots,k\}$ and
$j\in\{1,\ldots,|f|\}$, there is some $w_{i,j} \in
\event{\rho_{i_j}}{M^*}$ such that $\gambM{\gamma_i}(w_{i,j}) =
\min_{w\in\eventM{\rho_{i_j}}}(\gambM{\gamma_i}(w))$; that is,
$w_{i,j}$ is a world where $\gambM{\gamma_i}$ attains its minimum
value.
Define $M'=(W',\Bel',\pi')$ as follows. Let the
set of states $W'$ be $\{w_{i,j} ~:~
i\in\{1,\ldots,k\},j\in\{1,\ldots,|f|\}\}$. 
Let $\Bel'$ be the belief function whose corresponding mass function
$m'$ is defined by setting
$m'(\{w_{1,j},\ldots,w_{k,j}\})=m^*(\event{\rho_{i_j}}{M^*})$ for
$j=1,\ldots,|f|$, and $m'(U)=0$ if $U$ is not 
$\event{\rho_{i_j}}{M^*}$ for some $j \in \{1, \ldots, |f|\}$.
The interpretation $\pi'$
is simply the restriction of $\pi$ to $W'$, with $\pi(w)(p)=\false$
for primitive propositions $p$ not appearing in $f$. This model
satisfies the size conditions of the theorem. 

We now check that $f'$ (and hence $f$) is satisfiable in $M'$. 
It is clearly sufficient to show that for every gamble $\gamma_i$ in
$f'$, we have
$E_{\Bel^*}(\gamb{\gamma_i}{M^*})=E_{\Bel'}(\gamb{\gamma_i}{M'})$. To
do this, we show that
\begin{equation}\label{e:mineq}
\gamb{\gamma_i}{M^*}(w_{i,j}) = \min_{1\le r\le
k}(\gamb{\gamma_i}{M'}(w_{r,j})).
\end{equation}
By choice of $w_{i,j}$, $\gamb{\gamma_i}{M^*}(w_{i,j}) =
\min_{w\in\event{\rho_{i_j}}{M^*}}(\gamb{\gamma_i}{M^*}(w))$. Since
$\{w_{1,j},\ldots,w_{k,j}\}$ is a subset of $\event{\rho_{i_j}}{M^*}$, 
it follows that 
$\gamb{\gamma_i}{M^*}(w_{i,j}) = \min_{1\le r\le
k}(\gamb{\gamma_i}{M^*}(w_{r,j}))$. 
Next note that for all $w\in\{w_{1,j},\ldots,w_{k,j}\}$, we have that
$\gamb{\gamma_i}{M^*}(w)=\gamb{\gamma_i}{M'}(w)$.
For if $\gamma_i = 
c_1\rho_{j_1}+\ldots+c_l\rho_{j_l}$, then $\gamb{\gamma_i}{M^*}=c_1
X_{\event{\rho_{j_1}}{M^*}}+\ldots+c_l X_{\event{\rho_{j_l}}{M^*}}$
 and $\gamb{\gamma_i}{M'}=c_1 X_{\event{\rho_{j_1}}{M'}}+\ldots+c_l
X_{\event{\rho_{j_l}}{M'}}$. 
Since $\pi'$ is the restriction of $\pi^*$ to $W'$, it follows that
$\event{\rho_{j_i}}{M^*} = \event{\rho_{j_i}}{M'} \inter W'$.  Since
$\{w_{1,j},\ldots,w_{k,j}\} \subseteq W'$, it follows that
$X_{\event{\rho_{j_i}}{M^*}}(w) =  X_{\event{\rho_{j_1}}{M'}}(w)$ for all
$w \in \{w_{1,j},\ldots,w_{k,j}\}$.  Thus,
$\gamb{\gamma_i}{M^*}(w)=\gamb{\gamma_i}{M'}(w)$ for all 
$w \in \{w_{1,j},\ldots,w_{k,j}\}$.  
We can now show that
$E_{\Bel^*}(\gamb{\gamma_i}{M^*})=E_{\Bel'}(\gamb{\gamma_i}{M'})$:
\[\begin{array}{ll}
  & E_{\Bel^*}(\gamb{\gamma_i}{M^*})\\
= & \sum_{i=1}^{2^{2^N}}m^*(\event{\rho_i}{M^*})\min_{w\in\event{\rho_i}{M^*}}(\gamb{\gamma_i}{M^*}(w))\\
= & \sum_{j=1}^{|f|}
m^*(\event{\rho_{i_j}}{M^*})\min_{w\in\event{\rho_i}{M^*}}(\gamb{\gamma_i}{M^*}(w))
  \quad\mbox{(since $m^*(U) = 0$ if $U \ne \event{\rho_{i_j}}{M^*}$)}\\
= & \sum_{j=1}^{|f|}
m^*(\event{\rho_{i_j}}{M^*})\gamb{\gamma_i}{M^*}(w_{i,j})
  \quad\mbox{(by choice of $w_{i,j}$)}\\
= & \sum_{j=1}^{|f|}
m'(\{w_{1,j},\ldots,w_{k,j}\})\gamb{\gamma_i}{M^*}(w_{i,j})\\
= & \sum_{j=1}^{|f|}
m'(\{w_{1,j},\ldots,w_{k,j}\})\min_{1\le r\le
k}(\gamb{\gamma_i}{M'}(w_{r,j})) \quad\mbox{(by (\ref{e:mineq}))}\\
= & \sum_{U\subseteq W'}
m'(U)\min{w\in U}(\gamb{\gamma_i}{M'}(w)) \quad\mbox{(adding sets for
which $m'(U)=0$)}\\
= & E_{\Bel'}(\gamb{\gamma_i}{M'}).
  \end{array}\]
{F}rom this result, it is easy to see that $M'\sat f'$, and hence
$M'\sat f$. 
This establishes the small-model
result for $\LE$ interpreted over belief functions.

Essentially the same argument works in the case of possibility measures.
Recall that a possibility measure 
$\Poss$
is just a plausibility function. 
whose corresponding  
mass function $m$ 
is
\emph{consonant}; that is, for all sets $U,V$
such that $m(U)>0$ and $m(V)>0$, we have either $U\subseteq V$ or
$V\subseteq U$ \cite{DuboisPrade82}. In other words, the sets of
positive mass $U_1,\ldots,U_k$ can be ordered such that
$U_1\subseteq\ldots\subseteq U_k$. 
We then proceed much as for
belief functions. 
We construct a system of inequalities
over the variables $x_{1},\ldots,x_{k}$, where, 
Using Lemma~\ref{l:chv} again, because $f'$ is satisfiable in $M$, there
is a small nonnegative solution 
$x^*_{1},\ldots,x^*_{k}$, with at most $|f|$ positive
entries, each of small size. We can then use this solution to build a
structure satisfying $f$ where the mass function is consonant and
positive on at 
most $|f|$ sets. 
This follows directly from the 
fact that the
only sets of positive mass will be among 
$U_1,\ldots,U_k$, which are already
such that $U_1\subseteq\ldots\subseteq U_k$.) The remainder of the proof
goes through as in the belief function case.
\eprf

\othm{t:decproc3}
The problem of deciding whether a formula in $\LE$ is
satisfiable 
in $\MProb$ (\respc $\MLP$, $\MBel$, $\MPoss$) is NP-complete.
\eothm
\prf
The result for $\MProb$ is immediate from the proof that the
satisfiability problem for $\LPP_1$ (the restriction of $\LPP$ to
rational coefficents) is NP-complete, together with the 
argument in Theorem~\ref{thm:expressive} showing that every formula in
$\LE$ is equivalent to a formula in  $\LPP_1$ of the same length
(where the formula is essentially given by the translation of 
Lemma~\ref{l:transf1}).

Consider $\MLP$. The lower bound follows from the fact that we can
reduce propositional satisfiability to the decision problem for
$\LE$; hence the problem is NP-hard
(by replacing each proposition $p$ by the formula $e(p) = 1$).
The upper bound follows from
Theorem~\ref{t:smallmodel1}. Given a formula $f$, first guess a small
model $M=(W,\cP,\pi)$, of the form guaranteed to exist by
Theorem~\ref{t:smallmodel1}. (The fact that $\pi(w)(p)=\false$ for
every world $w\in W$ and every primitive proposition $p$ not appearing
in $f$ means that we must describe $\pi$ only for propositions that
appear in $f$.)  We can verify that $M\sat f$ inductively. For
inequality formulas, let $e(\gamma)$ be an arbitrary expectation term in the
formula, with $\gamma$ of the form $b_1\phi_1+ \cdots+b_n\phi_n$. For
each $\phi$ in $\gamma$, we compute $\eventM{\phi}$ by checking the
truth assignment of each world in $W$ and seeing whether this truth
assignment makes $\phi$ true. We then replace each occurrence of
$e(\gamma)$ by
$\min_{\mu\in\cP}\{\sum_{i=1}^{n}\sum_{w\in\eventM{\phi_i}}b_i\mu(w)\}$
and verify that the resulting inequality holds. It is easy to see
that this verification can be done in time polynomial in $|f|$ and
$||f||$.  Therefore, the decision problem is in NP, and hence is
NP-complete.

Finally, consider $\MBel$ and $\MPoss$. As earlier, the lower bound
follows from the fact that we can reduce propositional satisfiability
to the decision problem for $\LE$; hence, the problem is
NP-hard. Again, the upper bound follows from
Theorem~\ref{t:smallmodel2}. Given a formula $f$, guess a small
model $M=(W,\nu,\pi)$ of the form guaranteed to exist by
Theorem~\ref{t:smallmodel2}, along with sets $U_1,\ldots,U_s$
($s\le|f|$) such that the mass function corresponding to the belief
function (respectively, possibility measure) $\nu$ is positive only
on $U_1,\ldots,U_s$. 
We then verify that $f$ is indeed true at some (and hence all) states in
the model, just as in the case of $\MLP$. 
\eprf
\subsection{Proofs for Section~\protect{\ref{s:gambleineq}}}\label{appa7}

The following lemmas are useful in the proof of
Theorem~\ref{t:gamblecompl}. 
\lem\label{l:phinonneg}
The formula $\phi\ge \tilde{0}$ is provable in $\AXg$.
\elem
\prf
Here is a sketch of the derivation:
\begin{enumerate}
\item $\truep=\tilde{1}$ (\axiom{G3})
\item $\phi\lor\neg\phi=\truep$ (\axiom{G4})
\item $\phi \lor \neg \phi = \phi+\neg\phi$  (\axiom{G1})
\item $\phi+\neg\phi=\tilde{1}$ (1, 2, 3, \axiom{IneqF}, \axiom{Taut}, \axiom{MP})
\item $\phi+\neg\phi\ge\tilde{0}$ (4, \axiom{IneqF})
\item $\phi\ge\tilde{0}$  (5, \axiom{G2}). 
\end{enumerate}
\eprf

\lem\label{l:phizero}
The formula $(a\phi+b\neg\phi\ge\tilde{0})\rimp(\phi=\tilde{0})$ is
provable in $\AXg$, for $a<0$.
\elem
\prf
Here is a sketch of the derivation:
\begin{enumerate}
\item $a\phi+b\neg\phi\ge\tilde{0} \rimp a \phi \ge \tilde{0}$
(\axiom{G2},\axiom{Taut},\axiom{MP})
\item $a\phi\ge\tilde{0} \rimp \phi \le \tilde{0}$ (Ineq{F}, since $a<0$)
\item $\phi\ge\tilde{0}$ (Lemma~\ref{l:phinonneg})
\item $a\phi+b\neg\phi\ge\tilde{0} \rimp \phi = \tilde{0}$
(1, 2, 3, \axiom{Taut}, \axiom{MP}, definition of $=$).
\end{enumerate}
\eprf

\othm{t:gamblecompl}
$\AXg$ is a sound and complete axiomatization of $\Lg$ with respect to
$\Mg$.
\eothm
\prf 
Soundness is straightforward.  For completeness, we show that an
unsatisfiable formula $f$ is inconsistent. We first reduce $f$ to a
canonical form. Let $g_1\lor\ldots\lor g_r$ be a disjunctive normal
form expression for $f$ (where each $g_i$ is a conjunction of gamble
inequalities and their negations). Using propositional reasoning
(axioms \axiom{Taut} and \axiom{MP}), we can show that $f$ is provably
equivalent to this disjunction. Since $f$ is unsatisfiable, each $g_i$
must also be unsatisfiable. Thus, it is sufficient to show that any
unsatisfiable conjunction of gamble inequalities and their
negations is inconsistent.

Let $f$ be such an unsatisfiable conjunction of gamble inequalities
and their negations. Let $p_1,\ldots,p_N$ be the primitive
propositions appearing in $f$, and let $\delta_1,\ldots,\delta_{2^N}$
be a canonical listing of the atoms over $p_1,\ldots,p_N$. We first
show that any gamble inequality $\gamma \ge \tilde{c}$ is provably equivalent
to a gamble 
inequality
$a_1\delta_1 + \cdots + a_{2^N}\delta_{2^N}\ge
\tilde{0}$. Consider a term $a\phi$ appearing in $\gamma \ge \tilde{c}$.
Since $\phi$ is equivalent to a disjunction
$\delta_{i_1}\lor\ldots\lor\delta_{i_k}$, we have that
$\phi=\delta_{i_1}\lor\ldots\lor\delta_{i_k}$, and hence
$a\phi=a(\delta_{i_1}\lor\ldots\lor\delta_{i_k})$,
is provable by \axiom{G4} and 
\axiom{IneqF}. By
repeated applications of \axiom{G1}, we have that
$a(\delta_{i_1}\lor\ldots\lor\delta_{i_k})=a\delta_{i_1}+ \cdots+a\delta_{i_k}$
is provable.
(Note that if $\phi$ and $\phi'$ are mutually exclusive, then $\phi \lor
\phi'$ is equivalent to $\phi \land \phi \lor \phi' \land \neg \phi$, so
by \axiom{G1}, $\phi \lor \phi'  = \phi + \phi'$.)
By \axiom{IneqF}, $a\phi=a\delta_{i_1}+\cdots+a\delta_{i_k}$ is provable too.
Using \axiom{IneqF} again, it follows that 
$a\phi=a_1\delta_1 + \cdots  + a_{2^N}\delta_{2^N}$ is provable, where
$a_i=a$ if $i\in\{i_1,\ldots,i_k\}$, and $a_i=0$ otherwise. 
Doing this to every term in $\gamma$ shows that here exist $b_1, \ldots,
b_{2^n}$ such that $\gamma=b_1\delta_1+\cdots+b_{2^N}\delta_{2^N}$ is
provable.  Hence, 
using \axiom{IneqF}, so is
$a_1\delta_1+ \cdots+a_{2^N}\delta_{2^N} \ge \tilde{c}$. 
By \axiom{G3}, $\truep=\tilde{1}$ is provable, hence so is
$c~\truep=\tilde{c}$.  It then easily follows using \axiom{IneqF} that
$c\delta_1+ \cdots+c\delta_{2^N}=\tilde{c}$ is provable. Thus,
$\gamma \ge \tilde{c}$ is provably equivalent to
$(a_1-c)\delta_1+ \cdots+(a_{2^N}-c)\delta_{2^N}\ge \tilde{0}$,
as required. 

It immediately follows that $f$ is provably equivalent to a formula $f'$
that is a 
conjunction of gamble inequalities of the form
$a_1\delta_1+ \cdots+a_{2^N}\delta_{2^N}\ge \tilde{0}$ and their
negations. 
Say that $f'$
consists of $r$ gamble inequalities and $s$ negations of gamble
inequalities. Consider two arrays of coefficients of $f''$:
$P=(a_{i,j})$, where $a_{i,j}$ is the coefficient of $\delta_j$ in
$a_{i,1}\delta_1+ \cdots+a_{i,2^N}\delta_{2^N}\ge \tilde{0}$ ($1\le i\le
r$), and $N=(b_{i,j})$, where $b_{i,j}$ is the coefficient of
$\delta_j$ in
$\neg(b_{i,1}\delta_1+ \cdots+b_{i,2^N}\delta_{2^N}\ge \tilde{0})$ ($1\le
i\le s$). Let $I=\{1,\ldots,2^N\}$.
Note that, since $f$ is unsatisfiable, so is $f'$.
If $a_{i,j} < 0$ for some $i, j$, then, by 
Lemma~\ref{l:phizero}, we must have $\delta_j=\tilde{0}$. 
Let $I'' = \{j: a_{i,j} < 0 \mbox{ for some } i\}$; let $I' = I - I''$.
Let $P'$ and $N'$ be the result of setting all entries in column $j$ of
$P$ (resp., $N$) to 0, for all $j \in I''$.  
The formula $f''$ corresponding to the matrices $P'$ and $N'$ is
provably equivalent to the original $f'$, by \axiom{IneqF}:~since
$\delta_j=\tilde{0}$ for all $j \in I''$, it must be the case that
$a\delta_j=\tilde{0}$. 
By construction, all entries in $P'$ and $N'$ are nonnegative; moreover, 
$N'$ is nonempty (since
$\neg(-\delta_1-\cdots-\delta_{2^N}\ge \tilde{0})$ is in $f''$).

There are now two cases.
Taking $N' = (b_{i,j}')$, note that if $N'$ has a row $i$ with
all entries nonnegative, then $f''$ provably implies
that $\neg(0\delta_1+ \cdots+0\delta_{2^N} \ge
\tilde{0})$.  But using \axiom{IneqF}, it is easy to show that 
$ 0\delta_1+ \cdots+0\delta_{2^N} = \tilde{0}$.  This shows that $f''$,
and hence $f'$ and $f$, is inconsistent.
On the other hand,
if all the rows in $N'$ have a negative entry,
then the formula corresponding to $P',N'$ is in fact satisfiable.
We can construct a structure $M$ satisfying $f'$ by taking
$M=(\{\delta_i ~:~ i\in I'\},\pi)$, where $\pi(\delta_i)(p)=\true$ if
$\delta_i\rimp p$ is a propositional tautology.
This contradicts the assumption that $f'$ is unsatisfiable.
\eprf

The proof of Theorem~\ref{t:decprocLg} relies
on the following small-model result.

\lem\label{l:smallmodelLg}
Suppose that $f\in\Lg_1$ is satisfied in some structure in $\Mg$. Then $f$ is
satisfied in a structure $(W,\pi)$ where $|W|\le|f|$, and
$\pi(w)(p)=\false$ for every world $w\in W$ and every primitive
proposition $p$ not appearing in $f$. 
\elem
\prf
Suppose that $f\in\Lg_1$ is satisfied in a structure $M=(W,\pi)$. 
conjunction of gamble inequalities and their negations. 
We want to show that $f$ is in fact satisfied in a small structure.
As usual, we can assume without loss of generality that $f$ is a 
conjunction of gamble inequalities and their negations. 

Suppose that there are $r$ gamble inequalities in $f$ and $s$
negations of gamble inequalities. We consider two cases.
If $s=0$, then pick any $w\in W$ and let
$M'=(W',\pi')$, with $W'=\{w\}$, and $\pi'$ is the restriction of
$\pi$ to $\{w\}$ (setting $\pi(w)(p)=\false$ for every primitive
proposition $p$ not appearing in $f$. It is easy to check that $M'\sat 
f$, since $w\in W$. 
If $s>0$, then for every negation of gamble inequality
$\neg(\gamma\ge \tilde{c})$ appearing in $f$, there exists a world $w\in W$
such that $\gambM{\gamma}(w)< \tilde{c}$. Let $w_1,\ldots,w_s$ be such worlds,
one corresponding to each of the $s$ negation of gamble inequalities
appearing in $f$. Let $M'=(W',\pi')$, where $W'=\{w_1,\ldots,w_s\}$,
and $\pi'$ is the restriction of $\pi$ to $W'$ (setting
$\pi'(w)(p)=\false$ for primitive propositions $p$ not appearing in
$f$). Clearly, every gamble inequality in $f$ is satisfied in $M'$,
since $W'$ is a subset of $W$. Moreover, by choice of $W'$, every
negation of gamble inequality in $f$ is also satisfied in $M'$ (since
our construction guarantees that there is world in $M'$ that is a
witness to the falsity of all the negated gamble inequalities in $f$).
Since $|W'|=s<|f|$, the result follows.
\eprf

\othm{t:decprocLg} The problem of deciding whether a formula of
$\Lg_1$ is satisfiable in $\Mg$ is NP-complete.
\eothm
\prf
For the lower bound, observe that $\Lg_1$ includes propositional
reasoning. Hence, the decision problem for  $\Lg_1$ is at least as
hard as propositional reasoning.

For the upper bound, let $f$ be a satisfiable formula of $\Lg_1$. We
first guess a small model $M=(W,\pi)$ for the formula, of the form
guaranteed to exist by Lemma~\ref{l:smallmodelLg}. (As usual, the fact that
$\pi(w)(p)=\false$ for every world $w\in W$ and every primitive
proposition $p$ not appearing in $f$ means that we must describe $\pi$
only for propositions that appear in $f$.) We verify that $M\sat f$
inductively. For basic formulas of the form $\gamma\ge c$, we must
check that for all $w\in W$, $\gambM{\gamma}(w)\ge c$. 
If $\gamma$ is of the form $b_1\phi_1+ \cdots+b_n\phi_n$, then we can
compute $\gambM{\gamma}(w)$ by summing all the $b_i$ such that $\phi_i$ is
true at $w$ (i.e., $\pi(w)(\phi_i)=\true$). It is easy to see that
this verification can all be done in time polynomial in $|f|$ and
$||f||$. 
\eprf

The following notation is useful for the proofs of
Theorem~\ref{t:ineqfuncompl} and Lemma~\ref{l:smallmodelLineqf}. Given
a linear inequality formula $f$ (over real-valued functions)
$t\ge\tilde{c}$, we write $\hat{f}$ for the linear inequality formula
$t\ge c$ over the reals. 
We extend this to Boolean combinations of linear inequality formulas
in the obvious way.

For the sake of our proof of completeness of $\AXf$, we need also
to show that the following formula is provable:
\begin{equation}\label{zero}
0 v_1 + \ldots + 0 v_n \ge \tilde{0}
\end{equation}
This formula can be viewed as saying that the right implication of
axiom \axiom{I5} holds when $d=0$.
\lem\label{zero1}
The formula~(\ref{zero}) is provable from $\AXf$.
\elem
\prf
By \axiom{F1}, $v_1 \geq v_1$, that is,
$v_1 - v_1 \geq \tilde{0}$, is provable.
By axiom \axiom{F3}, 
so is
$-v_1 + v_1 \geq \tilde{0}$.
If we add these latter two inequalities by \axiom{F4}, and
delete a 0 term by \axiom{F2}, we
obtain $0 v_1 \geq \tilde{0}$.
By using \axiom{F2} to add 0 terms, 
it follows that $0 v_1 + \ldots + 0 v_n \ge \tilde{0}$ is provable, as
desired. 
\eprf

\othm{t:ineqfuncompl}
$\AXf$ is sound and complete for reasoning about formulas about linear 
inequalities over real-valued functions with nonempty domain.
\eothm
\prf
Soundness is straightforward.  For completeness, we show that an
unsatisfiable formula $f$ is inconsistent. 
So suppose that $f$ is unsatisfiable.
As usual, without loss of generality, we can assume that $f$ is a
conjunction of inequalities and their negations, say, with $r$
inequalities and $s$ negations of inequalities.
We prove the result by reducing 
satisfiability of inequalities over real-valued functions to 
satisfiability of
inequalities over real numbers, and then apply techniques from FHM. 

There are two cases.  First, suppose that $s=0$, so that there are no
negations of inequalities in $f$. 
It is easy to see that since $f$ is
unsatisfiable over functions, $\hat{f}$ must be unsatisfiable over the
reals.  For if $\hat{f}$ were satisfiable over the reals with a solution
$x_1^*,\ldots,x_k^*$, then $f$ would be 
satisfied by taking $x_i$ to be the constant function that always
returns $x_i^*$.

Write $\hat{f}$ in matrix form as $A\vec{x}\ge \vec{b}$,
where $A$ is the $r \times k$ matrix of coefficients on the left-hand
side of the inequalities, $\vec{x}$ is the column vector $(x_1 , \ldots ,
x_k )$, and $\vec{b}$ is the column vector of the right-hand sides of the
inequalities. Since $\hat{f}$ is unsatisfiable, $A\vec{x}\ge \vec{b}$ is
unsatisfiable. As in FHM, we make use of the following variant of
Farkas' lemma \cite{Farkas} (see \citeN[page 89]{Schrijver}) from
linear programming. 
\lem\label{farkas}
If $A\vec{x} \geq \vec{b}$ is unsatisfiable, then there exists a row
vector 
$\vec{\sigma}$ such that
\begin{enumerate}
\item $\vec{\sigma} \geq \vec{0}$;
\item $\vec{\sigma} A = \vec{0}$;
\item $\vec{\sigma} \cdot \vec{b} > 0$.
\end{enumerate}
\elem
Intuitively, $\vec{\sigma}$ is a ``witness'' or ``blatant proof'' of the
fact that $Ax \geq b$ is unsatisfiable.  This is because if there were
a vector $\vec{x}$ satisfying $A\vec{x} \geq \vec{b}$, then $0 =
(\vec{\sigma} A)\vec{x} = \vec{\sigma} 
(A\vec{x}) \geq \vec{\sigma} \vec{b} > 0$, a contradiction.

We now show
that $f$ must be inconsistent.  Let $\vec{\sigma} = (\sigma_1 , \ldots,
\sigma_{r})$ be the row vector guaranteed to exist by Lemma~\ref{farkas}.
Either by \axiom{F5} or by Lemma~\ref{zero1} (depending on whether
$\sigma_j \gt 0$ or $\sigma_j = 0$), we can multiply 
both sides of
the $j^{\rm th}$
conjunct of $f$ by $\sigma_j$ (for $1 \leq j \leq r$), and then use
\axiom{F4} to add the resulting inequality formulas together.  The net
result (after deleting some 0 terms by \axiom{F2}) is the formula $ (0
v_1 \geq \tilde{c} ) $, where $c = \vec{\sigma} \cdot \vec{b} > 0$. From
this formula, 
by \axiom{F6}, we can conclude $ (0 v_1 \gt \tilde{0}) $, which by the
definition of $>$ implies $ \neg(0 v_1 \leq \tilde{0} ) $, which is in turn an
abbreviation for $ \neg(-0 v_1 \geq \tilde{-0} ) $, that is, $ \neg(0 v_1 \geq
\tilde{0}) $.  Thus $f \rimp \neg(0 v_1 \geq \tilde{0})$ is provable. However, by
Lemma~\ref{zero1}, $(0 v_1 \geq \tilde{0})$ is also provable.  It follows by
propositional reasoning that $\neg f$ is provable, that is, $f$ is
inconsistent, as required.

Now suppose that $s>0$. Let $f^+$ be the conjunction of
the inequalities in $f$, and 
let
$g_1,\ldots,g_s$ be the negations of
inequalities in $f$. 
That is, $f = f^+ \land g_1 \land \ldots \land g_s$. 
We first show that, 
$\hat{f}^+\land \hat{g}_i$ must be unsatisfiable over the reals for some
$i \in \{1, \ldots, s\}$.  Assume by
way of 
contradiction that this is not the case, that is, for all $1\le i\le
s$, $\hat{f}^+\land\hat{g}_i$ is satisfiable over the reals. Let
$x_{1,i}^*,\ldots,x_{k,i}^*$ be the real number solution to the
inequalities $\hat{f}^+\land\hat{g}_i$, and let $D=\{d_1,\ldots,d_s\}$.
Define  the functions $F_1^*,\ldots,F_k^*$ over $D$ by taking
$F_j^*(d_i)=x_{j,i}^*$. It is easy to verify that
$f=f^+\land g_1\land\ldots\land g_s$ is satisfied by those
functions. Intuitively, for every element $d_i$ of the domain,
$\hat{f}^+$ is satisifed by $F_1^*(d_i),\ldots,F_k^*(d_i)$, so that
$f^+$ is satisfied by $F_1^*,\ldots,F_k^*$; moreover, each $g_i$
is also 
satisfied, since by the choice of $d_i$, we have
$a_1F^*_1(d_i)+\ldots+a_k F^*_k(d_i)<c$. Hence, $f$ is satisfiable over
real-valued functions, a contradiction. Therefore, 
$\hat{f}^+\land\hat{g}_{i_0}$ is unsatisfiable over the real numbers for 
some $i_0$. As in the case $s=0$, we use this fact to show that $f$
is inconsistent. 

As before, we can write $\hat{f}^+$ in
matrix form as $A\vec{x}\ge \vec{b}$. 
Similarly, the formula
$\hat{g}_{i_0}=\neg(a_1x_1+\ldots+a_kx_k\ge c)$ can be written in matrix
form as $A'\vec{x}>-c$, where $A'$ is the $1 \times s$ matrix
$[-a_1,\ldots,-a_k]$, and $\vec{x}$ is the column vector
$(x_1,\ldots,x_k)$. Since $\hat{f}^+\land\hat{g}_{i_0}$ is 
unsatisfiable, the system $A\vec{x}\ge \vec{b},A'\vec{x}>-c$ must be
unsatisfiable.  
Farkas' lemma does not apply, but a variant of it, called
Motzkin's transposition theorem, which is due to Fourier
\citeyear{Fourier}, Kuhn \citeyear{Kuhn}, and Motzkin \citeyear{Motzkin}
(see \citeN[page 94]{Schrijver}), does. 
\lem\label{motzkin}
If the system $A\vec{x} \geq \vec{b}, A'\vec{x} \gt -c$ is
unsatisfiable, then there 
exist a row vector $\vec{\sigma}$ and a real $\sigma'$ such that 
\begin{enumerate}
\item $\vec{\sigma} \geq \vec{0}$ and $\sigma' \geq \vec{0}$;
\item $\vec{\sigma} A + \sigma' A' = \vec{0}$;
\item either
  \begin{enumerate}
  \item $\sigma' = 0$ and $\vec{\sigma} \cdot \vec{b}  \gt 0$, or 
  \item $\sigma' > 0$  and
  $\vec{\sigma}\cdot \vec{ b} - \sigma' c \geq 0$
  \end{enumerate}
\end{enumerate}
\elem
\commentout{
We now show that $\sigma$ and $\sigma'$ together form a witness to the
fact that the system $Ax \geq b, A'x \gt b'$ is unsatisfiable.  Assume
that there were $x$ satisfying $Ax \geq b$ and $A'x > b'$.  In case
(3a) of Lemma \ref{motzkin} ($\sigma b + {\sigma'} b' \gt 0$), we are
in precisely the same situation as in Farkas' lemma, and the argument
after Lemma \ref{farkas} applies.  In case (3b) of Lemma
\ref{motzkin}, let $\Delta = (A'x) - b'$; thus, $\Delta$ is a column
vector and $\Delta > 0$.  Then $0 = (\sigma A + {\sigma'} A')x =
(\sigma A)x + ({\sigma'} A')x = \sigma (Ax) + {\sigma'} (A'x) \geq
\sigma b + {\sigma'} (b' + \Delta ) = (\sigma b + {\sigma'} b' ) +
{\sigma'} \Delta \geq {\sigma'} \Delta > 0$, where the last inequality
holds since every $\sigma_j '$ is nonnegative, some $\sigma_j '$ is
strictly positive, and every entry of $\Delta$ is strictly positive.
This is a contradiction.
}

Since 
$Ax\ge b,A'x>-c$ is unsatisfiable, let $\vec{\sigma} =
(\sigma_1 , \ldots, \sigma_{r})$ and $\sigma'$ be the row vector
and real guaranteed to exist by Lemma \ref{motzkin}. 
If case 3(a) of Lemma~\ref{motzkin} applies, then the situation is
identical to that of Lemma~\ref{farkas}, and the same argument shows
that $f$ is inconsistent.  
If case (3b) of Lemma~\ref{motzkin} applies, then $\sigma' > 0$.
As before, either by axiom \axiom{F5} or by Lemma~\ref{zero1},
we can
multiply 
both sides of 
the $j^{\rm th}$ conjunct in the formula $f^+$
by $\sigma_j$,
for $1 \leq j \leq r$.
This results in the following system of inequalities:
\begin{eqnarray}
\sigma_{1} \theta_{1,1} v_1 + \cdots +
  \sigma_{1} \theta_{1,k} v_k & \geq &
    \widetilde{\sigma_1 c_1} \nonumber \\
                          &\cdots&   \label{ineqdd}\\
\sigma_{r} \theta_{r,1} v_1 + \cdots +
  \sigma_{r} \theta_{r,k} v_k
  & \geq & \widetilde{\sigma_{r}c_r}. \nonumber
\end{eqnarray}
Similarly, by \axiom{F5}, we can multiply both sides of $g_{i_0}$ by
$\sigma'$ to get 
$\neg(\sigma' a_1 v_1 + \ldots \sigma' a_k v_k \ge \widetilde{\sigma' c'})$.

Let $\theta''_1 v_1 + \cdots + \theta''_k v_k \geq \tilde{d}$ be the
result of ``adding'' all the inequalities in (\ref{ineqdd}).  This
inequality is provable from $f$ using \axiom{F4}.  Since $\vec{\sigma} A +
{\sigma'} A' = \vec{0}$, and 
$A' = [-a_1,\ldots,-a_k]$, 
we must have that $-\sigma' a_j = -\theta''_j$, 
for $j=1, \ldots, k$.  
Thus, $g_{i_0} \rimp 
\neg(\theta''_1 v_1 + \cdots + \theta''_k v_k \ge \tilde{\sigma' c'}) $
is provable.
Since
$\sigma b - \sigma'c' \geq 0$, 
it follows that $d\ge\sigma'c'$. 
Therefore, 
by \axiom{F6},
$\theta''_1v_1+\cdots\theta''_kv_k\ge \tilde{d} \rimp
\theta''_1v_1+\cdots\theta''_kv_k\ge\widetilde{\sigma'c'}$ is provable.
Taking the contrapositive, we get that 
$g_{i_0} \rimp\neg(\theta''_1v_1+\cdots+\theta''_kv_k\ge \tilde{d})$ is
provable.  
But $f^+ \rimp \theta''_1v_1+\cdots+\theta''_kv_k\ge \tilde{d}$ is also
provable.  
Since $f \rimp f^+ \land g_{i_0}$ is obviously provable, 
it follows by propositional
reasoning that $\neg f$ is provable, that is, $f$ is inconsistent, as
desired. 
\eprf

As with $\Lg_1$, the proof of NP-completeness for the decision problem
of formulas about linear inequalities (with integer coefficients) over 
real-valued functions with nonempty domain relies on the following
small-model result. Here, a ``small model'' for a formula is an
assignment of functions with a small domain. This can be done quickly,
as we shall see. 
\lem\label{l:smallmodelLineqf}
Suppose that $f$ is a satisfiable inequality formula. Then $f$ has a
satisfying assignment where there are at most $|f|$ functions in the
assignment with a non-zero range, the functions in the assignment have a
domain of size at most $|f|$, and every value in the range of the
functions is a rational number with size $O(|f| ||f||+|f|\log(|f|))$. 
\elem
\prf
As usual, we can assume without loss of generality that 
$f$ is a conjunction of inequality formulas and their
negations.  Suppose that $f$ has a satisfying assignment
$v_1^*,\ldots,v_k^*$, over a domain $D$.
Let $f^+$ be the conjunction of the inequality formulas in $r$,
and let $g_1,\ldots,g_s$ be the negations of inequality formulas in
$f$. Consider two cases, depending on whether $s=0$. If $s=0$, 
pick some element $d$ of $D$. 
Clearly, $v^*_1(d),\ldots,v^*_k(d)$ is a solution to $\hat{f}$, and
thus $\hat{f}$ is satisfiable over the reals. By Theorem~4.9 in FHM,
$\hat{f}$ is satisfiable with a solution $x^*_1,\ldots,x^*_k$ where at
most $|f|$ entries are nonzero, and each nonzero value is a rational
number of size $O(|f| ||f|| + |f|\log(|f|))$. From this solution, we
can construct a solution $F_1^*,\ldots,F_k^*$ on the domain $D'=\{d\}$
satisfying the conditions of the theorem, by simply taking
$F_i^*(d)=x^*_i$.  

Now suppose that $s>0$. 
Assume $f$ is satisfiable over functions, with a solution
$F^*_1,\ldots,F^*_k$ over a domain $D$. Clearly, for any $d\in D$,
$F^*_1(d),\ldots,F^*_k(d)$ is a solution to $\hat{f}^+$. For any $1\le
i\le s$, consider $g_i=\neg(a_1v_1+\ldots+a_kv_k\ge\tilde{c})$. There
must be a $d\in D$ such that $a_1F^*_1(d)+\ldots+a_kF^*_k(d)<c$. Thus,
$F^*_1(d),\ldots,F^*_k(d)$ is a solution to $\hat{g}_i$, and by the
above is also a solution to $\hat{f}^+$. Thus, $\hat{f}^+\land\hat{g}$
is satisfiable over the reals. Again by Theorem~4.9 in FHM, we have a
solution $x_{1,i}^*,\ldots,x_{k,i}^*$ of the inequalities
$\hat{f}^+\land\hat{g}_i$
such that at most $|f|$ entries in the solution are nonzero,
and each nonzero value is of size $O(|f| ||f|| + |f|\log(|f|))$. Let
$D=\{d_1,\ldots,d_s\}$.
We construct a solution to $f^+\land g_1\land\dots\land g_s$ from those 
solutions. 
Define the functions $F_1^*,\ldots,F_k^*$ over $D$ such that
$F_j^*(d_i)=x_{j,i}^*$.  it is easy to verify that
$f=f^+\land g_1\land\ldots\land g_s$ is satisfied by those
functions. Intuitively, for every element $d_i$ of the domain,
$\hat{f}^+$ is satisfied by $F_1^*(d_i),\ldots,F_k^*(d_i)$; moreover, 
each $g_i$ (of the form $\neg(a_1F_1+\ldots+a_kF_k\ge\tilde{c})$ must
be satisfied, since by choice of $d_i$, we have
$a_1F^*_1(d_i)+\ldots+a_kF^*_k(d_i)<c$. 
Clearly this assignment
$F_1^*,\ldots,F_k^*$ satisfies the statement of the lemma.
\eprf

\othm{t:decprocLineqf} The problem of deciding whether a formula
about linear inequalities (with integer coefficients) 
is satisfiable
over real-valued functions with nonempty domain is NP-complete.
\eothm
\prf
For the lower bound, note that we can reduce propositional
satisfiability to satisfiability in the logic of linear inequalities by
simply replacing each primitive proposition $p_i$ in a propositional
formula by the inequality $v_i > 0$.  

For the upper bound, let $f$ be a satisfiable inequality formula. We
guess a small satisfying assignment where the domain $D$ of the
functions has size at most $|f|$; such an assignment is guaranteed to
exist if $f$ is satisfiable, by Lemma~\ref{l:smallmodelLineqf}.
It is easy to verify that this assignment does indeed satisfy $f$ in 
time polynomial in $|f|$ and $||f||$. 
\eprf

\end{document}